\documentclass{article}
\usepackage[T1]{fontenc}
\usepackage{iftex}
\ifPDFTeX
  \usepackage[utf8]{inputenc}
\fi
\usepackage[preprint]{colm2025_conference}

\ifPDFTeX\else
  \usepackage{newunicodechar}
  \newunicodechar{–}{\textendash}
  \newunicodechar{—}{\textemdash}
  \newunicodechar{×}{\ensuremath{\times}}
  \newunicodechar{í}{\'{i}}
  \newunicodechar{á}{\'{a}}
  \newunicodechar{ä}{\"{a}}
\fi

\usepackage[letterpaper,margin=1.25in]{geometry}

\usepackage{graphicx}
\usepackage{amsmath,amsfonts,amssymb,amsthm,mathtools}
\usepackage{array,tabularx,booktabs,multirow,colortbl}
\usepackage{enumitem}
\usepackage{float}
\usepackage{adjustbox}
\usepackage{siunitx}
\usepackage{xspace}

\usepackage{algorithm,algpseudocode}

\usepackage{hyperref}
\usepackage{cleveref, eucal}

\definecolor{mblue}{HTML}{0064E0}
\hypersetup{
    colorlinks=true,
    linkcolor=mblue,
    citecolor=mblue,
    urlcolor=mblue
}

\newcommand{\modelname}{InternW0-$\Delta$\xspace}

\newif\ifshowowners
\showownersfalse
\DeclareRobustCommand{\owners}[1]{%
  \ifshowowners\quad{\normalfont\small\color{gray}#1}\fi}

\newif\ifshowedits
\showeditsfalse

\hypersetup{pdftitle={InternW0-Delta: An Embodied World Model Bridging Predictive Dynamics and Actions}}

\title{\modelname: A World Action Model Bridging Predictive Dynamics and Actions with 20K+ Hours of Open Data}

\author{%
Physical Intelligence Team, Shanghai AI Laboratory\\[2pt]
{\normalfont\textbf{Project page:} 
\url{https://internrobotics.github.io/InternW0-Delta/}}
}

\date{September 2026}

\begin{document}
\maketitle

\begin{figure*}[!h]
    \centering
    \includegraphics[width=\linewidth]{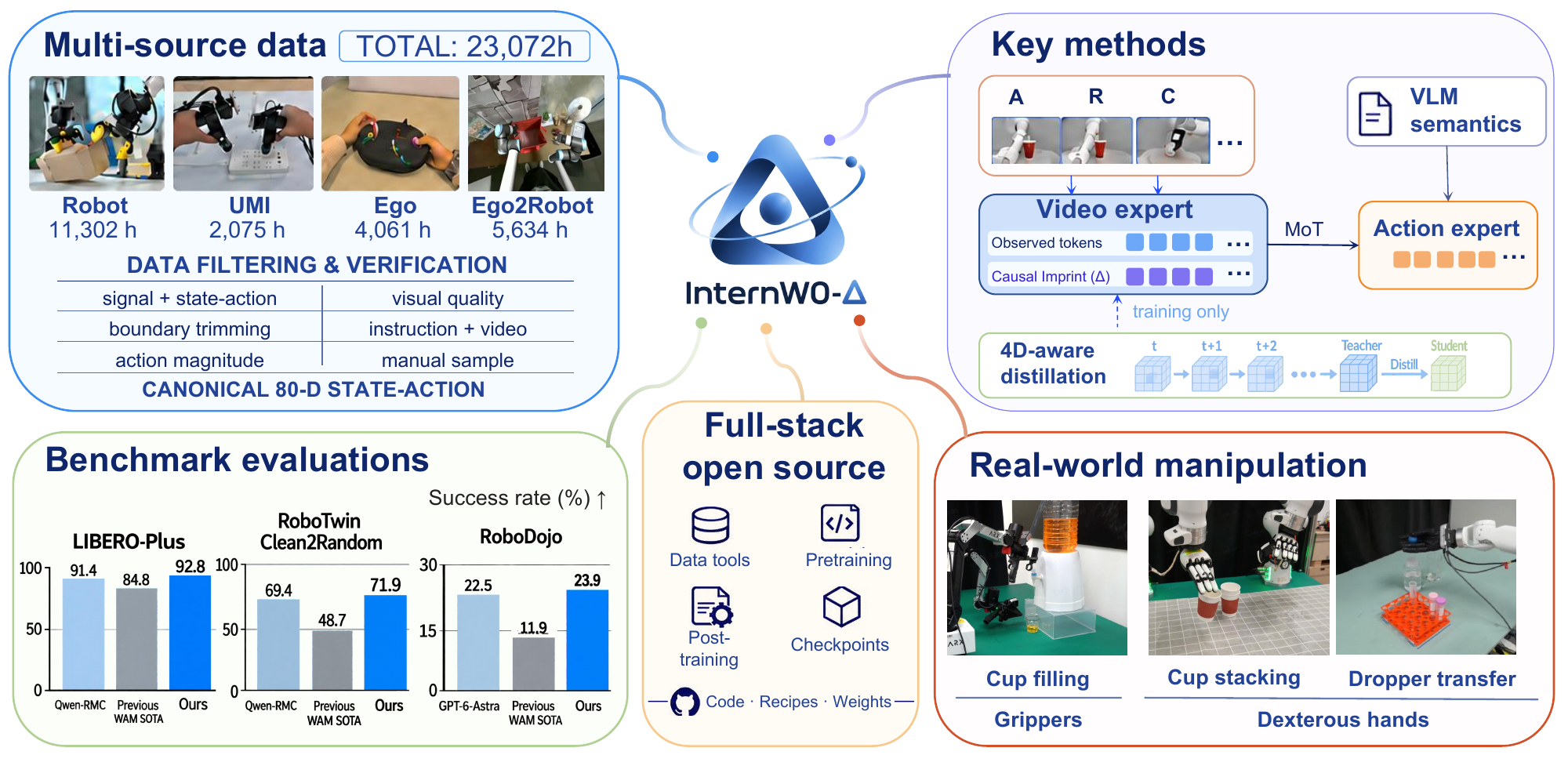}
    \label{fig:teaser}
\end{figure*}

\begin{abstract}
World Action Models (WAMs) have emerged as a promising paradigm for generalist robot manipulation by jointly modeling visual dynamics and action generation. A central challenge is how to effectively integrate complementary priors from large-scale pretrained models---including visual dynamics, scene semantics, and geometric and motion understanding---into a unified framework for robot action generation. We introduce \modelname, a unified World Action Model that meets this challenge: pretrained on a large-scale heterogeneous corpus, it outperforms prior methods across diverse simulation benchmarks and real-robot platforms.

\modelname
brings together pretrained visual dynamics, scene-level semantic understanding, 4D geometric and motion priors, and action generation within a {\it Mixture-of-Transformers (MoT) framework}. Within the World–Action MoT, a pretrained video expert and an action expert interact under scene-grounded semantic guidance from a frozen VLM, while a pretrained 4D foundation model injects geometric and motion priors through training-only distillation. To translate predictive visual dynamics into representations useful for action prediction, we introduce {\it  Causal Imprint},  which learns future-relevant scene changes from training-only future supervision and makes these predictive representations directly available to the action expert without requiring future-video rollout at inference. 

To support large-scale joint training, we construct a heterogeneous corpus spanning robot demonstrations, UMI data, egocentric human demonstrations, and Ego2Robot data---carefully curated and filtered, unified under a common state-action representation, and temporally aligned---yielding
over \textbf{20K} hours of processed training data, {\it to our knowledge the largest open-source corpus of its kind}.   We pretrain \modelname on this heterogeneous corpus and demonstrate strong performance across diverse simulation benchmarks and real-robot platforms. We will open source training code and the model weights, infrastructure, and data-processing pipeline, together with processed data where licenses permit, to accelerate progress in embodied intelligence and physical AI.

\vspace{10pt}

\end{abstract}

\section{Introduction}
World Action Models (WAMs) offer a promising approach to generalist robot manipulation by jointly modeling visual dynamics and robot actions~\citep{ye2026dreamzero,bi2025motus}. A key motivation behind this formulation is that large-scale video pretraining can provide strong visual and temporal knowledge about how scenes evolve over time. However, the ability to predict future observations does not directly translate into effective robot control. Action generation further requires identifying task-relevant changes, understanding object geometry and motion, and grounding these cues in the current instruction and scene. The central challenge\footnote{This challenge aligns with the broader goal of the InternW series: connecting perception, physical prediction, and action under limited sensing and computation~\citep{chen2026definition}.} is therefore to transfer predictive knowledge from visual dynamics modeling into representations that are directly useful for action generation, without requiring explicit future generation during online control~\citep{yuan2026fastwam}.

To face this challenge, we introduce \modelname, a directed world-action architecture that combines predictive visual dynamics, temporal context, and task-conditioned scene semantics for action generation.
At its core, a pretrained video expert and an action expert are coupled through a directed Mixture-of-Transformers architecture.
The video expert processes a lightweight sparse memory of anchor, recent, and current observations, providing both episode-level context and recent interaction history, while a frozen vision-language model (VLM) supplies task-conditioned scene semantics to the action expert.
To make predictive dynamics directly useful for control, Causal Imprint learns future-relevant scene changes from training-only future supervision and makes these representations available to the action expert.
In parallel, training-only 4D-aware distillation from a Track4World~\citep{lu2026track4world} teacher injects geometric and motion priors into the video expert through auxiliary supervision.
The directed information flow ensures that neither Causal Imprint nor the action expert takes future observations as input, with future information used only as training supervision. This allows \modelname to directly predict actions at inference without sampling future videos or invoking the distillation branch.

To support large-scale joint training, we curate a heterogeneous corpus spanning robot demonstrations, UMI data, egocentric human demonstrations, and Ego2Robot data, drawing primarily on public datasets.
These sources differ substantially in robot embodiment, control space, camera configuration, and temporal convention.
We therefore convert them into a canonical state-action representation and apply systematic quality filtering and temporal alignment, resulting in over \textbf{20K} hours of processed training data.
On this corpus, we adopt a two-stage training recipe that first pretrains \modelname to jointly learn visual dynamics and action generation, and then adapts the resulting checkpoint to target embodiments and tasks through post-training.

In addition, we develop complementary infrastructure to support efficient model iteration and online execution.
For model development, we optimize the training pipeline to reduce the cost of repeated architecture and hyperparameter experiments.
Caching video autoencoder latents and frozen vision-language features avoids redundant encoding across repeated training runs, while layerwise compilation and activation checkpointing improve backbone throughput and memory efficiency.
Together, these optimizations substantially accelerate model iteration and make large-scale experimentation more practical.
For deployment, context caching and compiled action execution reduce inference overhead and support asynchronous action-chunk execution.
On the physical dexterous-hand deployment, the optimized runtime achieves an average controller-observed round-trip latency of 152.8\,ms on a single NVIDIA RTX~5090 GPU, corresponding to a $5.11\times$ speedup over the standard runtime.

We will release code, checkpoints, recipes, and infrastructure covering the entire pipeline, from data processing and two-stage training to evaluation and deployment. Processed data will be shared where licenses permit, and versioned indices will reference the original samples and record our filtering decisions, so that the community can more easily reproduce our work.

We evaluate \modelname on LIBERO-Plus~\citep{libero_plus}, RoboTwin~2.0~\citep{robotwin2}, EBench~\citep{ebench}, and RoboDojo~\citep{robodojo}, spanning different embodiments, task demands, and distribution shifts.
Post-trained only on unperturbed demonstrations, \modelname achieves the best success rates under distribution shift on both LIBERO-Plus (92.8\%) and RoboTwin~2.0 Clean2Random (71.9\%), while also leading on Clean2Clean (90.0\%).
On EBench, which targets mobile bimanual manipulation, it obtains the highest overall score of 66.0.
On RoboDojo, whose memory, precision, and long-horizon tasks remain challenging for all methods, it achieves the best average success rate of 23.9\%, nearly double that of the strongest prior WAM.
We further demonstrate real-robot deployment on two gripper-based and two dexterous-hand platforms, adapting the same pretrained checkpoint to their respective control interfaces through post-training.

The main contributions of this work are as follows:
\begin{enumerate}
    \item \textbf{A World Action Model with action-relevant predictive representations.}
    \modelname couples a pretrained video expert and an action expert through a directed Mixture-of-Transformers. Causal Imprint learns future-relevant scene changes for the action expert without future-video sampling at inference, while training-only 4D-aware distillation adds geometric and motion priors to the video expert.

    \item \textbf{A scalable and reproducible data-to-deployment recipe.}
    We curate and unify over 20K hours of heterogeneous robot and human demonstrations under a canonical state-action representation, pretrain on this corpus, and adapt the resulting checkpoint to target embodiments through post-training. Our infrastructure further speeds up both model iteration and online execution. We will release the code, checkpoints, recipes, and filtered-data indices.

    \item \textbf{Comprehensive evaluation across embodiments, from simulation to real robots.}
    We evaluate \modelname on LIBERO-Plus, RoboTwin~2.0, EBench, and RoboDojo, covering single-arm, bimanual, and mobile manipulation under diverse task demands and distribution shifts. We further deploy it on gripper-based and dexterous-hand real robots, adapting the same pretrained checkpoint to each platform through post-training.
\end{enumerate}

\section{Related Work}

\subsection{Vision--Language--Action and World Action Models}

Vision--language--action (VLA) models transfer the semantic knowledge of pretrained vision--language models to robot control. RT-2~\citep{brohan2023rt2} and OpenVLA~\citep{kim2024openvla} adapt these backbones to predict tokenized actions, while $\pi_0$~\citep{pi0} and GR00T N1~\citep{bjorck2025groot} couple multimodal understanding with continuous action generation. Subsequent efforts expand the breadth of policy learning: $\pi_{0.5}$~\citep{pi05} combines heterogeneous training sources for open-world generalization, LingBot-VLA~\citep{wu2026lingbotvla,wu2026lingbotvla2} studies large-scale cross-embodiment learning and practical adaptation, and Qwen-RobotManip~\citep{qwen_robotmanip} emphasizes alignment across heterogeneous manipulation data. These works establish strong semantic and instruction-following foundations for generalist policies.

World Action Models (WAMs) additionally couple action learning with predictions of how the visual world evolves. DreamZero~\citep{ye2026dreamzero} transfers pretrained video priors through joint video--action prediction, while Motus~\citep{bi2025motus} integrates understanding, video generation, and action modeling within a mixture-of-transformers architecture. LingBot-VA~\citep{lingbot-va2026} adopts causal video--action modeling for streaming control, and LingBot-VA 2.0~\citep{zhang2026lingbotva2} further explores native video--action pretraining. A key design choice is whether action inference requires generating future video. Fast-WAM~\citep{yuan2026fastwam} separates future-video supervision from the action inference path, showing that video prediction can benefit policy learning without test-time future imagination. Following this separation, \modelname retains a trainable video expert and video-generation supervision, while using a frozen VLM for scene semantics. Its directed video--action interface allows the action expert to exploit learned predictive representations without sampling future video.

\subsection{Visual Representations for Robot Control}

Beyond the policy architecture, the choice of representation determines which aspects of visual dynamics are made available to action generation. Video Prediction Policy~\citep{hu2025videopredictionpolicy} extracts predictive visual features from a video diffusion model for control. V-JEPA~\citep{bardes2024vjepa} instead learns video representations by predicting in embedding space, and V-JEPA 2~\citep{assran2025vjepa2} extends this approach to latent planning through action-conditioned post-training. Within robot policies, VLA-JEPA~\citep{vla_jepa} uses future-state embedding prediction for pretraining, while JEPA-WAM~\citep{jepa_wam} couples spatially structured transition prediction and action generation through a shared predictor. InternVLA-A1.5~\citep{internvla_a15} distills a frozen video generator into foresight queries attached to a VLM-based policy. ST-WAM~\citep{st_wam} combines future VAE-latent and DINO-feature~\citep{caron2021emerging} prediction with semantic history retrieval, illustrating how generative and semantic prediction targets can complement each other.

Geometric supervision provides another source of action-relevant structure. Spatial Forcing~\citep{spatial_forcing} aligns intermediate VLA features with pretrained 3D representations, and LingBot-VLA 2.0~\citep{wu2026lingbotvla2} supervises current and future queries with depth and causal video features. Moving beyond per-frame geometry, 4D-WAM~\citep{wam4d} transfers trajectory-field knowledge through temporal feature-difference alignment and source-to-destination correspondence. Track4Action~\citep{track4action} predicts pooled Track4World~\citep{lu2026track4world} descriptors from policy observations and uses the resulting features to condition action generation.

\modelname combines predictive and geometric supervision within the video expert, with distinct roles for the two representations. Causal Imprint predicts clean video-latent changes and aligns with future video-expert features; its hidden states directly inform the action expert. Separately, we use Track4World~\citep{lu2026track4world} as a training-only teacher to distill clip-level geometry and motion information into the video expert. The teacher and distillation branch are discarded at inference, so this supervision does not add an extra policy inference path.

\subsection{Open-Source Systems for Robot Learning}

Open robot learning depends on reusable data interfaces, training implementations, and evaluation and deployment tools in addition to model checkpoints. LeRobot~\citep{lerobot} provides shared infrastructure for dataset handling, policy training, and robot interaction. OpenVLA~\citep{kim2024openvla} and openpi~\citep{openpi} make pretrained policies accessible through public implementations and adaptation workflows, while StarVLA~\citep{starvla} modularizes VLA development to support interchangeable components and controlled experimentation. These systems lower the cost of reproducing and extending robot policies, although their supported models, data pipelines, and deployment settings differ.

Recent foundation-model efforts also expose larger-scale training recipes. LingBot-VLA~\citep{wu2026lingbotvla} releases checkpoints and a training and evaluation codebase. Qwen-RobotManip~\citep{qwen_robotmanip} documents the data-alignment pipeline underlying its manipulation policy. OpenWAM~\citep{openwam_alpha} brings modular architectures, data processing, training, and deployment into a common framework for systematic WAM research. We share this emphasis on accessible research infrastructure. Alongside the policy, \modelname documents a unified data representation, multi-stage training, reusable frozen-encoder and teacher-feature caches, and compilation and memory optimizations. We will release these components together with the data-processing and filtering code, versioned filtered-data indices, training and evaluation code, model checkpoints, and deployment utilities, so that subsequent work can investigate model and representation choices with a concrete, inspectable training recipe.

\section{\modelname Model Design}
\label{sec:model-design}

\subsection{Architecture Overview}
\label{sec:architecture-overview}

\begin{figure}[!t]
    \centering
    \includegraphics[
        width=0.99\linewidth,
        trim=0bp 0bp 0bp 0bp,
        clip
    ]{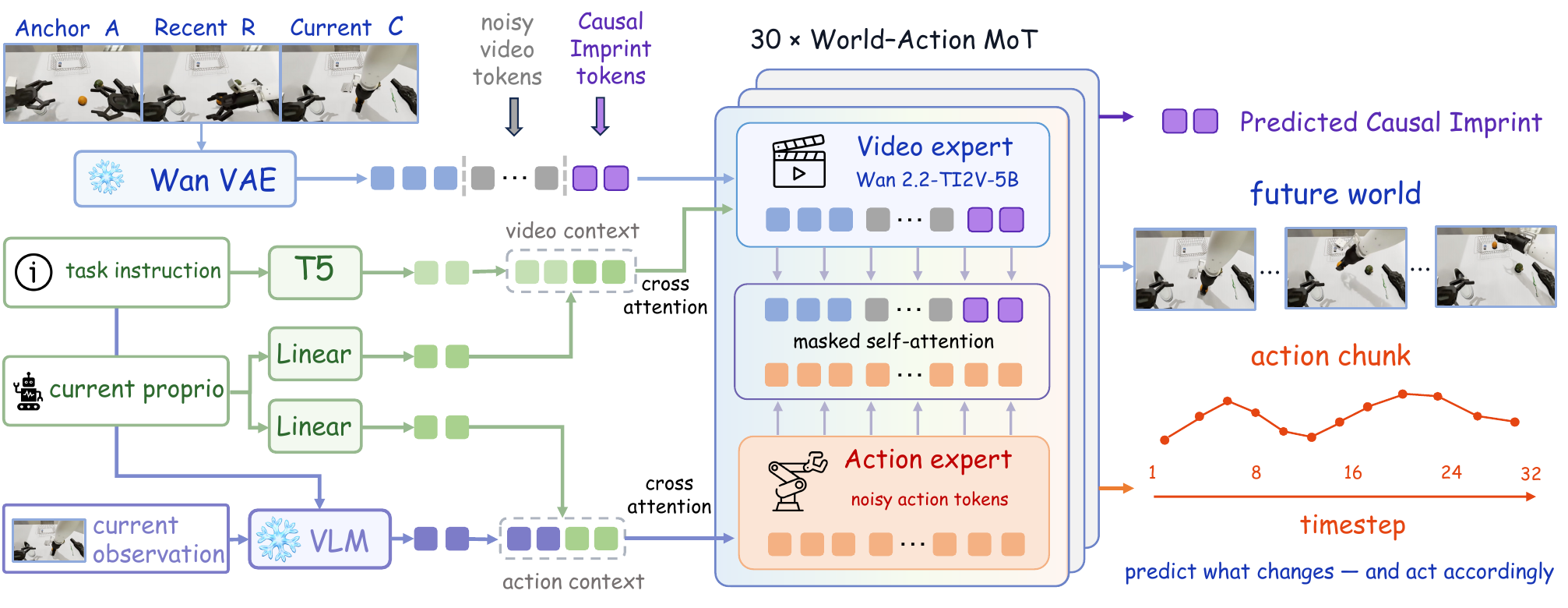}
    \caption{\textbf{Architecture overview of \modelname.} A frozen Wan VAE encodes sparse visual memory comprising anchor (A),
recent (R), and current (C) observations. The pretrained Wan2.2-TI2V-5B video expert and the ActionDiT action expert
are coupled through 30 directed Mixture-of-Transformers (MoT) blocks. T5 instruction embeddings condition the video expert, while a frozen VLM provides task-conditioned scene semantics to the action expert; current proprioception is independently projected for both experts. Causal Imprint tokens encode change-oriented predictive features from recent and current observations to guide action generation.}
    \label{fig:overview}
\end{figure}

As shown in~\Cref{fig:overview}, \modelname integrates pretrained visual dynamics, task-conditioned scene semantics, and action generation within a directed world-action architecture.
A pretrained video expert and an action expert are coupled through a directed Mixture-of-Transformers, while a frozen vision-language model provides scene-grounded task semantics to the action expert.
The video expert processes a lightweight sparse memory of anchor, recent, and current observations, with proprioceptive states conditioning both experts.
These components are detailed in~\Cref{sec:context-encoding,sec:sparse-memory,sec:directed-mot}.

To learn action-relevant dynamic representations, Causal Imprint uses training-only future supervision to capture future-relevant scene changes, while 4D-aware distillation from a frozen Track4World~\citep{lu2026track4world} teacher further introduces geometric and motion priors.
The directed information flow prevents any forward activation path from realized future observations to the action prediction path, allowing \modelname to directly generate actions at inference without future-video rollout.
We describe these representation-learning objectives and the resulting inference procedure in~\Cref{sec:causal-imprint,sec:track-distillation,sec:training-objectives,sec:efficient-inference}.

\subsection{Multimodal Context Encoding}
\label{sec:context-encoding}

\paragraph{Multi-view visual encoding.}

At each time step $\tau$, \modelname receives observations from up to $K$ camera views. We denote the $k$-th view by $o_\tau^{(k)}$ and use a binary indicator $m_\tau^{(k)}\in\{0,1\}$ to specify whether the corresponding view is available. To support heterogeneous camera configurations across embodiments, valid views are first resized and arranged into a unified visual canvas through an embodiment-dependent composition operator $\Pi$,
\begin{equation}
    x_\tau
    =
    \Pi
    \left(
        \left\{o_\tau^{(k)}\right\}_{k=1}^{K},
        \left\{m_\tau^{(k)}\right\}_{k=1}^{K}
    \right).
    \label{eq:visual-canvas}
\end{equation}
The composed observation is then encoded by the pretrained video VAE,
\begin{equation}
    z_\tau
    =
    E_{\mathrm{VAE}}(x_\tau).
    \label{eq:vae-encoding}
\end{equation}
where $z_\tau$ serves as the visual latent input to the video expert.

\paragraph{Proprioceptive encoding.}

In addition to visual observations, \modelname conditions on the current proprioceptive state $s_t$, expressed in the canonical representation defined in~\Cref{sec:unified-representation}.
The canonical state stores the current joint positions, absolute end-effector (EEF) pose, and gripper or hand configuration in fixed semantic slots.
We independently project the canonical state into the embedding spaces of the video and action experts,
\begin{equation}
    e_t^v = E_s^v(s_t),
    \qquad
    e_t^a = E_s^a(s_t),
    \label{eq:state-projections}
\end{equation}
where $E_s^v$ and $E_s^a$ are separately parameterized state encoders, each implemented as a single linear layer, and $e_t^v$ and $e_t^a$ denote the resulting proprioceptive embeddings for the video and action experts, respectively.

\paragraph{Action encoding.}

To support heterogeneous embodiments, robot-specific control signals are mapped into the canonical action representation defined in~\Cref{sec:unified-representation}, following the fixed semantic layout in~\Cref{tab:canonical-action-layout}.
Joint, gripper, and hand actions specify absolute target configurations, whereas EEF actions specify motion relative to the current EEF pose.
We denote an action chunk and its validity mask by
\begin{equation}
    A_t \in \mathbb{R}^{H_a \times D_a},
    \qquad
    M_A \in \{0,1\}^{H_a \times D_a},
    \label{eq:canonical-action}
\end{equation}
where $H_a$ denotes the action horizon, $D_a=80$ is the dimensionality of the canonical action space, and $M_A$ indicates the valid timesteps and action dimensions for the active embodiment.
The canonical action vectors are projected into the action expert's embedding space through an action encoder,
\begin{equation}
    e_t^A = E_A(A_t),
    \label{eq:action-encoding}
\end{equation}
where $E_A$ is implemented as a single linear layer and $e_t^A$ denotes the resulting action-token embeddings.

\paragraph{Task and scene semantic encoding.}

The pretrained video expert inherits the T5-based~\citep{raffel2020t5} language conditioning pathway from Wan2.2-TI2V-5B~\citep{wan2025wan}. Given the instruction $\ell$, we construct the video-side conditioning sequence as
\begin{equation}
    C_t^v
    =
    \left[
        E_{\mathrm{T5}}(\ell);
        e_t^v
    \right].
    \label{eq:video-conditioning}
\end{equation}
Retaining this pathway preserves the language conditioning learned during video pretraining. However, T5 only encodes the linguistic content of the instruction and does not directly perceive the current environment. As a result, it provides limited information about the scene in which the instruction must be executed, including the objects present in the scene, their spatial configuration, the robot--object relationships, and the current interaction state. Such scene understanding is essential for action prediction, since the same language instruction may correspond to different actions under different visual states.

We therefore introduce an additional frozen vision--language model to complement the T5 pathway with scene-level visual understanding. The VLM jointly processes all valid current views together with their view identities and the instruction,
\begin{equation}
    H_t^{\mathrm{VLM}}
    =
    f_{\mathrm{VLM}}
    \left(
        \left\{
            \left(
                o_t^{(k)},\operatorname{id}(k)
            \right)
            \,\middle|\,
            m_t^{(k)}=1
        \right\},
        \ell
    \right).
    \label{eq:vlm-encoding}
\end{equation}
The resulting multimodal tokens provide the action expert with a richer understanding of the current scene while relating this visual context to the task instruction. We combine these features with the action-side proprioceptive embedding as
\begin{equation}
    C_t^a
    =
    \left[
        H_t^{\mathrm{VLM}};
        e_t^a
    \right].
    \label{eq:action-conditioning}
\end{equation}
which is used to condition action prediction.

The two pathways are therefore complementary. T5 preserves the pretrained language prior of the video expert, while the VLM supplies the scene understanding required for observation-conditioned control. Importantly, introducing the VLM also establishes a multimodal semantic interface that can support future agentic capabilities, such as incorporating subtask descriptions, intermediate goals, persistent memory, and execution feedback~\citep{ichter2023saycan,brohan2023rt2,jiang2023vima,huang2023innermonologue}.

\subsection{Sparse Memory Context}
\label{sec:sparse-memory}
Effective action prediction requires awareness of both recent execution dynamics and the broader task context, while retaining a dense observation history introduces unnecessary computational overhead. Motivated by the observation that nearby history is most informative for short-term motion and interaction continuity, whereas distant history mainly serves as a coarse global reference, we equip \modelname with a lightweight sparse memory that preserves both levels of temporal context. Specifically, \modelname maintains a sparse visual memory consisting of an episode-level global memory, a short-term memory from the previous action chunk, and the current observation,
\begin{equation}
    \mathcal{X}_t
    =
    \left\{
        x_a,\,
        x_{t-H_c},\,
        x_t
    \right\},
    \label{eq:observation-triplet}
\end{equation}
where $x_a$ denotes the composed multi-view observation at the beginning of the episode, $x_{t-H_c}$ provides a reference to the state before the previous action chunk, and $x_t$ is the current observation. Their corresponding video latents, $\{z_a,z_{t-H_c},z_t\}$, provide sparse episode-level and recent execution context to the video expert without maintaining a dense observation history.

\begin{figure}[t]
    \centering
    \includegraphics[
        width=0.99\linewidth,
        trim=0bp 0bp 0bp 0bp,
        clip
    ]{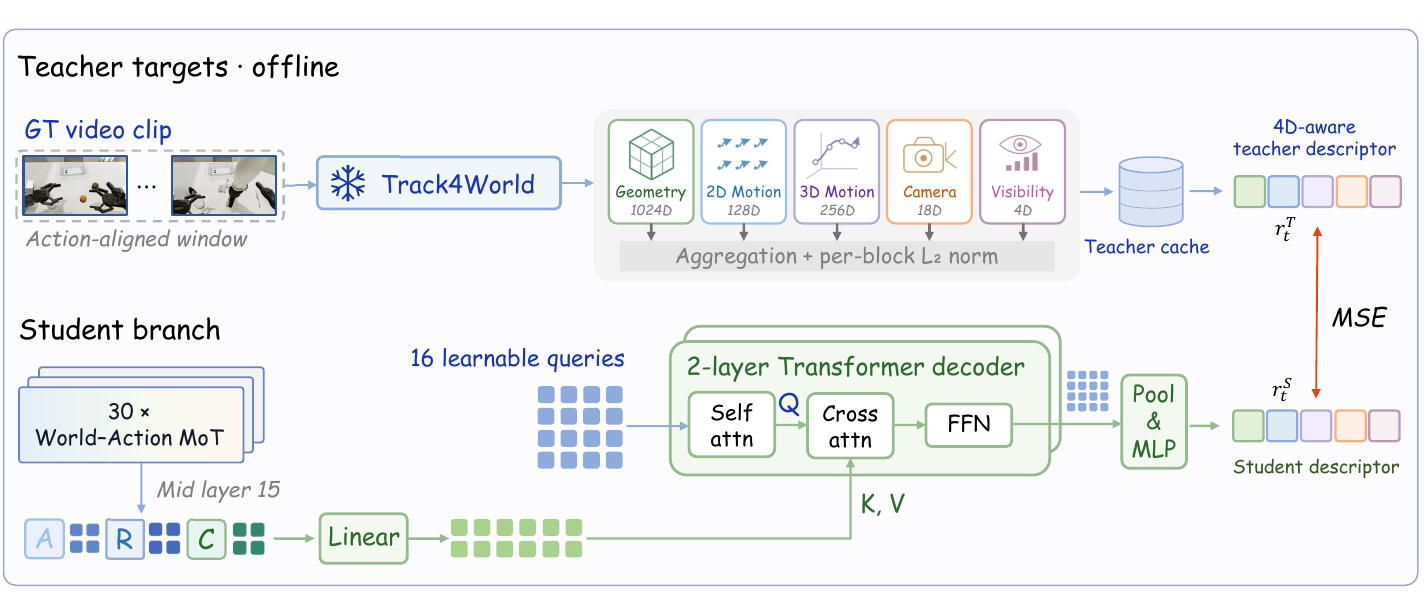}
    \vspace{-2mm}
     \caption{\textbf{4D-aware representation distillation.}
      A query-based student branch aggregates clean A/R/C features from
      the video expert and aligns them with cached Track4World descriptors
      via an auxiliary MSE loss, transferring geometric and motion priors
      without additional policy inference cost.}
    \label{fig:4d_overview}
\end{figure}

\subsection{Causal Imprint}
\label{sec:causal-imprint}

Causal Imprint is motivated by the idea that supervision from future outcomes can leave a training-time imprint on representations formed from the observed context.\footnote{The name is loosely inspired by the temporal perspective portrayed in \emph{Interstellar}, where information is depicted as leaving traces across different moments in time.}
Instead of exposing realized future observations to the action prediction path, we use them only as supervision, encouraging the model to encode future-relevant scene changes from observations available at inference time.

Concretely, we introduce a set of learnable Causal Imprint tokens into the self-attention layers of the video expert.
These tokens aggregate recent and current visual features to capture motion, interaction, and state transitions that are informative for action generation, complementing the appearance information contained in the observed visual features.
Their spatial layout follows that of the current video latent, allowing each Causal Imprint token to interact with observation tokens at the corresponding spatial resolution.

During training, we construct explicit change targets from the clean video latents before noise injection. Given the current latent $z_t$ and future latent slices $\{z_{t+i\rho_v}\}_{i=1}^{T_v}$, we define the adjacent latent differences as
\begin{equation}
    \Delta z_{t,i}
    =
    z_{t+i\rho_v}
    -
    z_{t+(i-1)\rho_v},
    \qquad
    i=1,\ldots,T_v,
    \label{eq:future-latent-delta}
\end{equation}
where $z_{t+0\rho_v}=z_t$. We stack these differences along the temporal dimension to form the supervision target $\Delta Z_t$. These adjacent latent differences encourage Causal Imprint to encode visual changes along the future trajectory. The resulting Causal Imprint representations are made available to the action expert, while the clean future latents are used only to construct training targets. The complementary future-feature alignment is illustrated in~\Cref{fig:objective}, with the corresponding objectives detailed in~\Cref{sec:training-objectives}.

\subsection{4D-Aware Representation Distillation}
\label{sec:model-3d-4d}
\label{sec:track-distillation}

To enrich the video expert with geometric and motion-aware representations, we introduce a training-only distillation objective using a frozen Track4World~\citep{lu2026track4world} teacher (See~\Cref{fig:4d_overview}).
For each training sample, the teacher processes a ground-truth video window aligned with the action horizon. 
We aggregate its geometry and 2D/3D motion features, together with camera-motion and visibility statistics, into a clip-level 4D-aware descriptor $r_t^{\mathrm{T}} \in \mathbb{R}^{1430}$ following the scheme of~\citep{track4action}.
Each component is independently $\ell_2$-normalized before concatenation. 
Teacher descriptors are precomputed and cached offline, avoiding teacher evaluation during policy training.

The student branch uses only observations available to the policy at inference. Specifically, we extract the clean-condition hidden tokens corresponding to the anchor, recent, and current frames from the 15th VideoDiT block and linearly project them into the decoder embedding space.
We then introduce 16 learnable queries and aggregate the projected visual tokens through a two-layer Transformer decoder. 
Each layer comprises query self-attention, cross-attention to the projected visual tokens, and a feed-forward network.
Finally, the decoded queries are mean-pooled and projected to the teacher feature dimension, producing the student descriptor $r_t^{\mathrm{S}}$.

We align the student descriptor with the cached teacher descriptor using an auxiliary MSE loss, jointly optimized with the original WAM objectives, to distill geometric and motion priors into the video expert. 
This distillation introduces no additional policy inference cost, as neither the teacher nor the student branch is required at inference.

\subsection{Mixture-of-Transformers and Attention Mask}
\label{sec:directed-mot}

\begin{figure}[t]
    \centering
    \includegraphics[
        width=0.99\linewidth,
        trim=0bp 0bp 0bp 0bp,
        clip
    ]{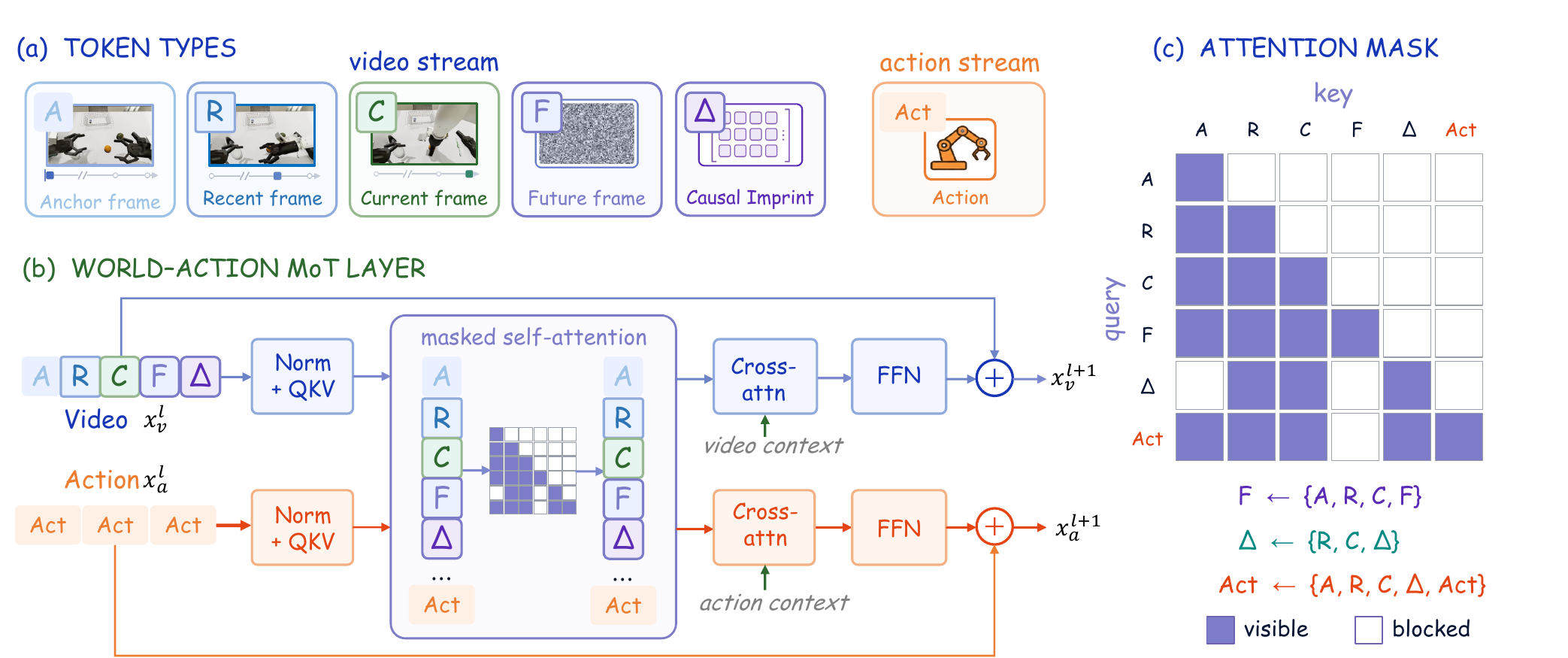}
    \caption{\textbf{World-Action MoT layer and attention mask.} (a) The video stream contains anchor (A), recent (R), current (C),
future (F), and Causal Imprint ($\Delta$) tokens, while action tokens
(Act) form a separate stream.
(b) The two experts compute separate query, key, and value projections
for joint masked self-attention, followed by expert-specific
cross-attention, feed-forward, and residual pathways.
(c) Rows and columns index query and key token groups, respectively.}
    \label{fig:am}
\end{figure}

The video and action experts retain separate modality-specific parameters, while exchanging information through masked joint attention. The token groups and a coupled World-Action MoT layer are illustrated
in~\Cref{fig:am}(a,b). At each coupled transformer block, the two experts independently compute their query, key,
and value projections, which are concatenated for a single attention operation:
\begin{equation}
    Y
    =
    \operatorname{Attn}
    \left(
        \begin{bmatrix}Q^v\\Q^a\end{bmatrix},
        \begin{bmatrix}K^v\\K^a\end{bmatrix},
        \begin{bmatrix}V^v\\V^a\end{bmatrix};
        M
    \right),
    \qquad
    (Y^v,Y^a)=\operatorname{Split}(Y).
    \label{eq:joint-attention}
\end{equation}
The resulting features are then routed back to their corresponding streams, where the video and action experts continue with their own cross-attention, feed-forward, and residual pathways. Thus, joint attention serves as the
communication interface between the two experts without merging their modality-specific parameters. The cross-attention modules use the video-side context $C_t^v$
and action-side context $C_t^a$ defined
in~\Cref{eq:video-conditioning,eq:action-conditioning}, respectively. After the final MoT block, the action decoder uses a single linear layer to map the action expert's output features to $D_a$-dimensional flow-velocity predictions in the canonical action space.

As illustrated in~\Cref{fig:am}, the attention mask further controls the direction of information flow among the anchor, recent, current, future, Causal Imprint, and action tokens. Future-frame tokens can use the observed context for future prediction, while Causal Imprint and action tokens are prevented from directly accessing the realized future. Instead, the Causal Imprint tokens learn predictive change representations from the recent and current observations under future-derived supervision, and expose these representations to the action stream. The action expert therefore conditions on observed context and the learned Causal Imprint representation, rather than privileged future frames.

This directed information flow separates future supervision from policy inference. Future observations provide a training signal to the video expert, but no forward activation path exposes the realized future to the action policy at inference time.

\subsection{Training Objectives}
\label{sec:training-objectives}

We jointly optimize \modelname for video generation, action generation, Causal Imprint learning, and 4D-aware representation distillation.

\paragraph{Flow matching objectives.}
Following the flow-matching formulation of the pretrained video model, for a target variable $y$, we sample Gaussian noise $\epsilon\sim\mathcal{N}(0,I)$ and a flow time $\sigma\in(0,1)$, and construct
\begin{equation}
    y^{\sigma}
    =
    (1-\sigma)y+\sigma\epsilon.
\end{equation}
The corresponding flow-matching objective is
\begin{equation}
    \mathcal{L}_{\mathrm{FM}}(y)
    =
    \mathbb{E}_{\epsilon,\sigma}
    \left[
        \left\|
            f_{\theta}(y^{\sigma},\sigma)
            -
            (\epsilon-y)
        \right\|_2^2
    \right].
    \label{eq:flow-matching}
\end{equation}

We apply this objective to both future video generation and action prediction:
\begin{equation}
    \mathcal{L}_{\mathrm{video}}
    =
    \mathcal{L}_{\mathrm{FM}}(Z_t^F),
    \qquad
    \mathcal{L}_{\mathrm{action}}
    =
    \mathcal{L}_{\mathrm{FM}}(A_t),
    \label{eq:video-action-loss}
\end{equation}
where $Z_t^F$ denotes the future video latent sequence and $A_t$ denotes the target action chunk.

\begin{figure}[t] 
    \centering
    \includegraphics[
        width=0.99\linewidth,
        trim=0bp 0bp 0bp 0bp,
        clip
    ]{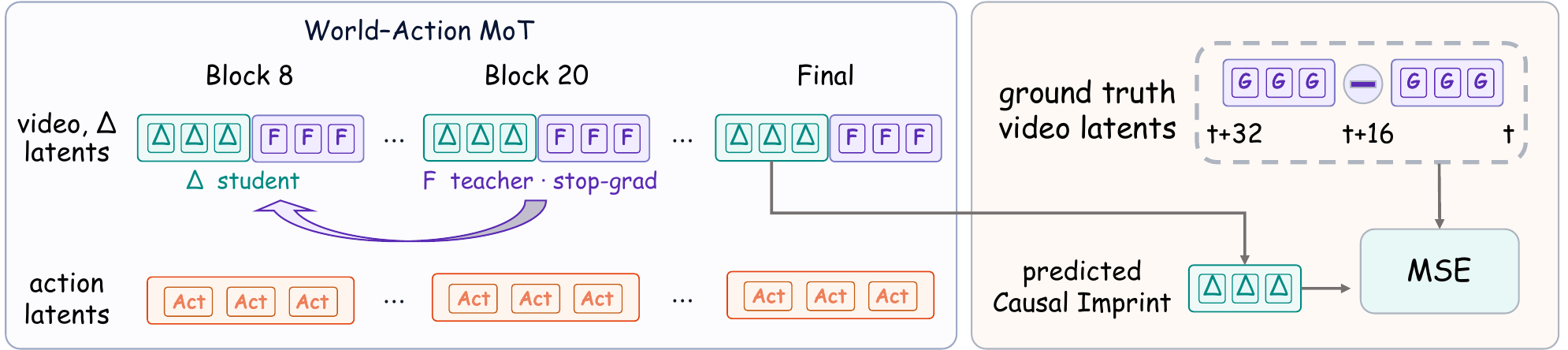}
    \caption{\textbf{Causal Imprint supervision.} Causal Imprint is trained with two complementary objectives. For future-feature alignment (left), Causal Imprint features at block 8 are aligned with spatially corresponding features of the terminal future slice at block 20 using the cosine alignment loss. For direct Causal Imprint supervision (right), the final Causal Imprint features regress adjacent differences between clean ground-truth video latents using a squared-error loss.}
    \label{fig:objective}
\end{figure}

\paragraph{Causal Imprint objectives.}
We train Causal Imprint with the two complementary objectives illustrated in~\Cref{fig:objective}. First, the final-layer Causal Imprint representation $h_t^\Delta$ is directly supervised by the adjacent clean-latent differences $\Delta Z_t$ defined in~\Cref{eq:future-latent-delta}. The corresponding objective is
\begin{equation}
    \mathcal{L}_{\Delta}
    =
    \left\|
        h_t^\Delta-\Delta Z_t
    \right\|_2^2.
    \label{eq:future-delta-loss}
\end{equation}

This objective provides direct supervision for Causal Imprint, encouraging it to encode changes associated with the subsequent evolution of the scene. In addition, video diffusion transformers have been shown to develop rich semantic, temporal, and physical representations in their intermediate features~\citep{king2026gen4u,esmati2026invisible}. We therefore introduce a representation-alignment objective to transfer such predictive information from the video expert to Causal Imprint. Specifically, each Causal Imprint token at block $\ell_s=8$ is aligned with the spatially corresponding feature of the terminal future slice at block $\ell_t=20$:
\begin{equation}
    \mathcal{L}_{\mathrm{align}}
    =
    1-
    \operatorname{cos}
    \left(
        h_{b,p}^{\Delta,\ell_s},
        \operatorname{sg}
        \left[
            h_{b,p}^{F,\ell_t}
        \right]
    \right),
    \label{eq:future-feature-loss}
\end{equation}
where $\operatorname{sg}[\cdot]$ denotes stop-gradient. The future-video features serve only as detached training targets and are not connected to Causal Imprint through the forward attention path.

\paragraph{4D-aware representation distillation objective.}
We align the student descriptor $r_t^{\mathrm{S}}$ with the cached teacher 4D-aware descriptor $r_t^{\mathrm{T}}$ using an auxiliary mean squared error loss:
\begin{equation}
  \mathcal{L}_{\mathrm{4D}}
  =
  \frac{1}{D}
  \left\|
      r_t^{\mathrm{S}}
      -
      \operatorname{sg}\!\left[r_t^{\mathrm{T}}\right]
  \right\|_2^2,
  \label{eq:track-distillation-loss}
\end{equation}
where $D=1430$ is the descriptor dimension and $\operatorname{sg}[\cdot]$ denotes stop-gradient.
The loss is averaged over training samples with valid teacher descriptors. This auxiliary supervision transfers geometric and motion priors to the video expert by backpropagating
through the student branch and the selected VideoDiT hidden tokens, while the teacher remains frozen.

\begin{figure}[!t]
    \centering
    \includegraphics[
        width=0.99\linewidth,
        trim=0bp 0bp 0bp 0bp,
        clip
    ]{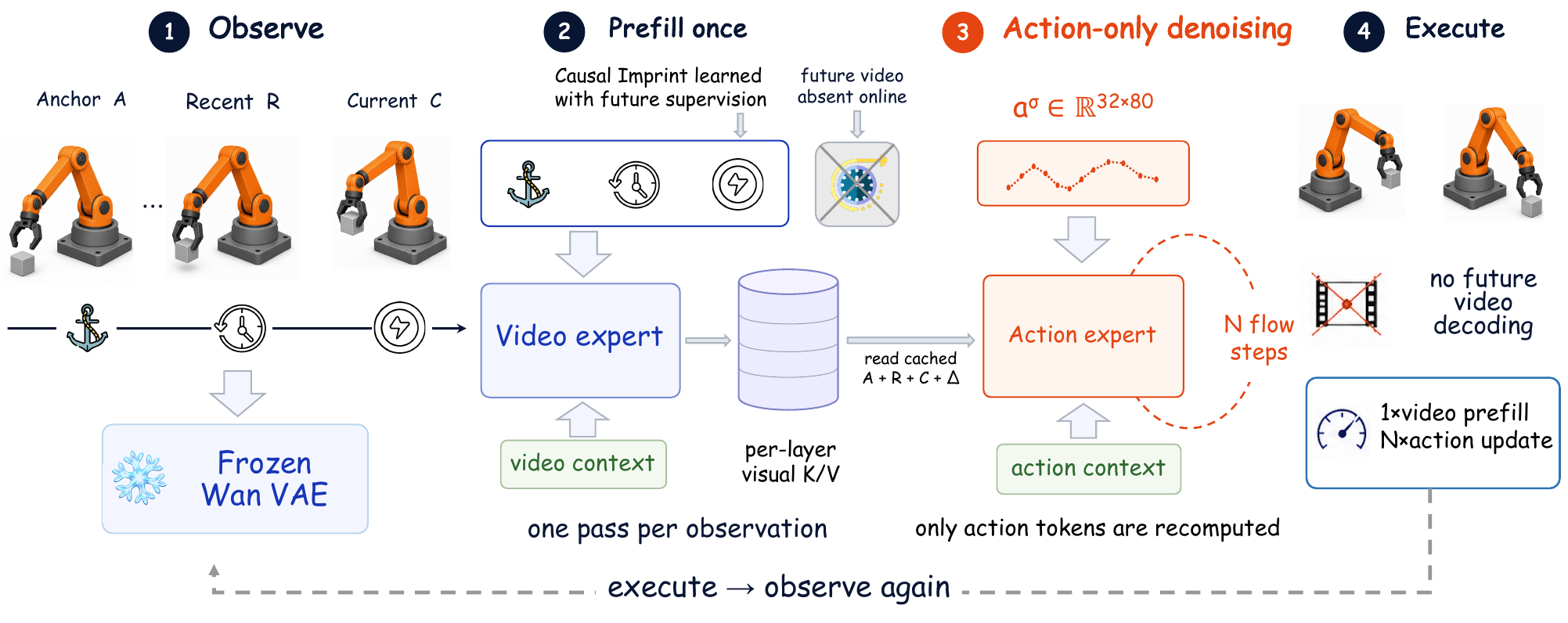}
    \caption{\textbf{Efficient inference with cached visual context.} (1) Anchor (A), recent (R), and current (C) observations are encoded by the frozen Wan VAE. (2) Conditioned on the video context, the video expert processes the observation tokens and Causal Imprint tokens once, caching their key/value (K/V) projections at each MoT layer. Future-video tokens are omitted. (3) Starting from Gaussian noise, the action expert generates an action chunk through $N$ flow steps, reusing the cached A/R/C/$\Delta$ features and the action-side context. Only the action stream is recomputed during denoising. (4) The predicted actions are executed, and the process repeats with an updated cache when a new observation is acquired. Each observation-to-action cycle requires one video-expert prefill and $N$ action-expert updates, without future-video sampling or decoding.}
    \label{fig:efficient-inference}
\end{figure}

\paragraph{Overall objective.}
The complete training objective is
\begin{equation}
    \mathcal{L}
    =
    \lambda_v\mathcal{L}_{\mathrm{video}}
    +
    \lambda_a\mathcal{L}_{\mathrm{action}}
    +
    \lambda_{\Delta}\mathcal{L}_{\Delta}
    +
    \lambda_{\mathrm{align}}\mathcal{L}_{\mathrm{align}}
    +
    \lambda_{\mathrm{4D}}\mathcal{L}_{\mathrm{4D}}.
    \label{eq:training-total-objective}
\end{equation}

\subsection{Efficient Inference}
\label{sec:efficient-inference}

As illustrated in~\Cref{fig:efficient-inference}, \modelname
generates action chunks without sampling future video.
Inference consists of a single video-expert prefill followed by iterative action-only denoising.
The visual context is computed once for each new observation and reused throughout the action denoising process.

\paragraph{Visual prefill and caching.}
At each policy step, we encode the anchor, recent, and current observations defined in~\Cref{eq:observation-triplet} using the frozen Wan VAE. The resulting observation tokens, together with the learned Causal Imprint tokens, are processed by the video expert conditioned on $C_t^v$, without instantiating future-video tokens. We cache the self-attention key/value (K/V) projections of the A/R/C/$\Delta$ tokens at each MoT layer:
\begin{equation}
    \mathcal{C}_t^{\mathrm{KV}}
    =
    \left\{
        \left(K_t^{v,l}, V_t^{v,l}\right)
    \right\}_{l=1}^{L},
    \label{eq:visual-kv-cache}
\end{equation}
where $L$ is the number of coupled MoT layers.
The directed mask in~\Cref{fig:am}(c) prevents
observation and Causal Imprint tokens from attending to action
tokens.
With the video-side inputs and conditioning held fixed during
action denoising, these cached features are independent of
the evolving action tokens and can be reused at every flow step.

\paragraph{Action-only denoising and execution.}
We initialize a noisy action chunk
$A_t^1 \sim \mathcal{N}(0,I)$ with shape
$H_a \times D_a$, where $H_a=32$ and $D_a=80$.
At each flow step, the action expert recomputes only the action
stream, attending to the cached visual K/V and the action tokens,
while using $C_t^a$ through its cross-attention pathway.
The predicted action flow velocity is
\begin{equation}
    v_t^\sigma
    =
    f_\theta^a
    \left(
        A_t^\sigma,\sigma;
        \mathcal{C}_t^{\mathrm{KV}},C_t^a
    \right),
    \label{eq:cached-action-flow}
\end{equation}
where $f_\theta^a$ denotes the action-side flow predictor.
Following the flow convention
in~\Cref{eq:flow-matching}, we perform $N$ integration steps
from $\sigma=1$ to $\sigma=0$ to obtain the predicted action
chunk $\hat{A}_t=A_t^0$.
No video-side token features are recomputed within this loop,
and neither future-video sampling nor VAE decoding is required.

The predicted action chunk is used for robot execution.
When a new observation is acquired, the sparse memory and
conditioning are updated, and the visual cache is recomputed
for the next cycle.
Each cycle therefore consists of one video-expert prefill
and $N$ action-expert updates.

\section{Data}
In this section, we introduce the data recipe of \modelname, which unifies heterogeneous embodied data under a common representation for joint training. We first describe the unified representation used across different data sources. We then describe the data sources, processing rules, and filtering statistics for robot manipulation data, egocentric and ego-to-robot data, and Universal Manipulation Interface (UMI) data. 

\subsection{Unified Representation}
\label{sec:unified-representation}
We train \modelname to use heterogeneous data sources that span diverse robot morphologies, degrees of freedom, control interfaces, coordinate conventions, and action dimensions. Directly mixing their native representations would assign inconsistent physical meanings to the same dimensions, hindering cross-embodiment learning. We therefore convert all trajectories into a canonical state--action representation with fixed semantic slots. In particular, all robot actions are mapped into a shared 80-dimensional action space, whose detailed layout is provided in \Cref{tab:canonical-action-layout}. Each slot is associated with a predefined physical quantity rather than the original ordering used by the source dataset.

For the robot state, we retain absolute quantities that describe the current robot configuration. These include the current joint positions, the absolute end-effector (EEF) pose, and the current gripper or hand configuration. The EEF pose is represented by a 3D position together with a continuous 6D rotation representation. For parallel-jaw grippers, the hand state is represented by the gripper opening, while dexterous hands use the corresponding hand joint configuration. All EEF poses are first transformed into a consistent coordinate convention before being assigned to the canonical slots.

For the action representation, we distinguish between joint-space and task-space control signals. Joint actions are represented as absolute target joint positions, corresponding to the configuration that the low-level controller is expected to reach at the next or a future control step. EEF actions consist of a 3D translation increment and a 3D rotation vector. The latter parameterizes the relative orientation change in the local coordinate frame of the current EEF. Thus, each EEF uses a 9D absolute pose representation in the state and a 6D relative motion representation in the action, before being embedded into the canonical format. An action segment contains $H_a$ consecutive canonical action vectors, each following the same fixed semantic layout.

\begin{table}[t]
  \centering
  \caption{Canonical layout of the 80-dimensional robot action
  representation. Slot ranges use zero-based, half-open indexing.
  Paired arm ranges are listed in left--right order.}
  \label{tab:canonical-action-layout}
  \small
  \setlength{\tabcolsep}{6pt}
  \renewcommand{\arraystretch}{1.1}
  \begin{tabular}{@{}lcc@{}}
    \toprule
    Component & Slot range(s) & Dim. \\
    \midrule

    \multicolumn{3}{@{}l}{\textit{Arm-specific slots}} \\
    \addlinespace[2pt]
    Arm joints
      & $[0,7),\ [40,47)$
      & $2 \times 7$ \\
    End-effector
      & $[7,16),\ [47,56)$
      & $2 \times 9$ \\
    Gripper
      & $[16,17),\ [56,57)$
      & $2 \times 1$ \\
    Hand
      & $[17,29),\ [57,69)$
      & $2 \times 12$ \\
    \midrule

    \multicolumn{3}{@{}l}{\textit{Body and mobility slots}} \\
    \addlinespace[2pt]
    Torso joints
      & $[29,34)$
      & $5$ \\
    Independent lift
      & $[39,40)$
      & $1$ \\
    Head
      & $[74,77)$
      & $3$ \\
    Mobile base
      & $[77,80)$
      & $3$ \\
    \midrule

    Reserved slots
      & $[34,39),\ [69,74)$
      & $2 \times 5$ \\
    \midrule

    \textbf{Total}
      & $[0,80)$
      & $\mathbf{80}$ \\
    \bottomrule
  \end{tabular}
\end{table}

\subsection{Robot Data}

\subsubsection{Data Sources}
Robot manipulation demonstrations serve as the main source of embodied supervision in our training corpus. We integrate a total of 15 datasets collected from both simulated and real-world environments, covering a wide range of manipulation scenarios, including single-arm and dual-arm tabletop tasks, dexterous manipulation, mobile manipulation, and humanoid interaction. The resulting collection spans diverse robot embodiments, task settings, sensing configurations, and control interfaces, providing broad coverage of manipulation behaviors for model training.

\paragraph{AgiBotWorld~\citep{contributors2024agibotworldrepo}}
AgiBotWorld is a large-scale real-world manipulation dataset collected using a fleet of dual-arm humanoid robots. Its full release contains over one million trajectories and approximately 3,000 hours of demonstrations, covering 217 tasks, 87 manipulation skills, more than 3,000 objects, and over 100 real-world scenes. The data includes dual-arm manipulation, dexterous-hand interaction, tool use, and mobile manipulation, with multimodal observations from multiple cameras and tactile sensors.

\paragraph{InternData-A1~\citep{contributors2025internroboticsrepo}}
InternData-A1 provides large-scale manipulation data across both simulated and real-world settings. It contains more than 630K trajectories and 7,400 hours of interaction across four robot embodiments, 70 tasks, and 227 scenes, covering rigid, articulated, deformable, and fluid-object manipulation. It further includes long-horizon tasks, multi-arm collaboration, and human--robot interaction, providing complementary supervision beyond standard tabletop manipulation.

\paragraph{RoboMIND~\citep{wu2025robomind} and RoboMIND~2.0~\citep{hou2025robomind20multimodalbimanual}}
RoboMIND contains 107K human-teleoperated trajectories across 479 tasks and four robot embodiments, ranging from single-arm manipulators to dual-arm and dexterous humanoid platforms. It also provides failure demonstrations in addition to successful trajectories. RoboMIND~2.0 further extends this collection to more than 310K real-world dual-arm trajectories across six embodiments and 739 tasks, together with tactile-enhanced and mobile-manipulation episodes, substantially increasing the coverage of contact-rich and spatially extended behaviors.

\paragraph{RW-RL Dataset~\citep{rw_rl_dataset_2026}}
The RW-RL Dataset focuses on real-world interaction data for policy improvement beyond offline imitation learning. It contains over 1,000 hours of interaction across multiple robot families, scenario domains, and task templates, combining teleoperated demonstrations with autonomous rollouts, human interventions, reward signals, and termination labels. These data expose the model to both successful behaviors and states encountered during policy execution and recovery.

\paragraph{Dexora~\citep{zhang2026dexoraopensourcevlahighdof}}
Dexora targets high-DoF bimanual dexterous manipulation with dual robot arms and dual dexterous hands. It combines 100K embodiment-matched simulated trajectories with 10K real-world teleoperated episodes, covering both coarse arm motion and fine-grained finger control. This data provides dense supervision for manipulation behaviors that require coordinated arm and finger motion beyond parallel-jaw grippers.

\paragraph{ABC-130K~\citep{abc2026}}
ABC-130K contains more than 130K real-world bimanual teleoperation episodes, corresponding to over 3,500 hours of interaction across 195 manipulation tasks. Collected with dual-arm YAM platforms, it covers manipulation primitives including pick-and-place, folding, handover, insertion, tool use, and assembly. Its scale and task composition provide extensive supervision for coordinated bimanual manipulation.

\paragraph{Galaxea Open-World Dataset~\citep{galaxea2025}}
The Galaxea Open-World Dataset contains more than 500 hours of real-world mobile manipulation collected with a consistent robot embodiment. The demonstrations span residential, kitchen, retail, and office environments and include fine-grained subtask-level language annotations. Compared with fixed tabletop data, it provides additional supervision for spatially extended and long-horizon manipulation that couples navigation with object interaction.

\paragraph{RoboCOIN~\citep{RoboCOINReport}}
RoboCOIN is a multi-embodiment bimanual manipulation dataset containing more than 180K demonstrations collected across 15 robot platforms. It covers 421 tasks in 16 real-world scenarios and organizes bimanual behaviors according to coordination patterns and object properties. Its hierarchical annotations provide supervision at trajectory, subtask, and frame levels, while its embodiment diversity complements datasets collected using a single robot platform.

\paragraph{RH20T~\citep{fang2024rh20t}}
RH20T contains more than 110K contact-rich real-world manipulation sequences covering 147 tasks and seven robot configurations. In addition to RGB observations and robot proprioception, it records force, audio, depth, and, for part of the data, tactile information. Each robot trajectory is also associated with a human demonstration and language description, providing rich multimodal supervision for contact-sensitive manipulation.

\paragraph{RDT-1B~\citep{liu2024rdt}}
The robot-data corpus used by RDT-1B aggregates 46 datasets into more than one million episodes spanning multiple robot embodiments and manipulation settings. It is further complemented by more than 6K demonstrations collected on the ALOHA dual-arm platform. We incorporate this data to increase the diversity of robot morphologies and to strengthen the coverage of coordinated bimanual behaviors.

\paragraph{RoboSet~\citep{RoboHive}}
RoboSet combines real-world teleoperation data with simulated human and expert-policy trajectories under a common data format. Its main real-world component contains approximately 31K kitchen manipulation trajectories across 40 tasks, accompanied by multi-view visual observations, actions, robot states, and rewards. The dataset therefore contributes both large-scale real-world tabletop interactions and additional simulated manipulation behaviors.

\paragraph{RealSource-World~\citep{realsourceworld}}
RealSource-World contains more than 14 million frames from over 11K real-world dual-arm manipulation episodes collected using the RS-02 humanoid robot. The dataset covers 35 tasks across household, kitchen, office, retail, and industrial environments, with synchronized head and wrist cameras and rich robot proprioception. It additionally provides atomic-skill segmentation and trajectory-quality annotations, making long-horizon demonstrations accessible at finer temporal granularity.

\paragraph{MolmoAct2-BimanualYAM~\citep{fang2026molmoact2actionreasoningmodels}}
MolmoAct2-BimanualYAM provides more than 720 hours of real-world bimanual manipulation demonstrations collected using dual YAM arms. The dataset contains multi-view observations and language-annotated tasks covering behaviors such as garment manipulation, cable handling, grocery organization, packing, and table bussing. These demonstrations add substantial coverage of long-horizon bimanual tasks involving coordinated interactions between both arms.

\paragraph{HABIT~\citep{song2026habit}}
HABIT focuses on robot manipulation in environments where humans are actively present. It contains more than 10K episodes and 160 hours of demonstrations across 60 tasks, organized around collaborative, shared-space, and human-directed interactions. This setting introduces behaviors such as human--robot synchronization, yielding, and gesture-conditioned manipulation that are largely absent from robot-only demonstration datasets.

\paragraph{ActionNet~\citep{fourier2025actionnet}}
ActionNet focuses on dexterous bimanual manipulation using humanoid robots. It contains more than 30K teleoperated trajectories, corresponding to approximately 140 hours of interaction, and covers both basic manipulation primitives and more complex two-hand behaviors using dexterous hands. The demonstrations are collected across multiple Fourier humanoid platforms and paired with manually verified language instructions, providing additional supervision for high-DoF humanoid manipulation.

\begin{table*}[t]
  \centering
  \caption{Dataset-level statistics before and after filtering.
  FPS denotes the frame rate declared in the training data metadata;
  comma-separated values indicate subsets with different frame rates.}
  \label{tab:pretraining-data-statistics}
  \small
  \setlength{\tabcolsep}{6pt}
  \renewcommand{\arraystretch}{1.08}
  \sisetup{
      group-separator = {,},
      group-minimum-digits = 4,
      table-number-alignment = center,
      detect-weight = true,
      mode = text
  }

  \begin{adjustbox}{max width=\linewidth}
\begin{tabular}{@{}lc
      S[table-format=5.2]
      S[table-format=7.0]
      S[table-format=5.2]
      S[table-format=7.0]
      S[table-format=4.3]@{}}
  \toprule
  Dataset
  & FPS
  & \multicolumn{2}{c}{Before filtering}
  & \multicolumn{3}{c}{After filtering} \\
  \cmidrule(lr){3-4}
  \cmidrule(lr){5-7}
  & & {Hours}
  & {Episodes}
  & {Hours}
  & {Episodes}
  & {Frames (M)} \\
  \midrule

  \multicolumn{7}{@{}l}{\textit{Robot data}} \\
  \addlinespace[2pt]

  InternData-A1~\citep{contributors2025internroboticsrepo}
  & 30 & 3528.27 & 573389 & 3375.51 & 559062 & 364.555 \\

  AgiBotWorld~\citep{contributors2024agibotworldrepo}
  & 30 & 2358.63 & 141563 & 1698.21 & 105227 & 183.407 \\

  Galaxea~\citep{galaxea2025}
  & 15, 62 & 681.22 & 29994 & 327.42 & 17703 & 17.684 \\

  RoboMIND~\citep{wu2025robomind}
  & 30 & 777.01 & 215240 & 625.79 & 195869 & 67.586 \\

  ABC-130K~\citep{abc2026}
  & 30 & 3533.30 & 128996 & 2587.94 & 94727 & 279.497 \\

  RoboSet~\citep{RoboHive}
  & $\approx 5.12$ & 21.64 & 9500 & 20.05 & 9382 & 0.370 \\

  MolmoAct2~\citep{fang2026molmoact2actionreasoningmodels}
  & 30 & 68.68 & 3331 & 33.28 & 1927 & 3.594 \\

  RoboCOIN~\citep{RoboCOINReport}
  & 30, 50 & 533.17 & 77843 & 485.77 & 77102 & 53.095 \\

  RDT-1B~\citep{liu2024rdt}
  & 25 & 35.66 & 6110 & 33.10 & 6110 & 2.979 \\

  RH20T~\citep{fang2024rh20t}
  & 10 & 1132.10 & 82894 & 1130.92 & 82823 & 40.713 \\

  ActionNet~\citep{fourier2025actionnet}
  & 30 & 157.72 & 32121 & 157.22 & 32120 & 16.980 \\


  Dexora~\citep{zhang2026dexoraopensourcevlahighdof}
  & 20 & 26.30 & 7867 & 26.30 & 7867 & 1.894 \\

  HABIT~\citep{song2026habit}
  & 10 & 165.18 & 10623 & 122.71 & 8389 & 4.417 \\

  RealSource~\citep{realsourceworld}
  & 30 & 345.46 & 26671 & 177.92 & 25504 & 19.216 \\

  RWRL~\citep{rw_rl_dataset_2026}
  & 15, 30 & 503.37 & 23861 & 500.06 & 23844 & 45.060 \\

  \addlinespace[2pt]
  \textbf{Total}
  & --
  & \bfseries 13867.71
  & \bfseries 1370003
  & \bfseries 11302.20
  & \bfseries 1247656
  & \bfseries 1101.047 \\

  \midrule
  \multicolumn{7}{@{}l}{\textit{UMI data}} \\
  \addlinespace[2pt]

  Hy-UMI-10K~\citep{zhang2026hy}
  & 30 & 2161.74 & 250135 & 2075.42 & 416053 & 224.146 \\

  \midrule
  \multicolumn{7}{@{}l}{\textit{Egocentric human data}} \\
  \addlinespace[2pt]

  EgoDex~\citep{hoque2025egodex}
  & 30 & 821.48 & 333682 & 772.16 & 310418 & 83.393 \\

  EgoVerse~\citep{punamiya2026egoverse}
  & 30 & 3818.56 & 1484714 & 3289.19 & 1312338 & 355.233 \\

  \addlinespace[2pt]
  \textbf{Total}
  & --
  & \bfseries 4640.04
  & \bfseries 1818396
  & \bfseries 4061.35
  & \bfseries 1622756
  & \bfseries 438.626 \\

  \bottomrule
  \end{tabular}
\end{adjustbox}%
  \end{table*}

\subsubsection{Processing Rules}
\label{sec:processing_rules}

Our data curation builds on practices used in prior work,
including Qwen-RobotManip~\citep{qwen_robotmanip}.
We apply signal anomaly and consistency filtering, static boundary
trimming, visual quality filtering, and an action magnitude guard
to the robot demonstrations.
We additionally perform automated checks of instruction correctness
and video--instruction consistency.
 The resulting data are further checked through manual inspection
of sampled episodes to verify robot motion and representation
conventions across datasets and embodiments.
We release the complete filtering pipeline and its implementation
as an open-source resource to support reproducible data curation
and facilitate community reuse and adaptation to additional
datasets and embodiments.

\paragraph{Signal anomaly and consistency filtering.}
Following Qwen-RobotManip~\citep{qwen_robotmanip}, we perform
sudden-change detection and state--action consistency checks.
Frames containing non-finite values or abrupt outliers, identified
through deviations from smoothed trajectories and higher-order
temporal changes, are removed while contiguous valid segments are
retained. For physically comparable state--action dimensions with
sufficient variation, we assess directional agreement after local
cross-correlation alignment. We use a default agreement threshold
of $0.65$ and reject episodes if any required dimension falls below
this threshold or if state changes systematically precede the
associated actions. Comparisons involving delta actions are
adapted to the corresponding control representation.

\paragraph{Static boundary trimming.}
We apply an adaptive trimming rule to remove prolonged inactivity
at episode boundaries. Motion is estimated from normalized
frame-to-frame state differences, with rotations represented
continuously over time to avoid artificial discontinuities.
An adaptive threshold derived from each episode's motion
distribution identifies sustained active segments. We retain
the interval spanning the first and last reliable active segments,
together with a small context margin at both ends. Intermediate
pauses are preserved to maintain the temporal structure of task
execution. Short episodes and those without a clear separation
between static and active motion are left unchanged.

\paragraph{Visual quality filtering.}
We perform frame-level visual quality checks to detect black frames, blur,
compression artifacts, and other image-level anomalies, using frames
resized to a unified resolution for all subsequent measurements. Black
frames are identified from the mean luminance and the fraction of pixels
below a dark-intensity threshold. For blur detection, Sobel gradients are
first used to locate edge pixels, and the mean absolute Laplacian response
is then computed only over sufficiently populated edge regions to avoid
false positives in textureless manipulation scenes. Compression artifacts
are quantified by comparing neighboring-pixel differences at periodic
$8\times8$ block boundaries with those at non-boundary locations. Abrupt
temporal changes are measured using frame differences normalized by the
episode-level median and median absolute deviation. Extreme temporal
changes, blockiness, and saturation anomalies are combined to identify
suspected corruption, while decoding and shape errors are recorded
separately.

\begin{figure*}[t]
    \centering
    \includegraphics[width=0.9925\linewidth]{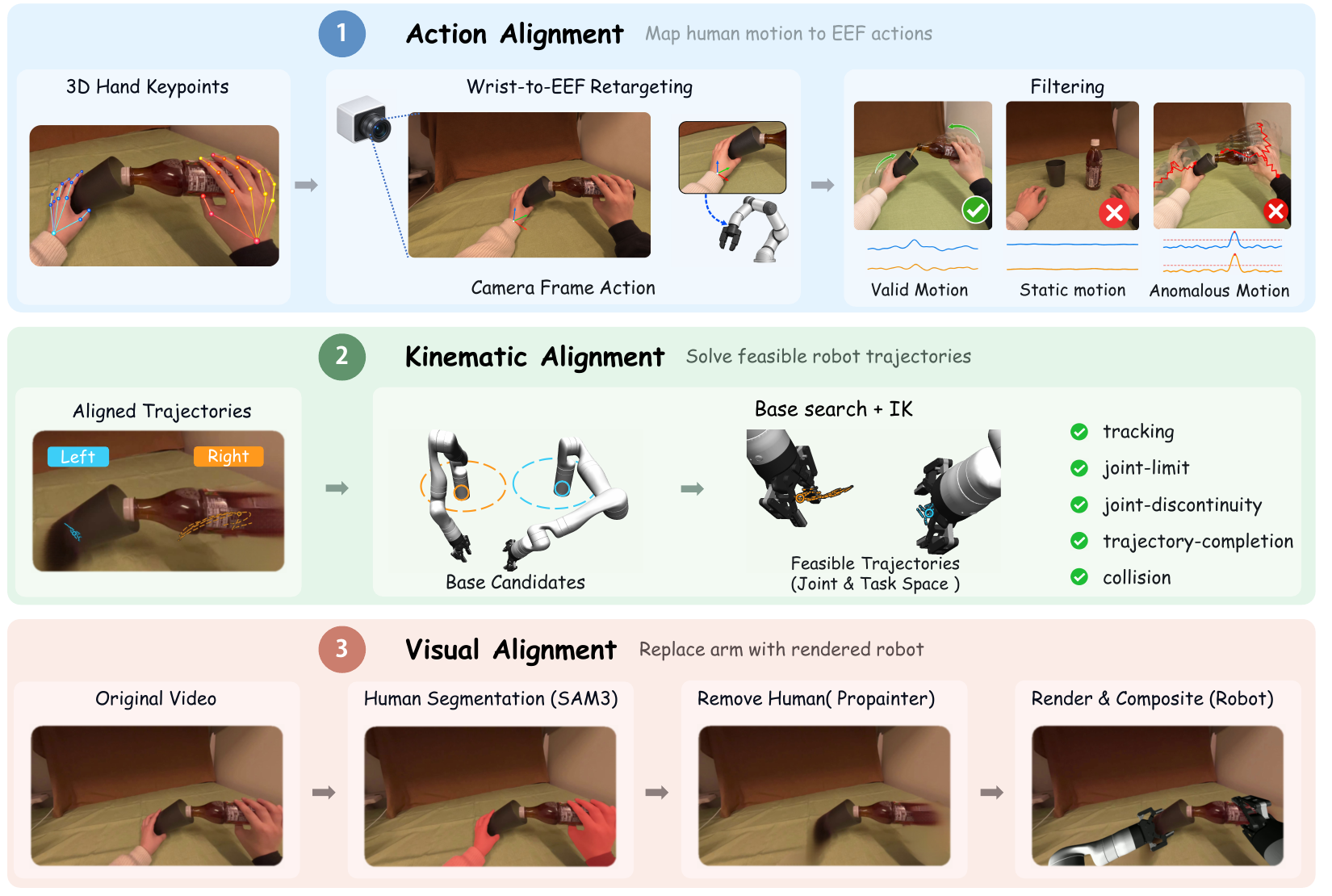}
    \caption{Ego and ego-to-robot data processing pipeline. Ego2Robot data undergo all
stages, whereas egocentric data undergo only
action alignment and action speed alignment. Action alignment maps
    hand motion to end-effector targets and gripper commands. Kinematic alignment
    couples base search with trajectory IK to obtain base poses and joint
    trajectories. Visual alignment combines the rendered robot with the
    human-removed scene using depth-aware compositing. }
    \label{fig:human2robot-pipeline}
\end{figure*}

\paragraph{Action magnitude guard.}
After the preceding signal and visual quality checks, we apply
an action magnitude guard to remove residual large-amplitude
commands. In some datasets, initialization and reset routines near episode boundaries introduce
sustained large actions, such as rapid arm retraction. These motions
can remain temporally smooth and consistent with the recorded
states, allowing them to survive the preceding filters. When
present over consecutive frames, they can also distort action
statistics despite normalization based on the 1st and 99th
percentiles ($q_{01}$ and $q_{99}$). We therefore impose additional
magnitude limits in physical units. For end-effector (EEF) actions,
we discard samples whose single-step translation magnitude exceeds
$0.2\,\mathrm{m}$ or whose single-step rotation magnitude exceeds
$0.5\,\mathrm{rad}$.

\paragraph{Instruction correctness and video-instruction consistency filtering.}
We apply an LLM-based filter to assess the correctness, clarity, and
executability of each natural-language instruction. Instructions that are
empty, malformed, gibberish, incomplete, or lack a clear action, object, or
target are removed before visual verification. For the remaining
instructions, we perform task-level video-instruction consistency checking
with a vision-language model. Using gripper-state signals recorded in the
robot data, we detect opening and closing events and split each episode into
action-centered video clips. The VLM first interprets these local clips
individually to obtain descriptions of the corresponding manipulation
actions. We then combine the clip-level descriptions with sparsely sampled
global frames from the full episode and the original instruction, and ask
the VLM to determine whether the observed behavior matches the instruction
and whether the task is completed. 

\paragraph{Manual semantic verification.}
After all automated filtering stages, we manually inspect
sampled episodes from every dataset--embodiment pair to verify
the interpretation of robot states and actions.
We replay the joint trajectories decoded by our data loaders
using the corresponding URDF and visualize the EEF poses
alongside the recorded videos from all available camera views. We assess whether the
visualized robot motion, end-effector poses, and gripper behavior
agree with the observations. Particular attention is paid to
dataset-specific conventions, including gripper opening and
closing definitions and coordinate-frame conventions. This
inspection helps identify semantic mismatches and verify that
our data loaders correctly interpret heterogeneous annotations
and map them consistently into the unified representation.

We apply the filtering and verification stages in a fixed order.
First, signal anomaly and consistency filtering removes invalid
or inconsistent data while retaining contiguous valid segments.
We then trim static episode boundaries, perform visual quality
filtering, and apply the action magnitude guard.
The remaining data undergo LLM-based instruction screening,
followed by VLM-based checks of video--instruction consistency
and task completion.
Finally, we manually inspect sampled episodes from each
dataset--embodiment pair to verify the decoded robot motion,
its agreement with the recorded observations, and the mapping
of dataset-specific conventions into the unified representation.

\subsection{Ego and Ego2Robot Data}
\subsubsection{Data Sources}
\paragraph{EgoDex~\citep{hoque2025egodex}}
EgoDex contains 829 hours of egocentric human demonstrations collected using Apple Vision Pro across 194 tabletop manipulation tasks. It pairs 30\,Hz RGB video with 3D head, upper-body, hand poses, camera intrinsics and poses, and language descriptions, providing fine-grained supervision for dexterous manipulation.

\paragraph{EgoVerse~\citep{punamiya2026egoverse}}
 EgoVerse is a continuously expanding collection of egocentric human demonstrations contributed by academic and industrial partners. We use an expanded EgoVerse snapshot of approximately 473K recordings, yielding 1.48M segmented episodes spanning 3,819 hours. The data pair egocentric video with 3D hand keypoints, 6-DoF head poses, and task descriptions, including subtask-level language annotations for industry-contributed recordings.

\subsubsection{Processing Rules}
The processing pipeline is organized into action alignment, kinematic
alignment, and visual alignment, as shown in~\Cref{fig:human2robot-pipeline}.
Our conversion procedure is inspired by Ego2Robot~\citep{wangEgo2RobotScalableRobot2026}
and Qwen-RobotManip~\citep{qwen_robotmanip}. Unlike their
representative-keyframe kinematic checks, we incorporate coarse-to-fine
trajectory validation into base selection. Ego2Robot data undergo all
stages, whereas egocentric data undergo only
action alignment and action speed alignment.

\paragraph{Action alignment.}
Action alignment converts heterogeneous egocentric human data into a standardized representation
of bimanual wrist states and actions. Specifically, we treat the wrist pose of
each 3D hand as the corresponding end-effector pose. To obtain a unified
camera-relative representation, we use the camera extrinsics to transform each
end-effector pose from the source frame into the instantaneous camera frame
and construct the corresponding state:
\begin{equation}
\label{eq:ego_state}
\mathbf{p}_{E_t}^{C_t} = R_s^{C_t}\mathbf{p}_{E_t}^{s} + \mathbf{t}_s^{C_t},
\quad
R_{E_t}^{C_t} = R_s^{C_t}R_{E_t}^{s},
\quad
\mathbf{s}_t = \bigl(\mathbf{p}_{E_t}^{C_t},\, R_{E_t}^{C_t},\, g_t\bigr).
\end{equation}
Here, $\mathbf{p}_{E_t}^{s}$ and $R_{E_t}^{s}$ denote the end-effector position
and orientation in the source frame $s$, respectively.
$R_s^{C_t}$ and $\mathbf{t}_s^{C_t}$ specify the transformation from the source
frame to the camera frame $C_t$, and $g_t$ denotes gripper openness. Notably, we estimate gripper openness from human fingertip geometry. The canonical mapping uses the distance between the thumb tip and the midpoint of the index and middle fingertips:
\begin{equation}
\label{eq:ego_gripper}
d_t = \left\|\mathbf{p}_{\mathrm{thumb},t} - \frac{\mathbf{p}_{\mathrm{index},t} + \mathbf{p}_{\mathrm{middle},t}}{2}\right\|_2,
\qquad
g_t = \operatorname{clip}\!\left(\frac{d_t - 0.05}{0.02},\, 0,\, 1\right).
\end{equation}
All fingertip positions are measured in meters and expressed in a common
coordinate frame at the same timestamp. Distances of $5\,\mathrm{cm}$ and
$7\,\mathrm{cm}$ map to fully closed ($g_t=0$) and fully open ($g_t=1$)
gripper targets, respectively. Following the action parameterization used for robot data, we define actions using camera-relative position deltas, EEF-local rotation
deltas, and absolute next-frame gripper targets:
\begin{equation}
\label{eq:ego_action}
\Delta\mathbf{p}_t = \mathbf{p}_{E_{t+1}}^{C_{t+1}} - \mathbf{p}_{E_t}^{C_t},
\quad
\Delta R_t = \bigl(R_{E_t}^{C_t}\bigr)^{\top} R_{E_{t+1}}^{C_{t+1}},
\quad
\mathbf{a}_t = \bigl(\Delta\mathbf{p}_t,\, \Delta R_t,\, g_{t+1}\bigr).
\end{equation}

After action alignment, we retain an episode only if at least one
hand has valid pose observations spanning more than $1\,\mathrm{s}$
and covering more than $70\%$ of the episode duration.
We then discard episodes in which both hands are static, defined
by each hand having position variance below
$5\times10^{-4}\,\mathrm{m}^2$ and rotation variance below
$0.1\,\mathrm{rad}^2$.
For hand trajectories, we apply the trajectory outlier detection
described in the robot data section and additionally reject episodes
with excessive linear or angular speeds.
We apply the same anomaly checks to the available camera trajectories.
To limit the effects of substantial locomotion during manipulation,
we also remove episodes with camera displacement greater than
$1\,\mathrm{m}$.

\begin{algorithm}[t]
    \caption{Coupled Base Search and Trajectory IK for Bimanual Arms}
    \label{alg:h2r-base-search}
    \small
    \begin{algorithmic}[1]
        \Require Robot model $\mathcal{M}$; $\{\mathcal{T}_a,\mathcal{W}_a\}_{a\in\{L,R\}}$; acceptance criteria and search limits
        \Ensure Base poses and joint trajectories $(b_L,b_R,\mathbf{Q}_L,\mathbf{Q}_R)$, or $\varnothing$
        \For{$a \in \{L,R\}$}
            \State $\mathcal{B}_a \gets \operatorname{Candidates}(\mathcal{T}_a,\mathcal{W}_a)$
            \For{$s\in(\mathrm{coarse},\mathrm{medium})$}
                \State $\mathcal{B}_a \gets \operatorname{SearchArmTrajectories}(\mathcal{B}_a,\mathcal{T}_a,\mathcal{W}_a;s)$
            \EndFor
        \EndFor
        \State $\mathcal{P} \gets \operatorname{PairBases}(\mathcal{B}_L,\mathcal{B}_R)$
        \For{$p=(b_L,b_R)\in\mathcal{P}$ in order}
            \State $(v,\mathbf{Q}_L,\mathbf{Q}_R) \gets \operatorname{SolveFullTrajectory}(p,\mathcal{T}_L,\mathcal{T}_R)$
            \If{$v$}
                \State \Return $(b_L,b_R,\mathbf{Q}_L,\mathbf{Q}_R)$
            \EndIf
        \EndFor
        \State \Return $\varnothing$
    \end{algorithmic}
\end{algorithm}

\paragraph{Kinematic alignment.}
Kinematic alignment couples base selection with inverse kinematics (IK): the
outer search proposes base configurations, while the inner solver holds each
base fixed and computes joint trajectories to assess its suitability.
Unlike the representative-keyframe checks described in Ego2Robot and
Qwen-RobotManip, we incorporate coarse-to-fine trajectory validation into the
search to detect tracking failures and discontinuous changes between IK
solution branches.

\Cref{alg:h2r-base-search} summarizes the coupled base search and trajectory IK
procedure for bimanual arms.
The shared robot model $\mathcal{M}$ is loaded in MuJoCo~\citep{todorov2012mujoco} from a robot description file
and includes the kinematic structure, joint limits, collision geometry, and
visual geometry required for IK, collision checking, and rendered occupancy
evaluation.
For arm $a$, $\mathcal{T}_a$ contains the end-effector targets and gripper commands,
$\mathcal{W}_a$ is its workspace index, and $\mathcal{B}_a$ contains base candidates
with their available joint trajectories and evaluation statistics.
The stage index $s$ takes the values coarse and medium in sequence,
with progressively denser trajectory evaluation frames.

$\operatorname{SearchArmTrajectories}$ uses workspace seeds and prior trajectories
for base-fixed IK, retaining a ranked shortlist subject to stage-specific tracking,
joint-limit, joint-discontinuity, and trajectory-completion criteria and search limits.
$\operatorname{PairBases}$ forms candidate pairs, rejects those violating minimum
base separation or exhibiting medium-stage cross-arm collisions, and orders the
survivors. Both functions use stage-specific priorities based on motion quality,
rendered image occupancy, and layout preferences without relaxing acceptance
criteria. Visual preferences favor less base and proximal-arm intrusion while
also considering overall robot coverage.
$\operatorname{SolveFullTrajectory}$ then solves full-frame trajectories with
cached arm/base solutions reused when available and returns a pass flag $v$
after trajectory-quality and cross-arm collision checks.
Both IK routines use hierarchical updates combining position and approach
tracking, contact-point and orientation refinement, and projected joint-space
regularization, as detailed in Appendix~\ref{app:h2r-ik}.
The search returns the first passing pair in the scheduled order.

\paragraph{Visual alignment.}
Visual alignment uses SAM3~\citep{carion_sam_2025} to segment human regions and ProPainter~\citep{zhou_propainter_2023} to remove
them from the source video. Using the selected base configurations and joint
trajectories, the target robot is rendered from the original camera viewpoint
and composited into the inpainted scene.

\begin{figure}[t]
    \centering
    \includegraphics[width=0.9925\linewidth]{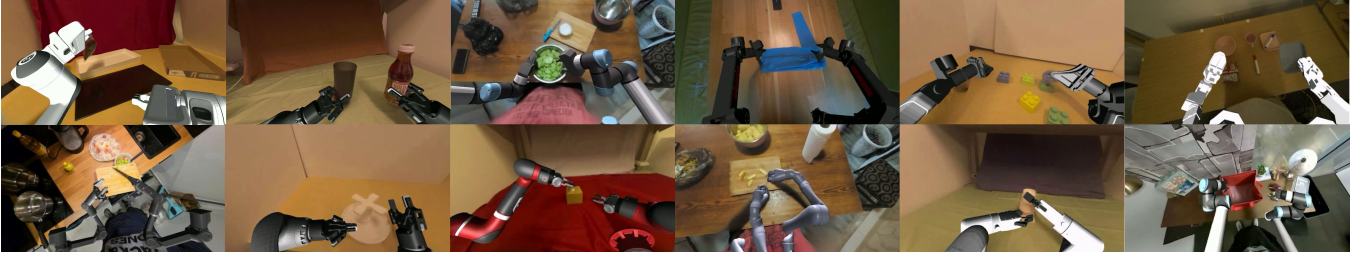}
    \caption{Qualitative Ego2Robot synthesis results.
        After action and kinematic alignment, the robot is rendered from the
        original egocentric viewpoint and composited into the human-removed
        scene with depth-aware occlusion.}
    \label{fig:ego2robot-results}
\end{figure}

Depth-aware compositing calibrates estimated scene depth using metric hand
keypoints and checks scale consistency near the manipulated objects. Object masks
constrain occlusion decisions, and temporal hysteresis reduces unstable visibility
changes across frames. The resulting videos are paired with the corresponding
smoothed end-effector targets and gripper commands, with validity masks retained
for downstream training, as shown in~\Cref{fig:ego2robot-results}.

\paragraph{Action speed alignment.}
To better match the motion speed of the robot data, we slow down EgoVerse and EgoDex trajectories by a factor of two. This doubles the trajectory duration while preserving the spatial path.

\begin{table*}[!t]
    \centering
    \caption{Source-video coverage and robot training data produced by the Ego2Robot pipeline.}
    \label{tab:human-to-robot-data-statistics}

    \small
    \setlength{\tabcolsep}{5pt}
    \renewcommand{\arraystretch}{1.08}
    \sisetup{
        group-separator = {,},
        group-minimum-digits = 4,
        table-number-alignment = center,
        detect-weight = true,
        mode = text
    }

    \begin{adjustbox}{max width=\linewidth}
\begin{tabular}{@{}l
        S[table-format=4.2]
        S[table-format=7.0]
        S[table-format=4.2]
        S[table-format=6.0]
        S[table-format=4.2]
        S[table-format=7.0]
        S[table-format=3.3]@{}}
    \toprule
    Dataset
    & \multicolumn{2}{c}{Source data}
    & \multicolumn{2}{c}{Converted data}
    & \multicolumn{3}{c}{Final data} \\
    \cmidrule(lr){2-3}
    \cmidrule(lr){4-5}
    \cmidrule(lr){6-8}
    & {Hours} & {Episodes}
    & {Hours} & {Episodes}
    & {Robot-hours} & {Episodes} & {Frames (M)} \\
    \midrule
    EgoDex~\citep{hoque2025egodex}
    & 772.16 & 310418 & 310.03 & 149753
    & 1702.33 & 1041346 & 183.852 \\
    EgoVerse~\citep{punamiya2026egoverse}
    & 3289.19 & 1312338 & 791.51 & 133697
    & 3931.43 & 2689167 & 424.410 \\
    \addlinespace[2pt]
    \textbf{Total}
    & \bfseries 4061.35 & \bfseries 1622756
    & \bfseries 1101.55 & \bfseries 283450
    & \bfseries 5633.77 & \bfseries 3730513 & \bfseries 608.262 \\
    \bottomrule
    \end{tabular}
\end{adjustbox}%
\end{table*}

\subsection{UMI Data}

\subsubsection{Data Sources}
\paragraph{Hy-UMI-10K~\citep{zhang2026hy}}
The public release of Hy-UMI-10K contains approximately 250K episodes spanning 2,162 hours of bimanual manipulation. It pairs  head- and wrist-camera RGB video with 6-DoF gripper poses, gripper openness, and task descriptions.

\subsubsection{Filtering}
Inspired by recent UMI data processing practices~\citep{zhang2026hy,team2026xiaomi}, we apply the following filtering procedure. We first perform basic segment cleanup and reject any segment containing
a gripper-pose quaternion $q$ with $\lVert q\rVert_2 \leq 10^{-8}$.
We then discard segments in which both hands are static, defined by
each hand having position variance below
$5\times10^{-4}\,\mathrm{m}^2$ and rotation variance below
$0.1\,\mathrm{rad}^2$. We follow the trajectory anomaly filtering procedure described in~\cref{sec:processing_rules} and additionally reject trajectories with
excessive linear or angular speeds or abrupt position or orientation
jumps between adjacent frames.

\subsection{Data Filtering and Conversion Statistics}
\label{sec:data-statistics}

\Cref{tab:pretraining-data-statistics} reports dataset-level hours and
episode counts before and after filtering, together with retained frame
counts, for robot, egocentric human, and UMI data.
\Cref{tab:human-to-robot-data-statistics} reports the Ego2Robot conversion
of EgoDex and EgoVerse only: the filtered human videos used as input,
the unique source coverage that completes synthesis, and the robot
training data obtained from conversion.

\paragraph{Data filtering.}
After the processing rules above, the robot collection is reduced from
13,867.71 hours across 1,370,003 episodes to 11,302.20 hours across
1,247,656 episodes, comprising 1,101.047 million frames.
EgoDex and EgoVerse jointly retain 4,061.35 of 4,640.04 hours and
1,622,756 of 1,818,396 episodes, totaling 438.626 million frames;
the retained durations are 772.16 and 3,289.19 hours, respectively.
Hy-UMI-10K is reduced from 2,161.74 hours across 250,135 episodes to
2,075.42 hours and 224.146 million frames.
Retained UMI trajectories are segmented into multiple training episodes,
so the episode count rises to 416,053.

\paragraph{Ego2Robot conversion.}
The source-data column of~\Cref{tab:human-to-robot-data-statistics}
is the filtered EgoDex and EgoVerse videos from~\Cref{tab:pretraining-data-statistics}.
The converted-data column is successful source coverage: the union of
intervals, across robot embodiments and processing batches, for which
at least one embodiment completes synthesis.
Each covered original source episode is counted once, even if only part
of it is converted; only the successful intervals contribute to covered
duration.
The final-data column is the exported robot training set.

Completing synthesis requires a selected base configuration and a joint
trajectory that satisfy the tracking, joint-limit, joint-discontinuity,
and trajectory-completion criteria in Kinematic Alignment, together with
base-separation and cross-arm collision checks for bimanual setups.
An interval with no such configuration cannot be converted.
Failures in subsequent synthesis stages, including visual alignment, likewise
omit an interval from successful coverage, so unsuccessful coverage is not
attributed solely to kinematic infeasibility.

Robot-hours sum usable durations across embodiments, reported before the
factor-of-two slowdown in Action Speed Alignment, and robot episodes count
the exported training segments of at least 32 action steps.
One source episode can therefore yield multiple robot episodes, and the
final training volume is larger than the unique source coverage.
Across EgoDex and EgoVerse, the pipeline covers 1,101.55 hours from
283,450 unique source episodes and yields 5,633.77 robot-hours across
3,730,513 training episodes, totaling 608.262 million frames.

\section{Training}
\label{sec:training-recipe}
We train \modelname in two stages. The first stage jointly optimizes a pretrained video expert and a randomly initialized action expert to a heterogeneous mixture of robot trajectories. The second stage specializes the resulting checkpoint to a target embodiment and task distribution. 

\subsection{Pretraining}
\label{sec:Pretraining}
\paragraph{Data and temporal sampling.}

Pretraining uses a heterogeneous hybrid dataset, including real robot demonstrations (80\%), ego-to-robot (Ego2Robot) data (10\%), UMI-collected demonstrations (8\%), and egocentric (Ego) demonstrations (2\%). These data sources cover different robot forms, camera layouts, state conventions, and task vocabularies. Each data source is converted to a canonical 80-dimensional state and action space by its adapter before entering the hybrid dataset. The dataset streams are weighted according to the number of valid start frames, and a trajectory-level balancing mechanism prevents a small number of longer segments from dominating the hybrid dataset. We use fixed time windows for training instead of using complete segments. A training sample contains 33 video frames and a 32-step action block. The video stream and action stream are sampled at a fixed frequency ratio of 4; the most recent observation lags the current observation by 32 action steps, and the anchor point is taken from the beginning of the episode. Images are packed into a 384×256 training canvas used by the pretraining scheme. 
We initially pretrain \modelname without 4D-aware representation
distillation, and then continue pretraining with the auxiliary
distillation objective described in~\Cref{sec:track-distillation} alongside the existing training
objectives.
During this final phase, each per-GPU mini-batch of 16 samples
includes one sample paired with an offline-cached Track4World~\citep{lu2026track4world}
descriptor.
The auxiliary distillation loss is applied only to this sample.

\paragraph{Initialization.}
The video expert model is initialized using Wan2.2-TI2V-5B~\citep{wan2025wan}. The action expert model is randomly initialized. The visual language encoder is initialized using RynnBrain1.1-2B~\citep{dang2026rynnbrainopenembodiedfoundation} and remains frozen during training. We use two language pathways, T5~\citep{raffel2020t5} and VLM. The two language paths assume different responsibilities during training. Instructions are converted into T5 embeddings for use by the video expert, thus preserving the pre-trained language-video interface of Wan2.2. T5 receives instructions but not the current scene, so its embeddings alone cannot determine which object is being referred to, its position in the camera view, or the progress of the task. To provide this missing observational basis, a valid current view and the same instructions are jointly passed to the RynnBrain1.1-2B vision language model. Its dense hidden states are appended with a proprioceptive token and provided to the action expert.

\begin{table}[!t]
  \centering
  \caption{Pretraining settings.}
  \label{tab:pretraining-settings}
  \small
  \setlength{\tabcolsep}{6pt}
  \renewcommand{\arraystretch}{1.1}
  \begin{tabular}{ll}
      \toprule
      Setting & Value \\
      \midrule
      Canonical state/action dimension
      & 80 \\
      Video resolution
      & $384\times256$ \\
      Video frames
      & 33 \\
      Action horizon
      & 32 \\
      Video-to-action frequency ratio
      & 4 \\
      Recent-frame offset
      & 32 action steps \\
      Batch size
      & 16 \\
      Gradient accumulation
      & 1 \\
      Optimizer
      & AdamW \\
      Learning rate
      & $5\times10^{-5}$ \\
      Weight decay
      & $1\times10^{-2}$ \\
      Learning-rate schedule
      & 5\% warm-up followed by cosine decay \\
      Mixed precision
      & bfloat16 \\
      Video expert
      & Wan2.2-TI2V-5B~\citep{wan2025wan} \\
      Action expert
      & ActionDiT with random initialization \\
      VLM
      & RynnBrain1.1-2B (frozen)~\citep{dang2026rynnbrainopenembodiedfoundation} \\
      Video flow-matching loss weight
      & 0.5 \\
      Action loss weight(Robot/UMI/Ego2Robot/Ego)
      & 1.0/0.5/0.1/0.1 \\
      Causal Imprint loss weight
      & 0.5 \\
      Future-feature alignment loss weight
      & 0.1 \\
      Distillation loss weight & 0.1 \\
      \bottomrule
  \end{tabular}
\end{table}

\paragraph{Implementation and hyperparameters.}

We pre-train our model on 256 NVIDIA A800 GPUs using bfloat16 mixed precision. The per-GPU batch size is 16, with no gradient accumulation, resulting in an effective global batch size of 4,096. We use the AdamW optimizer with a learning rate of $5 \times 10^{-5}$ and a weight decay of $1 \times 10^{-2}$. The learning-rate schedule consists of a 5\% warm-up phase followed by cosine decay. The weights for the video, Causal Imprint, and future-feature alignment loss are set to 0.5, 1.0, 0.5, and 0.1, respectively. We use data-source-dependent action loss weights: 1.0 for robot demonstrations, 0.5 for UMI-collected demonstrations, and 0.1 for both ego-to-robot (Ego2Robot) data and egocentric (Ego) demonstrations. Each weight scales only the action loss of samples from the
corresponding data category. Pretraining consists of two consecutive phases. We first train for 235K optimization steps without 4D-aware representation distillation. We then continue training for an additional 10K steps, adding the distillation objective described in~\Cref{sec:track-distillation} with a weight of 0.1 while retaining the existing training objectives. The complete pretraining run comprises 245K optimization steps and takes approximately 14 days. The main hyperparameters are summarized in~\Cref{tab:pretraining-settings}.

\begin{table}[t]
  \centering
  \caption{Simulation post-training configurations. LIBERO-Plus is used only for evaluation. RoboTwin~2.0 is trained on the clean split and evaluated on the randomized.}
  \label{tab:simulation-posttraining}
  \small
  \setlength{\tabcolsep}{5pt}
  \renewcommand{\arraystretch}{1.12}
\resizebox{\linewidth}{!}{%
\begin{tabular}{lccccc}
      \toprule
      Benchmark
      & Training split
      & Views
      & Resolution
      & Normalization
      & Epochs \\
      \midrule
      EBench~\citep{ebench}
      & EBench train
      & 3
      & $384\times256$
      & Z-score
      & 10 \\
      RoboDojo~\citep{robodojo}
      & RoboDojo train
      & 3
      & $384\times256$
      & Z-score
      & 10 \\
      LIBERO-Plus~\citep{libero_plus}
      & LIBERO~\citep{libero} train
      & 2
      & $384\times256$
      & Min--max
      & 15 \\
      RoboTwin~2.0~\citep{robotwin2}
      & Clean
      & 3
      & $384\times256$
      & Z-score
      & 10 \\
      \bottomrule
  \end{tabular}
}
\end{table}

\subsection{Post-Training for Simulation}
\label{sec:post-training-sim}

We post-train \modelname independently on four simulation benchmarks: EBench~\citep{ebench}, RoboDojo~\citep{robodojo}, LIBERO-Plus~\citep{libero_plus}, and RoboTwin~2.0~\citep{robotwin2}. Each experiment starts from \textbf{the same pretrained checkpoint} and uses a benchmark-specific data adapter to convert the native observation, state, and action interfaces into our unified representation. Native action dimensions are mapped into the canonical 80-dimensional action space, while dimensions that are not defined for a given embodiment are masked from the training objective.

Unless otherwise stated, we use a batch size of 16, eight data-loading workers per process, bfloat16 mixed precision, and a learning rate of $5\times10^{-5}$. Each sample is constructed from a 33-step trajectory window and contains a 32-step action chunk. Visual observations are sampled every four action steps, resulting in nine observation frames for each training sample. Multi-view observations are packed into a common $384\times256$ canvas across all four benchmarks. The benchmark-specific configurations are summarized in~\Cref{tab:simulation-posttraining}.

\paragraph{EBench~\citep{ebench}.}

EBench is a mobile bimanual manipulation benchmark containing 26 tasks that cover tabletop manipulation, pick-and-place, and long-horizon interaction. The benchmark is designed to test a broad range of manipulation capabilities and generalization factors, including tasks that require coordinated dual-arm manipulation and mobile-base motion. We use the three RGB observations provided by the benchmark, consisting of one external view and two wrist views, and pack them into the common $384\times256$ input canvas.

The control interface contains end-effector targets for both arms, gripper commands, and planar mobile-base motion. We convert each arm pose into the unified end-effector representation, reduce the two finger coordinates of each gripper to a single gripper value, and retain the three-dimensional base motion command. These components are then mapped into their corresponding entries in the canonical action space. We use z-score normalization and post-train for 10 epochs.

\paragraph{RoboDojo~\citep{robodojo}.}

RoboDojo provides 42 simulation tasks for bimanual manipulation and organizes them along five capability dimensions: generalization, memory, precision, long-horizon execution, and open-vocabulary instruction following. The tasks are designed to evaluate more than short-horizon pick-and-place behavior and include challenging multi-stage and precision-sensitive interactions. For post-training, we use the high camera together with the left and right wrist cameras and pack the three views into a $384\times256$ canvas.

The standard simulated embodiment uses a 14-dimensional bimanual joint-space interface, with six arm joints and one gripper coordinate for each arm. The benchmark-specific adapter maps the valid joint and gripper channels into the canonical action representation and masks dimensions that are not used by this embodiment. We use z-score normalization and post-train for 10 epochs.

\paragraph{LIBERO~\citep{libero}.}

LIBERO is a single-arm manipulation benchmark built around a Franka robot and contains four standard task suites: LIBERO-Spatial, LIBERO-Object, LIBERO-Goal, and LIBERO-10. We post-train jointly on demonstrations from these four suites. Each observation contains an external scene view and a wrist view, which are packed into the common $384\times256$ input representation.

The native action consists of a six-dimensional end-effector command and a scalar gripper command, which are mapped into the corresponding entries of the canonical action space. We use min--max normalization and post-train for 15 epochs.

We evaluate the resulting checkpoint on the LIBERO-Plus robustness setting. While standard LIBERO measures performance close to the post-training distribution, LIBERO-Plus preserves the underlying task semantics while introducing controlled distribution shifts along seven dimensions, including object layout, camera viewpoint, robot initialization, language instruction, lighting, background appearance, and sensor noise. This provides a direct measure of how well the post-trained policy generalizes when the visual, linguistic, and execution conditions deviate from those observed during training.

\paragraph{RoboTwin~2.0~\citep{robotwin2}.}

RoboTwin~2.0 is a bimanual manipulation benchmark containing 50 tasks for the ALOHA-AgileX embodiment. We use three RGB observations, consisting of one high camera and two wrist cameras, and pack them into a $384\times256$ training canvas. The native 14-dimensional action space contains six joint coordinates and one gripper coordinate for each arm, which are mapped to the corresponding bimanual entries of our canonical action representation.

Following the Clean2Random evaluation protocol, we post-train exclusively on demonstrations collected under the clean environment configuration and evaluate the same checkpoint under both clean and progressively randomized environments. The randomized evaluation introduces controlled changes in background appearance, lighting, scene clutter, and tabletop height, with the Hard setting combining multiple sources of variation. This setting directly measures whether the policy can retain its manipulation capability when the visual and physical environment deviates from the post-training distribution. We use z-score normalization and post-train for 10 epochs.

\begin{table}[t]
    \centering
    \caption{Real-robot datasets used for post-training and evaluation.}
    \label{tab:real_robot_datasets_real}

    \small
    \setlength{\tabcolsep}{5pt}
    \renewcommand{\arraystretch}{1.08}

    \begin{tabular}{lrrrr}
        \toprule
        Task & Episodes & Frames & Duration (min) & Epochs \\
        \midrule
        \rowcolor{gray!10}\multicolumn{5}{l}{\textbf{AC-One (gripper)}}\\
        \midrule
        Luminol reaction & 162 & 433{,}715 & 240.95 & 10 \\
        Get a drink & 137 & 244{,}440 & 135.80 & 10 \\
        Toast bread & 304 & 311{,}631 & 173.13 & 10 \\
        MOF experiment & 350 & 1{,}173{,}626 & 652.02 & 10 \\
        \midrule

        \rowcolor{gray!10}\multicolumn{5}{l}{\textbf{Arx5 (gripper)}}\\
        \midrule
        Magnetic stirrer & 202 & 66{,}353 & 36.86 & 10 \\
        Place tube & 2{,}664 & 555{,}702 & 308.72 & 10 \\
        \midrule

        \rowcolor{gray!10}\multicolumn{5}{l}{\textbf{Franka+XHand (dexterous hand)}}\\
        \midrule
        Pour water & 50 & 44{,}704 & 24.84 & 100 \\
        Place fruit into box & 48 & 39{,}253 & 21.81 & 100 \\
        Stack cups & 54 & 48{,}264 & 26.81 & 100 \\
        \midrule

        \rowcolor{gray!10}\multicolumn{5}{l}{\textbf{TianJi Marvin+Wuji Hand (dexterous hand)}}\\
        \midrule
        Use dropper & 101 & 107{,}703 & 59.84 & 50 \\
        Make a sandwich & 101 & 106{,}731 & 59.30 & 50 \\
        \bottomrule
    \end{tabular}
    \vspace{-3mm}
\end{table}

\subsection{Post-Training for Real Robots}
\label{sec:post-training-real}

\paragraph{Setup.}
We post-train \modelname on 4 robot setups, including two \textbf{gripper-based} ones and two \textbf{dexterous-hand} ones. For \textbf{gripper-based} setups, task-specific datasets are collected on the ARX AC-One real-robot platform and Arx5 single-arm robot. For \textbf{dexterous-hand} setups, we post-train the same pretrained checkpoint on the bimanual Franka+XHand platform and TianJi Marvin+Wuji Hand combination. We have collected 11 tasks across all 4 setups, whose details are summarized in
\Cref{tab:real_robot_datasets_real}. The post-training settings for real robots are included in \Cref{tab:real-robot-posttraining-settings}. It is worth mentioning that all post-training share the same settings and start from the same pretrained checkpoint.

\paragraph{RTC post-training.}
During real-robot post-training, we train the policy to predict
32-step action chunks. Following training-time RTC~\citep{rtc_training_time_arxiv2512_05964},
we simulate inference delay by sampling a committed prefix length
$d\sim\mathcal{U}\{0,\dots,16\}$ for each full-length training sample;
for padded samples, the maximum prefix length is capped by the valid
sequence length to retain at least one supervised suffix action.
The first $d$ ground-truth actions remain clean, with their per-token flow
timesteps set to zero, while noise is applied to the remaining actions.
The action loss is computed only over valid suffix tokens and action
dimensions. This trains the policy to predict an action continuation
conditioned on a known prefix. At deployment, the prefix is supplied by
the previously generated plan, as described in
\Cref{sec:real-robot-deployment}.

\begin{table}[!t]
  \centering
  \caption{Post-training settings for real robots.}
  \label{tab:real-robot-posttraining-settings}
  \small
  \setlength{\tabcolsep}{6pt}
  \renewcommand{\arraystretch}{1.1}
  \begin{tabular}{ll}
      \toprule
      Setting & Value \\
      \midrule
      Canonical state/action dimension
      & 80 \\
      Video resolution
      & $384\times256$ \\
      Video frames
      & 33 \\
      Action horizon
      & 32 \\
      Video-to-action frequency ratio
      & 4 \\
      Recent-frame offset
      & 32 action steps \\
      Batch size
      & 16 \\
      Gradient accumulation
      & 1 \\
      Optimizer
      & AdamW \\
      Learning rate
      & $5\times10^{-5}$ \\
      Weight decay
      & $1\times10^{-2}$ \\
      Learning-rate schedule
      & 5\% warm-up followed by cosine decay \\
      Mixed precision
      & bfloat16 \\
      Video expert
      & Wan2.2-TI2V-5B~\citep{wan2025wan} \\
      Action expert
      & ActionDiT \\
      VLM
      & RynnBrain1.1-2B (frozen)~\citep{dang2026rynnbrainopenembodiedfoundation} \\
      Video loss weight
      & 0.5 \\
      Action loss weight
      & 1.0 \\
      Causal Imprint loss weight
      & 0.5 \\
      Future-feature alignment loss weight
      & 0.1 \\
      Distillation loss weight
      & 0.1 \\
      \bottomrule
  \end{tabular}
\end{table}

\section{Infrastructure}
\label{sec:Infrastructure}
\label{sec:infra}

\subsection{Training Infrastructure for Model Development}
\label{sec:training}

During model iteration, we identified two main sources of overhead: repeated extraction of visual features from samples revisited across epochs, and the compute and activation memory required by the MoT backbone. We address these costs at complementary stages of the training pipeline. A unified cache reuses the outputs of the frozen VAE and VLM, while layerwise compilation and checkpointing improve the execution of the MoT backbone. Together, these optimizations reduce redundant encoder work and allow the backbone to be configured for either throughput or activation memory.

\subsubsection{Feature Caching}
\label{sec:cache}
\label{sec:vlm-vae-cache}

For a training sample $x_i$, let
\begin{equation}
    z_i = E_{\mathrm{VAE}}(x_i), \qquad
    h_i = E_{\mathrm{VLM}}(x_i,p_i,s_i),
\end{equation}
where $z_i$ is the video latent, $h_i$ is the VLM context, and $(p_i,s_i)$
denote the task prompt and state. Because the same samples are revisited during
training, repeatedly evaluating the frozen encoders expends computation
without changing these representations. Feature caching therefore aims to
amortize encoding across training steps while preserving preprocessing,
temporal alignment, camera ordering, and model semantics.

\begin{figure}[!t]
    \centering
    \includegraphics[width=\linewidth]{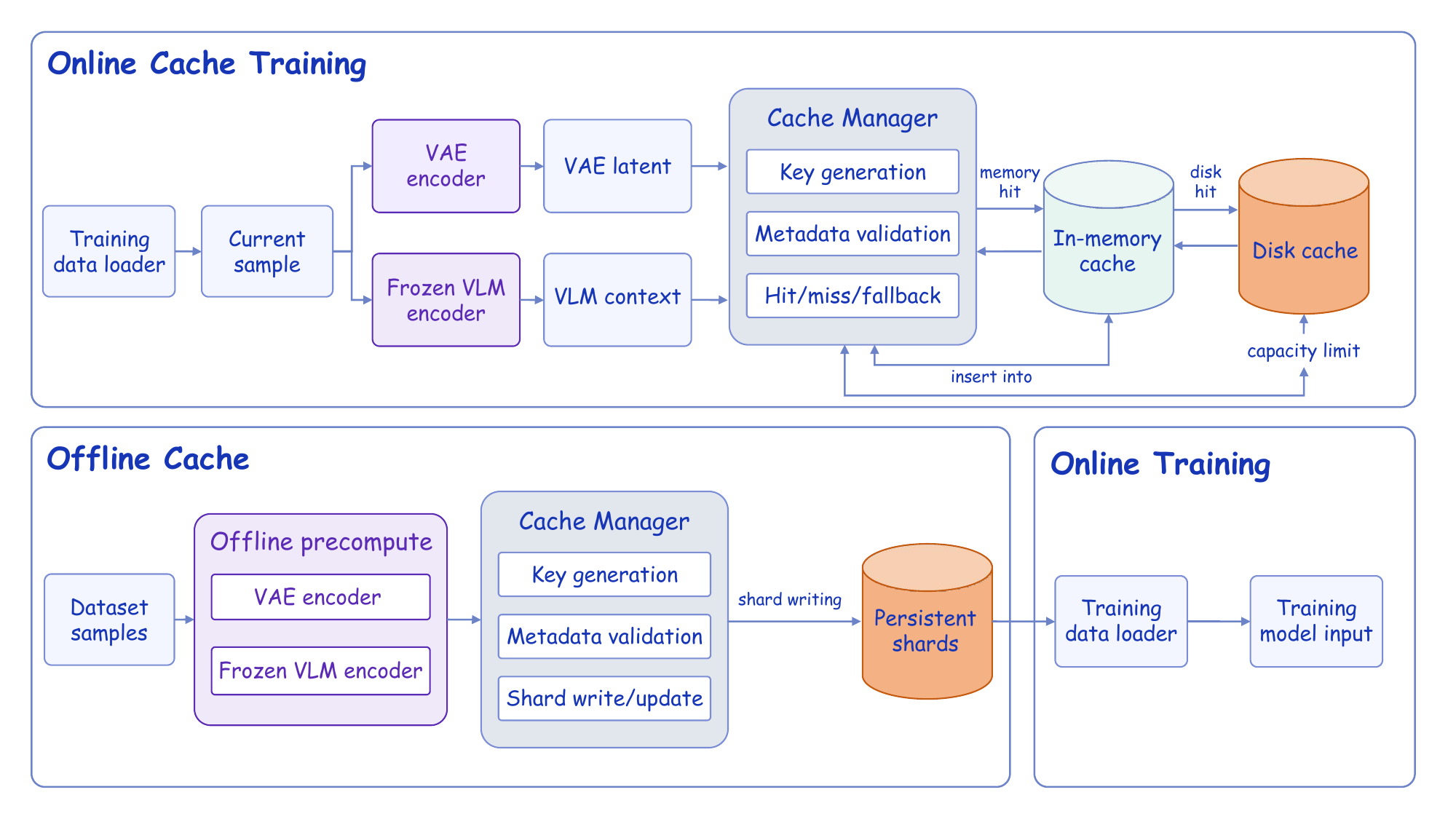}
    \caption{Feature caching for exploratory training. The Cache Manager serves VAE latents and VLM contexts to training
    through a common interface for online and offline materialization, reducing repeated feature extraction during model development.}
    \label{fig:vlm-vae-cache-manager}
\end{figure}

\Cref{fig:vlm-vae-cache-manager} shows the Cache Manager between the data
loader, feature encoders, and storage backends. Built on
LiteGen~\citep{contributors2026litegenrepo}, it performs lookups, validates
artifact metadata, dispatches reads and writes, and records hits, misses, and
fallbacks. To prevent reuse across incompatible data or encoder
configurations, each artifact is identified by its sample and representation
provenance. A representative key is
\begin{equation}
    \mathrm{key} = \mathrm{hash}(D,e,k,v,P,M),
\end{equation}
where $D$, $e$, $k$, and $v$ identify the dataset version, episode, temporal
chunk, and camera view, and $P$ and $M$ identify the preprocessing and model
versions. VLM artifacts must additionally account for any prompt or state
inputs that affect $h_i$. Metadata validation rejects incompatible or
corrupted artifacts; an online miss can then fall back to encoding rather
than silently consuming an invalid feature.

The same artifact contract supports two materialization strategies. In
\emph{online caching}, the manager first checks memory and then disk for each
sample. A miss invokes the corresponding VAE or frozen VLM encoder, after
which the feature is inserted into memory and, when needed, persisted to disk.
Entries are thus populated as the dataset is traversed: this avoids a
separate preprocessing job, although the first pass still pays the encoding
cost. Asynchronous persistence limits the impact of writes on training. In
\emph{offline caching}, batched encoder inference produces persistent feature
shards before training begins. A manifest records artifact keys, coverage,
tensor shapes and data types, preprocessing versions, and model fingerprints;
a startup check validates the manifest and index. The one-time preprocessing
cost can then be amortized over subsequent epochs and repeated experiments
using the same data and encoder versions.

\subsubsection{Layerwise MoT Optimization}
\label{sec:mot-compile}

\begin{figure*}[!t]
    \centering
    \includegraphics[width=\linewidth]{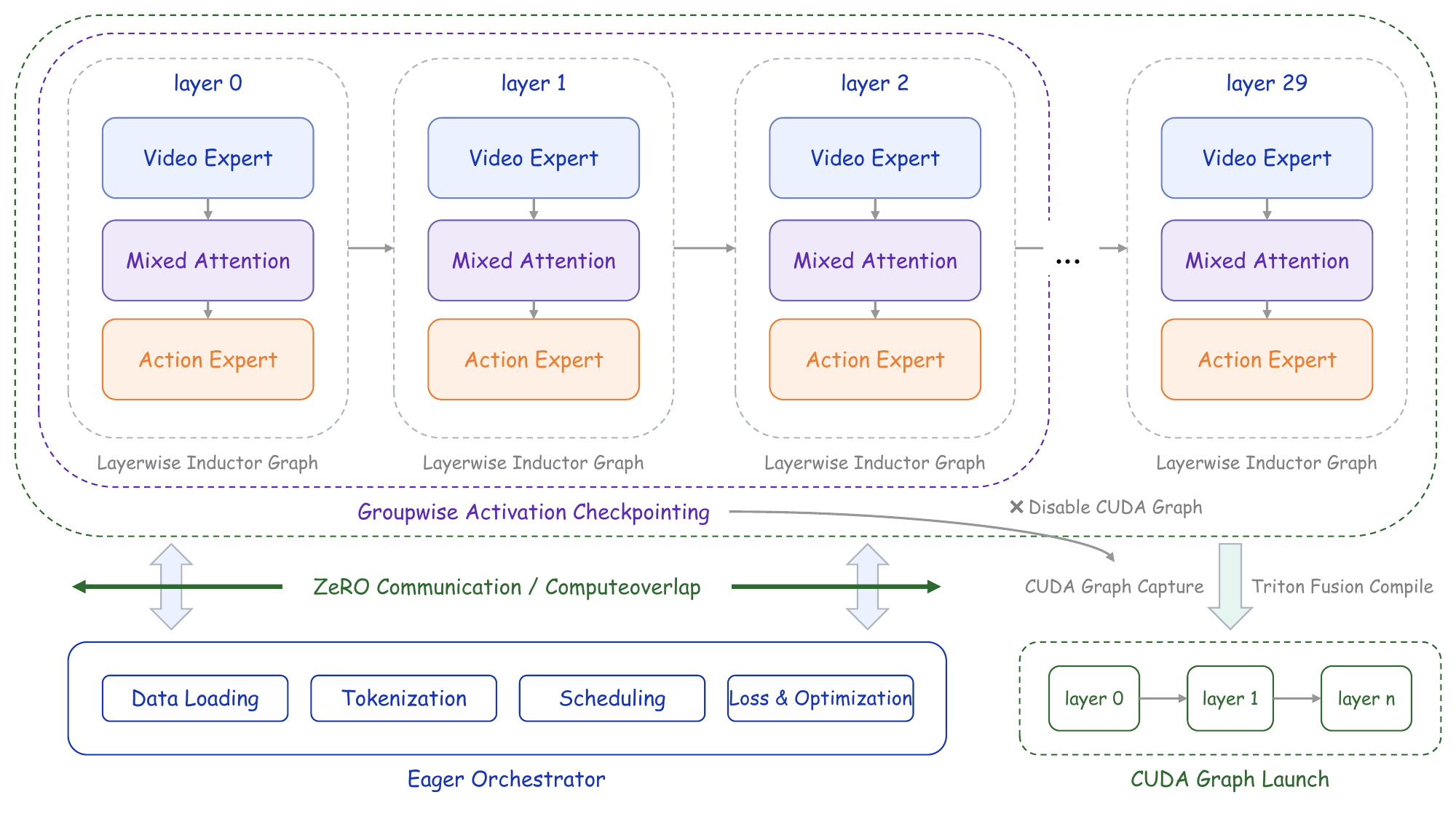}
    \caption{Layerwise compilation of the two-stream MoT backbone. An eager
    orchestrator invokes independently compiled layers and groups three
    consecutive layers for external non-reentrant checkpointing when memory
    reduction is required. VLM contexts are padded before action
    cross-attention.}
    \label{fig:mot-layerwise-compilation}
\end{figure*}

The MoT backbone contains 30 blocks. In each block, a video expert and an
action expert update separate token streams while sharing mixed
self-attention. The experimental configuration uses a 3{,}072-dimensional
Wan video stream and a 1{,}024-dimensional action stream; the latter also
attends to text, the current VLM context, and proprioceptive tokens. Our
objective is to improve the throughput and memory efficiency of this
backbone while preserving its attention routing and intended gradients.

As illustrated in~\Cref{fig:mot-layerwise-compilation}, we compile each
MoT layer as an independent PyTorch Inductor graph and invoke the layers from
an eager orchestrator. Layer boundaries constrain recompilation, allow
per-layer profiling, and retain opportunities to overlap ZeRO communication
with computation. We compile supported regions while leaving operations that
Inductor cannot lower in eager execution. Mixed attention uses PyTorch
FlexAttention with a 64-token block size; constructing the block mask outside
the compiled layer keeps the causal and stream-specific routing rules
consistent across execution modes. To stabilize the compiled action-layer
shapes, we pad VLM contexts and validity masks to 640 tokens. The current
LIBERO~\citep{libero} cache reaches 614 tokens after appending the proprioceptive token;
added positions are masked, and longer contexts are rejected rather than
truncated.

For throughput-oriented training, the \texttt{reduce-overhead} mode uses
layerwise compilation with CUDA Graph capture where eligible. When activation
memory is the limiting factor, the eager orchestrator instead places three
consecutive compiled layers inside each external non-reentrant checkpoint
region. This recomputes the grouped activations during backward while
retaining independent compiler graphs for individual layers; the
expert-internal and eager mixed-attention checkpoint switches are disabled
for this configuration. CUDA Graph capture is also disabled on the
checkpointed path because its interaction with AOTAutograd and non-reentrant
checkpointing produced incorrect gradients in validation.

\subsubsection{Training Efficiency}
\label{sec:infra-training-efficiency}

The proposed optimizations can be enabled independently according to the
data characteristics and resource constraints at different stages of model
development. On LIBERO~\citep{libero} and RoboTwin~\citep{robotwin2}, combining VAE/VLM feature caching
with the workload-specific layerwise MoT configuration yields end-to-end
training throughput speedups of $3.02\times$ and $2.11\times$,
respectively, over the corresponding unoptimized baselines. When stochastic
inputs preclude feature reuse, or when the dataset makes cache
materialization impractical, the throughput-oriented layerwise MoT
optimization remains applicable on its own and improves end-to-end
throughput by 20--30\%. These optimizations reduce the time for
architecture and hyperparameter exploration across training regimes.

\subsection{Real-Robot Deployment}
\label{sec:Deployment}
\label{sec:real-robot-deployment}
\paragraph{Deployment setup.}
We deploy \modelname with inference running locally on a single
NVIDIA RTX~5090 GPU with 32~GiB of memory and a control rate of 30~Hz.
The observation configuration varies by platform: AC-One uses one head
camera and two wrist cameras; Arx5 uses one wrist camera and two external
cameras; bimanual Franka+XHand uses a single external camera; and
TianJi Marvin+Wuji Hand uses one head camera and one wrist camera.

\paragraph{Asynchronous execution strategy.}
Asynchronous execution requires each new action chunk to remain consistent
with actions already committed to execution. Existing strategies address
this requirement through action blending, inference-time guidance, or
training-time prefix conditioning. A recent empirical study of real-time
World Action Models compares these approaches and documents their trade-offs
in precision, smoothness, and execution efficiency~\citep{motubrain2026realtime}.

We use the prefix-conditioned policy trained in
\Cref{sec:post-training-real}, supplying committed actions as clean inputs
during sampling. This avoids additional gradient-based RTC
guidance~\citep{rtc_inference_time_arxiv2506_07339} and is compatible with
our compiled deployment pipeline.

The controller retains a 32-step prediction horizon and launches a single
background inference request every 16 execution steps while the remaining
actions of the current plan execute. The worker receives a fresh observation and a
temporally aligned prefix from the previous plan. Prefix tokens are held
clean at timestep zero and re-clamped to the supplied actions throughout
sampling. The prefix length is estimated conservatively from recent
inference delays and capped at the trained maximum of 16 steps.

\paragraph{Inference optimization.}
\label{sec:Inference}
\label{sec:inference-optimization}
Asynchronous inference must complete within a bounded time window.
After a replan is issued, the controller must receive the next action chunk
before the remaining 16 actions are exhausted. At 30~Hz, this window
is 533~ms; observation transport, policy inference, prefix alignment, and
controller-side command dispatch all consume it, and robot-side scheduling
jitter reduces the remaining margin. The policy must therefore meet a
round-trip latency constraint, not merely attain high nominal throughput.

At inference, \modelname executes the action-generation path of the directed
MoT. The runtime first isolates policy inference in a separate rclpy-free
process, preventing ROS callbacks, timers, and Python GIL scheduling in the
robot node from interfering with CUDA launch execution. The remaining runtime
exploits two invariants of this path. First, conditioning features and their
attention K/V projections are fixed within one action invocation and are
reused across diffusion steps. Second, the action hot path has fixed tensor
shapes for a given deployment configuration. Cached K/V tensors are packed
across layers, groups of consecutive MoT action layers are compiled as
fixed-shape callables, and CUDA Graph Trees replay these groups to reduce
launch overhead. These options do not change the model inputs, diffusion
schedule, attention semantics, action interface, or RTC prefix contract.

\Cref{tab:rtc-inference-ablation} reports a cumulative ablation of
controller-observed round-trip latency on the dexterous-hand deployment.
Each request is timed from when the controller issues a replan
to when the returned action chunk is available. The RTT includes
controller-side request handling and transport to the isolated policy
process, as well as observation preprocessing, state normalization,
recent-frame/prefix alignment, conditioning, action inference, action
mapping, and output post-processing. Action inference runs on one
RTX~5090 GPU. Three warm-up requests are discarded, and the following 50
requests are averaged per configuration.

\begin{table}[t]
\centering
\caption{Cumulative inference-latency ablation on dexterous-hand
deployment. Success rates are from separate LIBERO-Plus~\citep{libero_plus} simulations.}
\label{tab:rtc-inference-ablation}
\small
\setlength{\tabcolsep}{5pt}
\renewcommand{\arraystretch}{1.1}
\begin{adjustbox}{max width=\linewidth}
\begin{tabular}{lrrr}
  \toprule
  Configuration & Round-trip time (ms) $\downarrow$ & Speedup &
  Success rate (LIBERO-plus, \%) \\
  \midrule
  Standard runtime & 780.5 & 1.00$\times$ & 92.78 \\
  \quad + process isolation & 374.1 & 2.09$\times$ & 92.67 \\
  \quad + feature caching \& compilation & 249.8 & 3.12$\times$ & 92.46 \\
  \quad + context caching & 217.3 & 3.59$\times$ & 92.41 \\
  \quad + grouped action execution & 186.2 & 4.19$\times$ & 92.33 \\
  \quad + CUDA-graph replay & 152.8 & 5.11$\times$ & 92.23 \\
  \bottomrule
\end{tabular}
\end{adjustbox}
\end{table}

\section{Experiments}

\subsection{Evaluation Protocol}
We evaluate \modelname on four simulation benchmarks, including LIBERO-Plus~\citep{libero_plus}, RoboTwin~2.0~\citep{robotwin2}, EBench~\citep{ebench}, and RoboDojo~\citep{robodojo}. The post-training setup for each benchmark has been described in~\Cref{sec:post-training-sim}. For each benchmark, we follow its official simulator, task definitions, and success criteria without additional adaptation on the evaluation environments. Baseline results are taken from the corresponding papers or public benchmark results under matching evaluation protocols.

\subsection{Simulation Benchmark Results}

\subsubsection{LIBERO-Plus}
We train \modelname on the standard LIBERO~\citep{libero} training set and directly evaluate it on LIBERO-Plus~\citep{libero_plus} without adaptation to the evaluation environments. As shown in~\Cref{tab:main_libero_plus}, \modelname achieves the best overall success rate of 92.8\%, outperforming Qwen-RobotManip-Context~\citep{qwen_robotmanip} by 1.4 percentage points and the strongest prior WAM, Being-H0.7~\citep{being_h07}, by 8.0 points. The largest gains appear under robot perturbations, where \modelname reaches 91.1\%, compared with 87.4\% for the previous best result. It also achieves the best performance under camera and language perturbations, reaching 90.6\% and 92.9\%, respectively. In addition, \modelname remains competitive on background and layout perturbations, with 99.4\% and 88.4\% success rates.

\begin{table}[!t]
    \centering
    \caption{
        Zero-shot robustness on LIBERO-Plus.
        All values are success rates (\%).
    }
    \label{tab:main_libero_plus}

    \small
    \setlength{\tabcolsep}{3.0pt}
    \renewcommand{\arraystretch}{1.06}

    \resizebox{\linewidth}{!}{%
    \begin{tabular}{l*{8}{c}}
        \toprule
        Method
        & Camera
        & Robot
        & Language
        & Light
        & Background
        & Noise
        & Layout
        & Total $\uparrow$ \\
        \midrule

        \rowcolor{gray!10}
        \multicolumn{9}{c}{\textbf{VLA}} \\
        \midrule

        $\pi_{0}$~\citep{pi0}
        & 13.8 & 6.0 & 58.8 & 85.0
        & 81.4 & 79.0 & 68.9 & 53.6 \\

        $\pi_{0}$-FAST~\citep{pi0_fast}
        & 65.1 & 21.6 & 61.0 & 73.2
        & 73.2 & 74.4 & 68.8 & 61.6 \\

        RIPT-VLA~\citep{ript_vla}
        & 55.2 & 31.2 & 77.6 & 88.4
        & 91.6 & 73.5 & 74.2 & 68.4 \\

        OpenVLA-OFT~\citep{openvla_oft}
        & 56.4 & 31.9 & 79.5 & 88.7
        & 93.3 & 75.8 & 74.2 & 69.6 \\

        StarVLA~\citep{starvla}
        & 52.5 & 49.8 & 88.5 & 95.7
        & 95.7 & 73.0 & 76.9 & 74.1 \\

        VLA-JEPA~\citep{vla_jepa}
        & 63.3 & 67.1 & 85.4 & 95.6
        & 93.6 & 66.3 & 85.1 & 79.5 \\

        VLAct~\citep{vlact}
        & 73.9 & 68.4 & 81.5 & 96.7
        & 96.7 & 86.0 & 83.3 & 82.6 \\

        $\pi_{0.5}$~\citep{pi05}
        & 78.4 & 73.6 & 80.8 & 96.2
        & 94.1 & 89.0 & 84.5 & 84.4 \\

        InternVLA-A1.5~\citep{internvla_a15}
        & 83.1 & 55.1 & 86.9 & 96.4
        & 98.2 & \underline{95.6} & 85.2 & 84.8 \\

        Qwen-RobotManip-Context~\citep{qwen_robotmanip}
        & \underline{89.9} & 83.9 & 86.5 & \textbf{98.6}
        & \textbf{99.9} & \textbf{97.9} & 87.5 & \underline{91.4} \\

        \midrule
        \rowcolor{gray!10}
        \multicolumn{9}{c}{\textbf{WAM}} \\
        \midrule

        Fast-WAM~\citep{yuan2026fastwam}
        & 16.4 & 44.5 & 68.9 & 78.2
        & 53.7 & 37.7 & 60.7 & 51.5 \\

        OpenWAM-$\alpha$~\citep{openwam_alpha}
        & 33.8 & 76.1 & 88.0 & 97.0
        & 87.1 & 39.8 & 77.5 & 69.2 \\

        4D-WAM~\citep{wam4d}
        & 45.2 & 64.3 & 90.6 & 94.3
        & 57.7 & 69.1 & 79.2 & 71.0 \\

        ST-WAM~\citep{st_wam}
        & 55.4 & 60.1 & 79.3 & 93.0
        & 74.2 & 79.5 & 74.3 & 72.8 \\

        Faster-WAM~\citep{faster_wam_depth}
        & 67.9 & 49.0 & \underline{92.1} & 94.3
        & 57.0 & 82.3 & 82.7 & 75.0 \\

        JEPA-WAM~\citep{jepa_wam}
        & 79.2 & 59.2 & 68.2 & 93.3
        & 94.6 & 83.6 & 76.1 & 79.2 \\

        Cosmos-Policy~\citep{cosmos_policy}
        & 75.8 & 63.3 & 81.7 & 96.5
        & 88.9 & 92.7 & 82.2 & 82.2 \\

        ImageWAM~\citep{imagewam}
        & 80.8 & 50.3 & 91.4 & \underline{98.1}
        & 85.5 & 93.8 & 80.5 & 83.1 \\

        ABot-M0.5~\citep{abot_m05}
        & 70.5 & \underline{87.4} & 88.6 & 94.0
        & 89.7 & 75.5 & 85.2 & 83.4 \\

        Being-H0.7~\citep{being_h07}
        & 82.0 & 59.0 & 82.8 & 97.8
        & 90.0 & 93.5 & \textbf{88.5} & 84.8 \\

        \midrule

        \rowcolor{gray!12}
        \textbf{\modelname}
        & \textbf{90.6}
        & \textbf{91.1}
        & \textbf{92.9}
        & 95.8
        & \underline{99.4}
        & 94.0
        & \underline{88.4}
        & \textbf{92.8} \\

        \bottomrule
    \end{tabular}
}
\end{table}

\subsubsection{RoboTwin 2.0}
Under the Clean2Random protocol, models are trained on clean demonstrations only and evaluated on both Clean2Clean and Clean2Random settings. As shown in~\Cref{tab:main_robotwin}, \modelname achieves the best performance in both settings, reaching 90.0\% on Clean2Clean and 71.9\% on Clean2Random, with an overall success rate of 81.0\%. Compared with Qwen-RobotManip-Context~\citep{qwen_robotmanip}, the strongest VLA baseline, \modelname improves Clean2Random by 2.5 percentage points and the overall score by 3.9 points. The advantage over existing WAMs is larger: compared with OpenWAM-$\alpha$~\citep{openwam_alpha}, \modelname improves Clean2Random from 48.7\% to 71.9\%, while also slightly improving Clean2Clean from 89.4\% to 90.0\%. These results show that the improvement on randomized environments is achieved without sacrificing performance on the original clean distribution.

\begin{table}[!t]
    \centering
    \caption{
        Evaluation results on RoboTwin 2.0~\citep{robotwin2} Clean2Random under clean-only training.
        Success rates (\%) are reported on Clean2Clean and
        Clean2Random.
    }
    \label{tab:main_robotwin}

    \small
    \setlength{\tabcolsep}{12pt}
    \renewcommand{\arraystretch}{1.08}

\resizebox{\linewidth}{!}{%
\begin{tabular}{lccc}
    \toprule
    Method
    & Clean2Clean $\uparrow$
    & Clean2Random $\uparrow$
    & Overall $\uparrow$ \\
    \midrule

    \rowcolor{gray!10}
    \multicolumn{4}{c}{\textbf{VLA}} \\
    \midrule

    GR00T-N1.7~\citep{groot_n17}
    & 43.6
    & 20.7
    & 32.2 \\

    StarVLA~\citep{starvla}
    & 58.1
    & 10.6
    & 34.4 \\

    X-VLA~\citep{x_vla}
    & 68.0
    & 20.9
    & 44.5 \\

    Spatial Forcing~\citep{spatial_forcing}
    & 77.2
    & 26.7
    & 52.0 \\

    ABot-M0~\citep{abot_m0}
    & 70.7
    & 36.0
    & 53.4 \\

    $\pi_{0.5}$~\citep{pi05}
    & 73.1
    & 47.9
    & 60.5 \\

    GigaBrain-0.7~\citep{gigabrain07}
    & 66.8
    & 67.9
    & 67.4 \\

    Qwen-RobotManip-Context~\citep{qwen_robotmanip}
    & 84.7
    & \underline{69.4}
    & \underline{77.1} \\

    \midrule
    \rowcolor{gray!10}
    \multicolumn{4}{c}{\textbf{WAM}} \\
    \midrule

    AHA-WAM~\citep{aha_wam}
    & 64.3
    & 3.2
    & 33.8 \\

    Fast-WAM~\citep{yuan2026fastwam}
    & 77.8
    & 1.9
    & 39.9 \\

    X-WAM~\citep{x_wam}
    & 70.0
    & 25.8
    & 47.9 \\

    4D-WAM~\citep{wam4d}
    & 81.5
    & 41.8
    & 61.7 \\

    OpenWAM-$\alpha$~\citep{openwam_alpha}
    & \underline{89.4}
    & 48.7
    & 69.0 \\

    \midrule

    \rowcolor{gray!12}
    \textbf{\modelname}
    & \textbf{90.0}
    & \textbf{71.9}
    & \textbf{81.0} \\

    \bottomrule
\end{tabular}
}
\end{table}

\subsubsection{EBench}

We further evaluate \modelname on EBench~\citep{ebench}, which covers precision-sensitive Table Top tasks, mobile pick-and-place tasks, and multi-stage Long Horizon manipulation. As shown in~\Cref{tab:main_ebench}, \modelname achieves an overall success rate of 49.2\% and the highest overall score of 66.0, demonstrating competitive performance across different manipulation settings.

The advantage of \modelname is particularly clear on Long Horizon tasks, where it achieves 49.4\% success rate and a score of 76.5, both the highest among the compared methods. Since these tasks require multiple dependent operations over extended interaction sequences, the results indicate that \modelname can effectively maintain task progress and execution consistency over long-horizon manipulation. Together with its strong performance on Simple PnP, these results show that \modelname generalizes well from basic mobile manipulation to more temporally extended behaviors.

\subsubsection{RoboDojo}
As shown in~\Cref{tab:main_multicategory_eval}, \modelname achieves highly competitive overall performance, reaching an average SR of 23.91\% and a score of 30.77. Compared with DM0.5~\citep{dm05}, a strong VLA baseline in terms of average performance, \modelname improves the average SR by 4.57 percentage points and the score by 5.87 points. The improvement over existing WAMs is more substantial: compared with OpenWAM-$\alpha$~\citep{openwam_alpha}, \modelname increases the average SR from 11.92\% to 23.91\% and the score from 17.18 to 30.77. In individual categories, \modelname shows particularly strong performance on Gen-Std and Precision, reaching SRs of 33.78\% and 23.25\%, respectively, while also achieving a Long-Horizon score of 46.29. It remains competitive on Gen-Rand and Open tasks, where GPT-6 Astra~\citep{zhang2026unexpectedrobotpolicyearly} shows stronger performance. Overall, \modelname also exceeds GPT-6 Astra in average SR and score, reaching 23.91\% and 30.77 compared with 22.48\% and 28.97, respectively.

\begin{table}[t]
    \centering
    \caption{
        Evaluation results on EBench~\citep{ebench}
    }
    \label{tab:main_ebench}

    \small
    \setlength{\tabcolsep}{12pt}
    \renewcommand{\arraystretch}{1.08}

    \resizebox{\linewidth}{!}{%
    \begin{tabular}{l*{8}{c}}
        \toprule
        \multirow{2}{*}{Method}
        & \multicolumn{2}{c}{Table Top}
        & \multicolumn{2}{c}{Simple PnP}
        & \multicolumn{2}{c}{Long Horizon}
        & \multicolumn{2}{c}{Overall} \\
        \cmidrule(lr){2-3}
        \cmidrule(lr){4-5}
        \cmidrule(lr){6-7}
        \cmidrule(lr){8-9}
        & SR $\uparrow$ & Score $\uparrow$
        & SR $\uparrow$ & Score $\uparrow$
        & SR $\uparrow$ & Score $\uparrow$
        & SR $\uparrow$ & Score $\uparrow$ \\
        \midrule

        \rowcolor{gray!10}
        \multicolumn{9}{c}{\textbf{VLA}} \\
        \midrule

        StarVLA-OFT~\citep{starvla}
        & -- & --
        & -- & --
        & -- & --
        & 0.0 & 0.2 \\

        $\pi_0$~\citep{pi0}
        & 15.7 & 30.0
        & 35.0 & 39.0
        & 17.0 & 41.0
        & 23.6 & 37.0 \\

        X-VLA~\citep{x_vla}
        & 8.6 & 24.0
        & 50.0 & 54.0
        & 6.2 & 25.0
        & 23.7 & 36.0 \\

        InternVLA-A1~\citep{internvla_a1}
        & 4.3 & 11.0
        & 43.0 & 47.0
        & 17.9 & 46.0
        & 23.9 & 36.0 \\

        $\pi_{0.5}$~\citep{pi05}
        & 12.9 & 32.0
        & 45.0 & 50.0
        & 18.1 & 39.0
        & 27.1 & 41.0 \\

        GigaBrain-0.7~\citep{gigabrain07}
        & -- & --
        & -- & --
        & -- & --
        & 33.3 & 46.0 \\

        Qwen-RobotManip~\citep{qwen_robotmanip}
        & \textbf{50.0} & \textbf{70.0}
        & 56.5 & 60.0
        & 29.9 & 55.0
        & 45.6 & 60.0 \\

        \midrule
        \rowcolor{gray!10}
        \multicolumn{9}{c}{\textbf{WAM}} \\
        \midrule

        Fast-WAM~\citep{yuan2026fastwam}
        & -- & --
        & -- & --
        & -- & --
        & 4.7 & 7.6 \\

        OpenWAM-$\alpha$~\citep{openwam_alpha}
        & 30.0 & 44.2
        & \textbf{67.5} & \textbf{72.0}
        & \underline{44.3} & \underline{72.6}
        & \textbf{49.4} & \underline{64.7} \\

        \midrule

        \rowcolor{gray!12}
        \textbf{\modelname}
        & \underline{33.6} & \underline{55.2}
        & \underline{60.0} & \underline{64.3}
        & \textbf{49.4} & \textbf{76.5}
        & \underline{49.2} & \textbf{66.0} \\

        \bottomrule
    \end{tabular}%
    }
\end{table}

\begin{table*}[t]
    \centering
    \caption{
        Evaluation results on RoboDojo~\citep{robodojo} across six evaluation categories.
    }
    \label{tab:main_multicategory_eval}

    \scriptsize
    \setlength{\tabcolsep}{1.8pt}
    \renewcommand{\arraystretch}{1.06}

    \resizebox{\textwidth}{!}{%
    \begin{tabular}{l*{14}{c}}
        \toprule
        \multirow{2}{*}{Method}
        & \multicolumn{2}{c}{Gen-Std}
        & \multicolumn{2}{c}{Gen-Rand}
        & \multicolumn{2}{c}{Precision}
        & \multicolumn{2}{c}{Long-Horizon}
        & \multicolumn{2}{c}{Memory}
        & \multicolumn{2}{c}{Open}
        & \multicolumn{2}{c}{Avg} \\
        \cmidrule(lr){2-3}
        \cmidrule(lr){4-5}
        \cmidrule(lr){6-7}
        \cmidrule(lr){8-9}
        \cmidrule(lr){10-11}
        \cmidrule(lr){12-13}
        \cmidrule(lr){14-15}
        & SR $\uparrow$ & Score $\uparrow$
        & SR $\uparrow$ & Score $\uparrow$
        & SR $\uparrow$ & Score $\uparrow$
        & SR $\uparrow$ & Score $\uparrow$
        & SR $\uparrow$ & Score $\uparrow$
        & SR $\uparrow$ & Score $\uparrow$
        & SR $\uparrow$ & Score $\uparrow$ \\
        \midrule

        \rowcolor{gray!10}
        \multicolumn{15}{c}{\textbf{VLA}} \\
        \midrule

        StarVLA-$\alpha$~\citep{starvla}
        & 5.00 & 7.54
        & 0.00 & 0.33
        & 4.33 & 9.90
        & 6.50 & 14.15
        & 2.44 & 3.34
        & 0.58 & 0.68
        & 3.24 & 6.40 \\

        X-VLA~\citep{x_vla}
        & 12.00 & 17.90
        & 1.00 & 3.04
        & 12.00 & 18.32
        & 9.75 & 16.53
        & 3.56 & 4.76
        & 0.50 & 0.55
        & 6.52 & 10.13 \\

        $\pi_{0.5}$~\citep{pi05}
        & 15.00 & 20.93
        & 1.00 & 5.82
        & 5.50 & 12.40
        & 14.67 & 23.54
        & 4.56 & 5.78
        & 1.67 & 1.98
        & 6.91 & 11.41 \\

        Spatial Forcing~\citep{spatial_forcing}
        & 15.00 & 21.25
        & 4.00 & 6.98
        & 10.58 & 17.33
        & 14.58 & 23.26
        & 4.11 & 5.43
        & 1.58 & 1.78
        & 8.04 & 12.38 \\

        Hy-Embodied-0.5-VLA~\citep{zhang2026hy}
        & 17.00 & 21.98
        & 0.00 & 1.57
        & 8.00 & 13.81
        & 14.92 & 25.74
        & 12.11 & 13.37
        & 0.58 & 0.65
        & 8.80 & 13.07 \\

        Xiaomi-Robotics-1~\citep{team2026xiaomi}
        & 28.00 & \underline{35.65}
        & 6.00 & 11.44
        & 18.83 & 26.69
        & 23.67 & 38.39
        & 6.56 & 7.81
        & 3.58 & 3.94
        & 13.93 & 20.07 \\

        Galaxea G0.5~\citep{galaxea_g05}
        & 20.00 & 26.74
        & 6.00 & 11.16
        & \underline{20.42} & \underline{28.25}
        & \textbf{32.25} & \underline{44.12}
        & 7.33 & 8.61
        & 1.58 & 1.73
        & 14.88 & 20.23 \\

        DM0.5~\citep{dm05}
        & 18.00 & 23.49
        & 4.00 & 8.06
        & 16.75 & 24.82
        & 19.50 & 33.70
        & \textbf{47.44} & \textbf{47.74}
        & 2.08 & 2.43
        & 19.34 & 24.90 \\

        \midrule
        \rowcolor{gray!10}
        \multicolumn{15}{c}{\textbf{LLM as Policy}} \\
        \midrule

        GPT-6 Astra~\citep{zhang2026unexpectedrobotpolicyearly}
        & \underline{32.67} & 35.32
        & \textbf{28.33} & \textbf{31.40}
        & 4.00 & 12.65
        & 8.25 & 21.45
        & \underline{38.67} & \underline{43.04}
        & \textbf{31.00} & \textbf{34.36}
        & \underline{22.48} & \underline{28.97} \\

        \midrule
        \rowcolor{gray!10}
        \multicolumn{15}{c}{\textbf{WAM}} \\
        \midrule

        Fast-WAM~\citep{yuan2026fastwam}
        & 2.00 & 4.33
        & 0.00 & 0.34
        & 0.00 & 1.96
        & 5.17 & 9.14
        & 3.44 & 3.55
        & 0.42 & 0.42
        & 2.03 & 3.48 \\

        AHA-WAM~\citep{aha_wam}
        & 6.00 & 10.32
        & 0.00 & 1.26
        & 2.42 & 5.86
        & 2.67 & 8.61
        & 2.78 & 2.97
        & 0.83 & 0.88
        & 2.39 & 4.82 \\

        GigaWorld-Policy~\citep{gigaworld_policy}
        & 6.00 & 10.28
        & 0.00 & 0.41
        & 1.83 & 6.15
        & 8.92 & 15.51
        & 2.22 & 3.46
        & 0.50 & 0.54
        & 3.27 & 6.20 \\

        X-WAM~\citep{x_wam}
        & 5.00 & 11.24
        & 1.00 & 3.54
        & 1.83 & 6.72
        & 9.08 & 17.47
        & 4.67 & 6.32
        & 0.25 & 0.57
        & 3.83 & 7.69 \\

        OpenWAM-$\alpha$~\citep{openwam_alpha}
        & 25.56 & 33.16
        & 4.11 & 8.26
        & 9.25 & 18.45
        & 25.33 & 34.93
        & 9.11 & 10.41
        & 1.08 & 1.41
        & 11.92 & 17.18 \\

        \midrule

        \rowcolor{gray!12}
        \textbf{\modelname}
        & \textbf{33.78} & \textbf{40.98}
        & \underline{11.78} & \underline{19.19}
        & \textbf{23.25} & \textbf{31.98}
        & \underline{29.33} & \textbf{46.29}
        & 34.00 & 34.67
        & \underline{10.17} & \underline{10.84}
        & \textbf{23.91} & \textbf{30.77} \\

        \bottomrule
    \end{tabular}%
    }
\end{table*}

\subsection{Real-Robot Experiments}
\label{sec:real-robot-experiments}

\begin{figure*}[t]
  \centering
  \includegraphics[width=\textwidth]{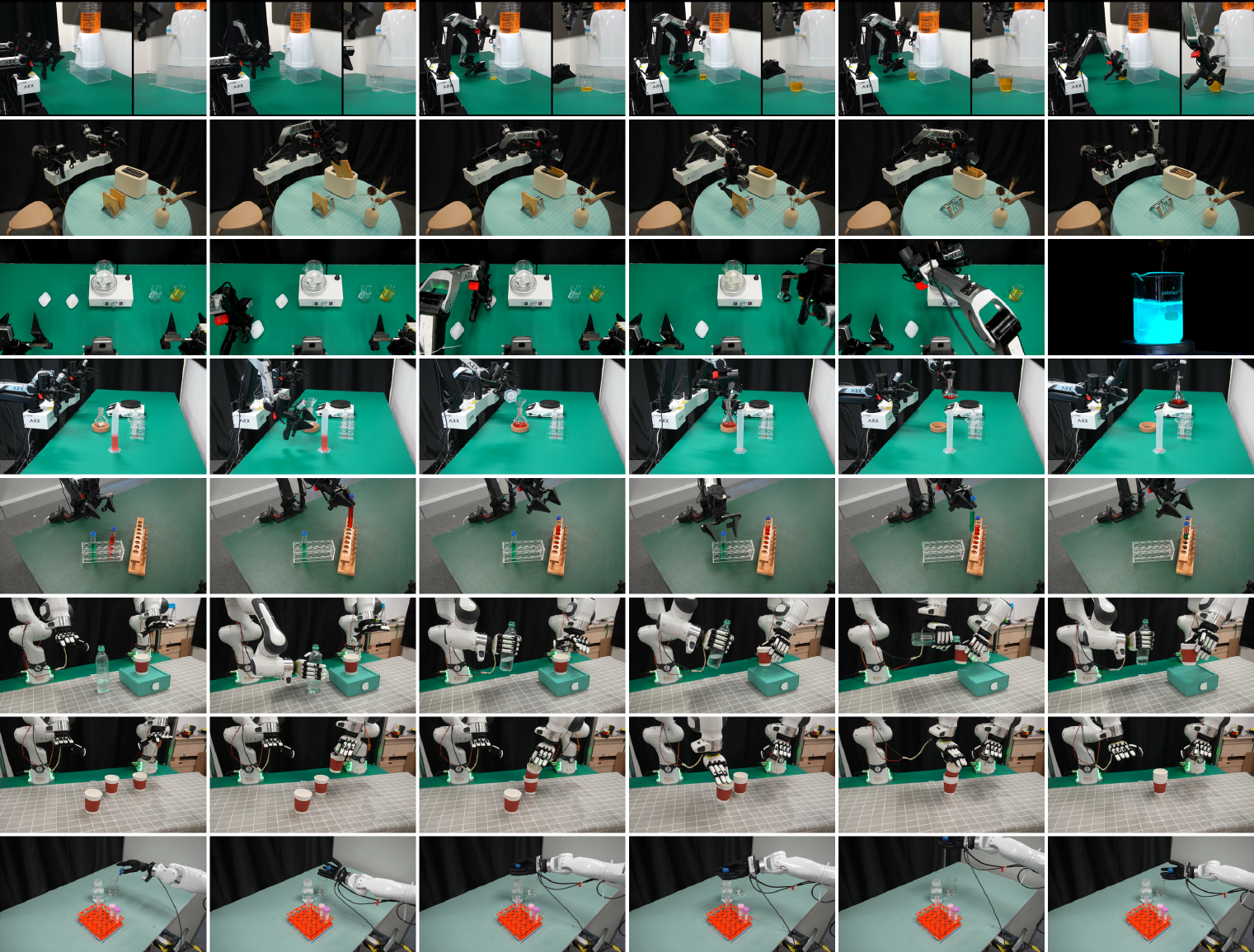}
  \caption{\textbf{Real-robot task execution.}
  Selected video frames illustrate eight tasks, with each row ordered
  chronologically from left to right. From top to bottom:
  \textit{get a drink}, \textit{toast bread},
  \textit{Luminol reaction}, \textit{MOF experiment},
  \textit{place tube}, \textit{pour water}, \textit{stack cups},
  and \textit{use dropper}.
  The final frame of the Luminol sequence shows a close-up of the
  reaction outcome.}
  \label{fig:real-robot-task-sequences}
\end{figure*}

\paragraph{Tasks and evaluation procedure.}
We evaluate eleven tasks across four real-robot platforms: AC-One,
Arx5, bimanual Franka+XHand, and TianJi Marvin+Wuji Hand. 
On AC-One, the tasks comprise \textit{get a drink}, \textit{toast bread},
\textit{Luminol reaction}, and \textit{MOF experiment}. \textit{Get a drink} episodes
place a cup under the dispenser and fill it to the marked level before moving
it to the right side of the dispenser; \textit{toast bread} episodes place two bread slices
into the toaster in sequence and press its switch; \textit{Luminol reaction} episodes add NaOH,
luminol, hydrogen peroxide, and potassium ferricyanide in sequence and mix the
reaction; \textit{MOF experiment} episodes pour a measured solution through a funnel into a flask,
return the funnel, and place and seal the flask on the stirrer. On Franka+XHand,
we evaluate \textit{pour water}, \textit{place fruit into box}, and
\textit{stack cups}. These tasks respectively pour water from a bottle into a disposable
paper cup, place the language-specified fruit into a box, and stack disposable
cups. On Arx5, \textit{magnetic stirrer} places a magnetic stir bar into a
beaker and then places the beaker on the magnetic stirrer;
\textit{place tube} transfers a test tube to a wooden test-tube rack.
On TianJi Marvin+Wuji Hand, \textit{use dropper} picks up a rubber-bulb
dropper, draws liquid from a reagent bottle, and transfers it into a
test tube held in a rack. The \textit{make a sandwich} task places a
bread slice from the bread rack onto a plate, sequentially layers
cheese, egg, and lettuce on top, and adds a second bread slice
to complete the sandwich. On these four platforms, each rollout starts from the robot's home
pose with a text instruction and runs under asynchronous training-time RTC
(\Cref{sec:real-robot-deployment}), which achieves real-time deployment. \Cref{fig:real-robot-task-sequences} illustrates representative
execution sequences for eight of these tasks.

\paragraph{Effect of pretraining on real-robot performance.}
We compare models with and without our pretraining stage after task-specific
post-training on two AC-One tasks, using 20 evaluation trials per task and
condition. As shown in \Cref{tab:real-pretraining-comparison}, pretraining
increases success rates from 20\% to 95\% on \textit{toast bread} and
from 0\% to 95\% on \textit{Luminol reaction}, supporting its benefit for
these downstream real-robot tasks. 

\begin{figure}[t]
  \centering
  \includegraphics[width=0.6725\linewidth]{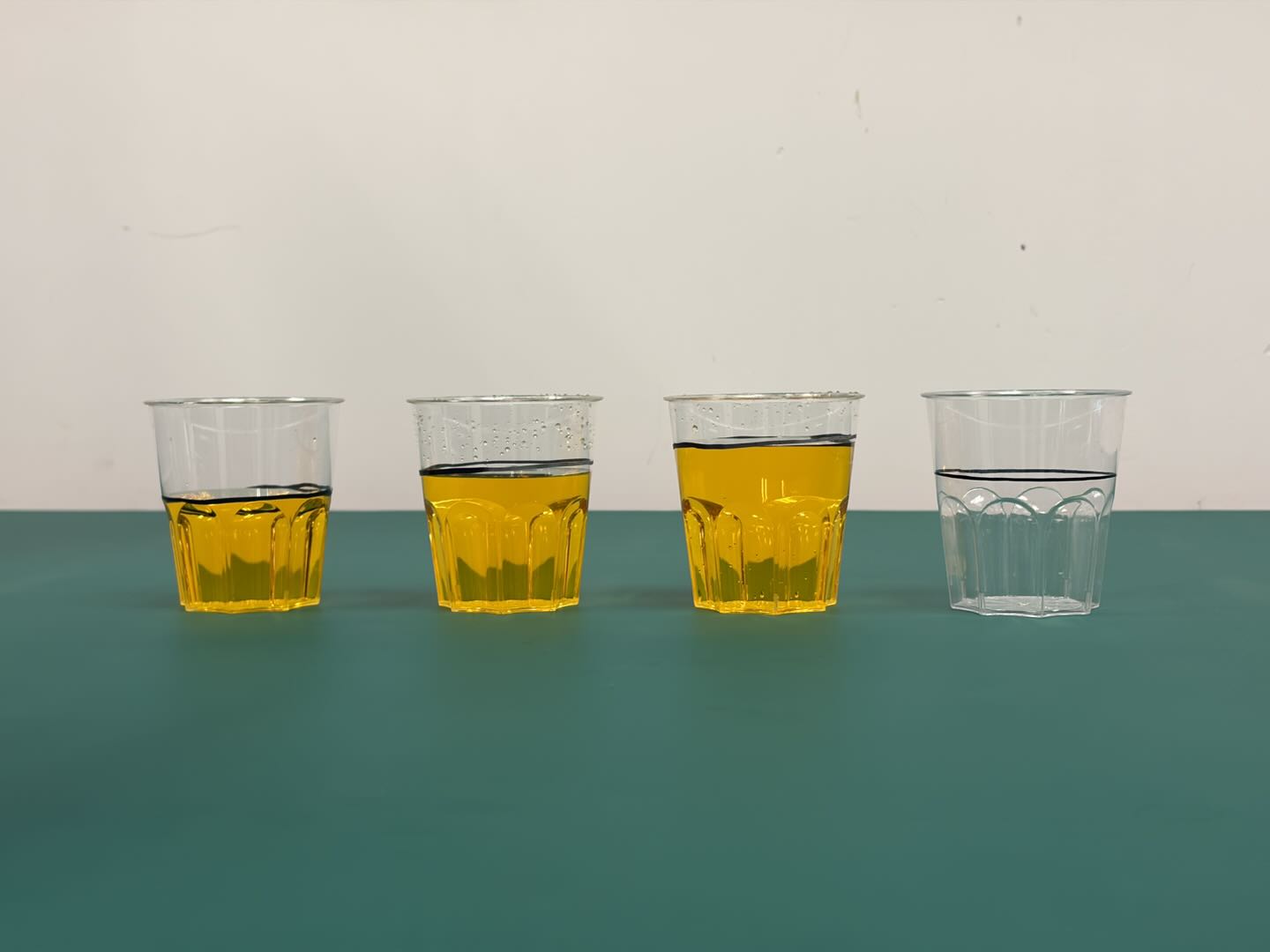}
  \caption{\textbf{Generalization to different target water levels.}
  For the AC-One \textit{get a drink} task, the rightmost cup
  shows the marked target level used throughout post-training data
  collection. The three cups on the left show the outcomes of three
  successful executions at different marked target levels.
  Black markings indicate the target levels.}
  \label{fig:water-level-generalization}
\end{figure}

\begin{table}[t]
  \centering
  \caption{Real-robot performance with and without our pretraining stage,
  followed by task-specific post-training. Each entry reports successful
  trials out of 20 and the corresponding success rate.}
  \label{tab:real-pretraining-comparison}
  \small
  \setlength{\tabcolsep}{8pt}
  \begin{tabular}{lcc}
    \toprule
    Task & Without pretraining & With pretraining \\
    \midrule
    Toast bread & 4/20 (20\%) & 19/20 (95\%) \\
    Luminol reaction & 0/20 (0\%) & 19/20 (95\%) \\
    \bottomrule
  \end{tabular}
\end{table}

\paragraph{Generalization across physical conditions.}
Beyond the training conditions, the post-trained policy exhibits
qualitative generalization capabilities to unseen target water levels.
For the AC-One \textit{get a drink} task, all 137 post-training
demonstrations use the same marked target water level.
At deployment, the policy successfully fills cups to different marked
target levels, indicating that its behavior is not restricted to the
single target level represented in the demonstrations. \Cref{fig:water-level-generalization} shows three successful
executions alongside the training target-level reference.

\paragraph{Cross-embodiment adaptation.}
The same pretrained checkpoint is post-trained separately for the \textbf{gripper-based} and
\textbf{dexterous hands} interfaces. For grippers, AC-One uses 14-dimensional bimanual
absolute joint targets and Arx5 utilizes 6-dimensional end-effector poses plus a 1-dimensional gripper. For dexterous hands, Franka+XHand uses 6-dimensional end-effector
commands per Franka arm together with 12-dimensional dexterous hand commands
per hand, whereas TianJi Marvin+Wuji Hand is controlled by 7-dimensional arm joint action and 20-dimensional hand action. All four settings therefore require different post-training action
adapters while sharing the same pretrained model.

\begin{figure*}[t]
  \centering
  \includegraphics[width=\textwidth]{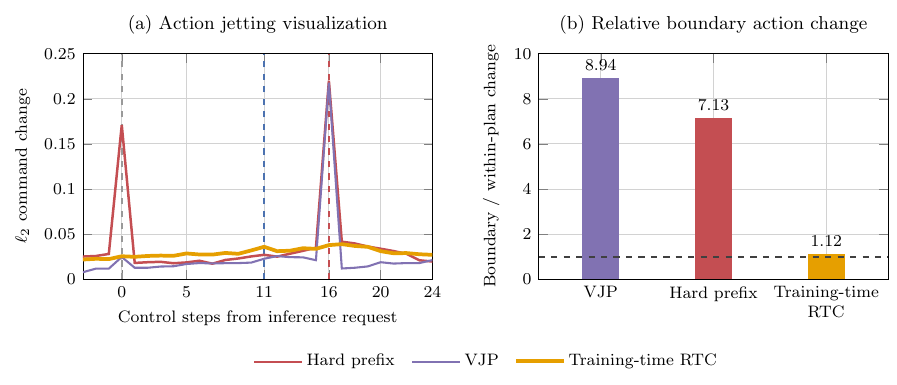}
  \caption{\textbf{Action jetting visualization.}
  (a) $\ell_2$ changes between consecutive published 14-dimensional action
  commands, aligned to the inference request at $t=0$, with $t$ measured
  in published control steps. Hard prefix and training-time RTC are shown
  as mean curves with chunk switches at $t=10$; VJP is shown as an
  individual trace with its chunk switch at $t=16$.
  The gray dashed line at $t=0$ also marks the previous hard-prefix plan's
  prefix-to-suffix boundary. The blue dashed line at $t=11$ marks the
  training-time prefix-to-suffix boundary. The red dashed line at $t=16$
  marks both the hard-prefix prefix-to-suffix boundary and the VJP chunk
  boundary.
  (b) Mean boundary command change divided by the mean consecutive-command
  change within the executed new-plan segments, excluding the evaluated
  boundary. The evaluated boundary is the chunk boundary for VJP and the
  prefix-to-suffix boundary for hard prefix and training-time RTC.
  The dashed horizontal line denotes a ratio of one.}
  \label{fig:rtc-action-jetting}
\end{figure*}

\paragraph{RTC continuity comparison. }
The three RTC strategies exhibit distinct continuity behaviors in the
recorded AC-One runs (\Cref{fig:rtc-action-jetting}).
In the illustrated trace, VJP~\citep{rtc_inference_time_arxiv2506_07339}
shows an abrupt change in the published action commands at the chunk
boundary, with a boundary-to-within-plan change ratio of $8.94$.
Hard prefix~\citep{motubrain2026realtime} preserves the committed actions
at the chunk boundary, but introduces a sharp jetting event at the
prefix-to-suffix boundary, where the model switches from the committed
prefix to freely generated actions. Its corresponding ratio is $7.13$.
Training-time conditioning~\citep{rtc_training_time_arxiv2512_05964}
keeps changes at the prefix-to-suffix boundary comparable to ordinary
within-plan changes, with a ratio of $1.12$.
These diagnostic traces support our use of training-time prefix
conditioning for continuous asynchronous execution.

\subsection{Ablation}

\subsubsection{Component-Wise Ablations}
We conduct cumulative ablations on LIBERO-Plus to investigate the contribution of the main components in \modelname. All configurations are trained independently from scratch using the same training data, optimization recipe, and evaluation protocol. Therefore, each row in \Cref{tab:cumulative-ablation} corresponds to a separate training run rather than continued training from the preceding configuration. Starting from the baseline, we progressively introduce Sparse Memory Context (SMC), VLM, Causal Imprint (CI), representation alignment, and 4D-aware representation distillation.

\begin{table}[t]
  \centering
  \caption{Cumulative ablation study on LIBERO-Plus~\citep{libero_plus}. All variants are independently trained from scratch under the same training and evaluation settings. Starting from the baseline, we progressively introduce Sparse Memory Context (SMC), vision-language conditioning, Causal Imprint (CI), the additional CI alignment objective $\mathcal{L}_{\mathrm{align}}$, and 4D-aware representation distillation $\mathcal{L}_{\mathrm{4D}}$. Success rate denotes the overall success rate (\%). }
  \label{tab:cumulative-ablation}

  \small
  \renewcommand{\arraystretch}{1.15}

  \begin{tabular}{lc}
    \toprule
    Method & Success rate (\%) $\uparrow$ \\
    \midrule
    Baseline & 49.59 \\
    Baseline + SMC & 53.47 \\
    Baseline + SMC + Qwen3.5-2B & 60.37 \\
    Baseline + SMC + RynnBrain1.1-2B & 69.08 \\
    Baseline + SMC + RynnBrain1.1-2B + CI & 70.80 \\
    Baseline + SMC + RynnBrain1.1-2B + CI + $\mathcal{L}_{\mathrm{align}}$ & 76.45 \\
    Baseline + SMC + RynnBrain1.1-2B + CI + $\mathcal{L}_{\mathrm{align}}$ + $\mathcal{L}_{\mathrm{4D}}$ & \textbf{78.37} \\
    \bottomrule
  \end{tabular}
\end{table}

\paragraph{Sparse Memory Context.} We first investigate whether lightweight temporal context benefits action prediction. Adding SMC improves the success rate from 49.59\% to 53.47\%, a gain of 3.88 percentage points. SMC augments the current observation with an episode-level anchor and a recent observation preceding the previous action chunk, providing both coarse task context and short-term execution history. The improvement indicates that a single current observation is insufficient to fully capture the interaction state, while a small number of carefully selected historical observations already provides useful temporal information without maintaining a dense visual history. 

\paragraph{VLM.} We next investigate the effect of the vision-language representation supplied to the action expert. With SMC fixed, introducing Qwen3.5-2B improves the success rate from 53.47\% to 60.37\%, showing the benefit of jointly reasoning over the current visual scene and task instruction. Replacing Qwen3.5-2B with RynnBrain1.1-2B further increases the success rate to 69.08\%, corresponding to an additional gain of 8.71 percentage points. Since all variants use the same training and evaluation settings, this comparison shows that the representation provided by the VLM has a strong effect on downstream manipulation performance. We therefore adopt RynnBrain1.1-2B in the remaining experiments. 

\paragraph{Causal Imprint.} We then investigate Causal Imprint, which is designed to encode information about how the observed scene is likely to evolve while using only observations available to the policy at inference time. CI is trained with its latent-difference objective $\mathcal{L}_{\Delta}$, which directly supervises the imprint representation using changes between consecutive future video latents. Introducing CI improves the success rate from 69.08\% to 70.80\%. We further introduce the representation alignment objective $\mathcal{L}_{\mathrm{align}}$, which aligns Causal Imprint tokens with spatially corresponding representations from the future video features. This increases the success rate substantially from 70.80\% to 76.45\%, a gain of 5.65 percentage points. The two objectives provide complementary signals: $\mathcal{L}_{\Delta}$ directly describes the temporal changes along the future trajectory, whereas $\mathcal{L}_{\mathrm{align}}$ transfers the richer semantic and temporal representation learned by the video expert. Together, they encourage Causal Imprint to capture not only low-level visual changes but also higher-level information about the subsequent evolution of the scene. Importantly, the future observations are used only to construct supervision during training. The Causal Imprint tokens themselves cannot directly attend to the realized future, and the action expert therefore receives no privileged future observation during either training or inference. This allows future trajectories to shape the learned representation without requiring explicit future-video generation for online action prediction.

\paragraph{4D-aware representation distillation. }

As shown in~\Cref{tab:cumulative-ablation}, adding $\mathcal{L}_{\mathrm{4D}}$ improves the overall success rate on LIBERO-Plus from 76.45\% to 78.37\%, a gain of 1.92 percentage points. We further investigate the teacher choice and action feature injection in~\Cref{tab:4d-distillation-ablation}. Distillation from Track4World~\citep{lu2026track4world} achieves the highest success rate of 78.37\%, compared with 76.15\% for CoWTracker~\citep{lai2026cowtracker} and 73.63\% for Pi3X~\citep{wang2025pi3}. Additionally, injecting the student descriptor into the action expert yields a slightly lower success rate of 77.78\%. We therefore adopt auxiliary distillation of the video expert without action injection, allowing the student branch to be removed at inference with no additional policy inference cost.

\begin{table}[t]
    \centering
    \caption{
      Ablation of distillation design on LIBERO-Plus~\citep{libero_plus}.
      All variants include SMC, RynnBrain1.1-2B~\citep{dang2026rynnbrainopenembodiedfoundation}, CI, and
      $\mathcal{L}_{\mathrm{align}}$.
      Action injection indicates whether the student descriptor
      is additionally fed into the action expert.
    }
    \label{tab:4d-distillation-ablation}
    \small
    \setlength{\tabcolsep}{5pt}
    \renewcommand{\arraystretch}{1.15}
    \begin{tabular}{lcc}
      \toprule
      Teacher & Action injection & Success rate (\%) $\uparrow$ \\
      \midrule
      None        & No  & 76.45 \\
      CoWTracker  & No  & 76.15 \\
      Pi3X        & No  & 73.63 \\
      Track4World & Yes & 77.78 \\
      Track4World & No  & \textbf{78.37} \\
      \bottomrule
    \end{tabular}
  \end{table}

\subsubsection{Data Source Ablations }
We investigate how different data sources contribute to model performance. We consider four types of data, including raw robot demonstrations, raw human egocentric videos (Ego), human demonstrations converted through our Ego2Robot pipeline (Ego2Robot), and UMI data. All variants use the same \modelname architecture and are pretrained using data from different sources. The subsequent training and evaluation pipelines are kept identical across all settings, allowing us to study the impact of data composition under a controlled setup.

Across the evaluated settings, as shown in \Cref{tab:ego2robot-data-ablation} and \Cref{tab:ego2robot-robotwin-ablation}, robot demonstration data provides the strongest and most consistent benefit. On LIBERO-Plus, robot pretraining improves the overall success rate from 78.59\% to 83.73\%. The effect is particularly clear on RoboTwin 2.0 under the Clean2Random setting, where the success rate increases from 4.38\% to 32.34\%, indicating a substantial gain in robustness to visual and environmental variations. In comparison, raw egocentric video brings only a modest improvement on the current benchmarks, increasing LIBERO-Plus performance to 80.45\% and RoboTwin Clean2Random from 2.80\% to 3.13\%.  Ego2Robot data yields a larger gain, reaching 81.86\% on LIBERO-Plus and 6.66\% on RoboTwin Clean2Random. UMI data provides an even stronger improvement, achieving 83.24\% on LIBERO-Plus and increasing RoboTwin Clean2Random from 2.80\% to 13.90\%. Overall, these results suggest that human data become more effective when their observation and action spaces are better aligned with those of robots.

The results are specific to our current data scale and evaluation setting, with both LIBERO-Plus and RoboTwin primarily focusing on gripper-based manipulation. Under this setting, raw egocentric video provides limited gains in task success rate, but this does not rule out stronger benefits at a larger scale or on tasks that are more closely aligned with human hand-object interaction. In particular, human demonstrations may be more useful for dexterous manipulation, where fine-grained contact patterns and hand motions are less well covered by standard robot datasets. Moreover, our current evaluation only measures downstream success rate. Therefore, the limited improvement from Ego data should not be interpreted as evidence that it does not improve the learned representation, since such effects may not be fully reflected by success rate on the current benchmarks.

\begin{table}[H]
    \centering
\caption{Pretraining data-source ablation on LIBERO-Plus~\citep{libero_plus}.
    The baseline uses no pretraining, whereas Robot, Ego, Ego2Robot, and UMI
    are pretrained exclusively on their respective data sources.
    All values are success rates (\%).}
    \label{tab:ego2robot-data-ablation}
    \small
    \setlength{\tabcolsep}{4.5pt}
    \renewcommand{\arraystretch}{1.08}
    \begin{tabular}{lcccccccc}
        \toprule
        Method & Camera & Robot & Language & Light & Background & Noise & Layout & Overall $\uparrow$ \\
        \midrule
        Baseline
        & 60.10 & 76.77 & 88.68 & 95.01
        & 81.23 & 75.58 & 78.69 & 78.59 \\
        Robot
        & \textbf{74.36} & 68.39 & 93.69 & 96.67
        & 81.97 & \textbf{93.07} & 80.85 & \textbf{83.73} \\
        Ego
        & 65.60 & 74.26 & 91.93 & 94.83
        & \textbf{84.20} & 79.01 & 78.82 & 80.45 \\
        Ego2Robot
        & 63.10 & 79.81 & 93.69 & 96.76
        & 82.62 & 82.70 & 79.15 & 81.86 \\
        UMI
        & 65.92 & \textbf{80.26} & \textbf{96.49} & \textbf{97.20}
        & 79.83 & 84.82 & \textbf{81.38} & 83.24 \\
        \bottomrule
    \end{tabular}
\end{table}

\begin{table}[H]
    \centering
\caption{Pretraining data-source ablation on RoboTwin 2.0~\citep{robotwin2}.
    Baseline-Joint and Baseline-EEF use no pretraining, whereas Robot-Joint,  Ego2Robot-Joint, Ego-EEF and UMI-EEF
    are pretrained exclusively on their respective data sources.
    All values are success rates (\%).}
    \label{tab:ego2robot-robotwin-ablation}
    \small
    \setlength{\tabcolsep}{12pt}
    \renewcommand{\arraystretch}{1.08}
    \begin{tabular}{lccc}
        \toprule
        Method & Clean2Clean $\uparrow$ & Clean2Random $\uparrow$ & Overall $\uparrow$ \\
        \midrule
        Baseline-Joint & 78.20 & 4.38 & 41.29 \\
        Robot-Joint & \textbf{81.36} & \textbf{32.34} & \textbf{56.85} \\
        Ego2Robot-Joint & 76.92 & 6.66 & 41.79 \\
        \midrule
        Baseline-EEF & 61.58 & 2.80 & 32.08 \\
        Ego-EEF & 61.49 & 3.13 & 32.31 \\
        UMI-EEF & \textbf{66.05} & \textbf{13.90} & \textbf{39.98} \\
        \bottomrule
    \end{tabular}
\end{table}

\subsection{Exploration of GPT-Guided Policy}

Recent work on GPT as an embodied policy has explored using GPT-6 Astra~\citep{zhang2026unexpectedrobotpolicyearly} together with $\pi_{0.5}$, where GPT monitors the execution of a VLA policy and provides corrective actions when necessary~\citep{su2026astra}. Their experiments on RoboDojo show that such test-time correction can improve the execution of a pretrained VLA policy~\citep{su2026astra}. Motivated by these results, we conduct a small-scale exploration to examine whether the same mechanism can also complement a World Action Model.

We select five RoboDojo tasks on which the standalone \modelname exhibits relatively low performance under seed 0, covering both semantic reasoning and manipulation challenges. As shown in~\Cref{tab:wam_astra_five_tasks}, \modelname obtains an average success rate of 10.80\% and an average score of 13.92 across these tasks. We then introduce GPT-6 Astra as an execution-time assistant that observes the interaction and provides end-effector (EEF) corrections when necessary. This increases the average success rate to 47.20\% and the average score to 52.00, corresponding to absolute gains of 36.40 percentage points in success rate and 38.08 score points, respectively. The improvement is particularly clear on ``classify objects'', where the success rate increases from 14\% to 80\%, and on ``general pickup'', from 36\% to 76\%. Tasks with zero standalone success also become partially solvable: ``solve equation'' and ``arrange largest number'' reach 20\% and 40\% success rates, respectively. These results suggest that GPT-based execution correction can provide complementary high-level reasoning and online adjustment even when the underlying policy is a world-action model rather than a conventional VLA.

We further observe that 
{\it 
the reasoning configuration of GPT-6 Astra can have a noticeable effect on correction quality, especially for tasks that require semantic understanding.}  For example, on ``classify objects by language'', using GPT-6 Astra with the xhigh reasoning setting together with \modelname successfully completes all five evaluated episodes. In contrast, the medium setting obtains only partial task scores and does not complete the full task in these trials. Although this comparison is based on a small number of episodes, it indicates that stronger test-time reasoning may be particularly useful when policy correction requires interpreting language instructions, identifying task-relevant objects, and deciding how the underlying policy should be adjusted.

\begin{table*}[!t]
\centering
\caption{Comparison of Ours and Ours+Astra on five RoboDojo~\citep{robodojo} tasks.}
\label{tab:wam_astra_five_tasks}

\scriptsize
\setlength{\tabcolsep}{3.0pt}
\renewcommand{\arraystretch}{1.06}

\resizebox{0.98\textwidth}{!}{%
\begin{tabular}{l*{8}{c}*{2}{w{c}{1.6cm}}*{2}{c}}
    \toprule
    \multirow{2}{*}{Method}
    & \multicolumn{2}{c}{classify objects}
    & \multicolumn{2}{c}{press by number}
    & \multicolumn{2}{c}{solve equation}
    & \multicolumn{2}{c}{general pickup}
    & \multicolumn{2}{c}{arrange largest number}
    & \multicolumn{2}{c}{Average} \\
    \cmidrule(lr){2-3}
    \cmidrule(lr){4-5}
    \cmidrule(lr){6-7}
    \cmidrule(lr){8-9}
    \cmidrule(lr){10-11}
    \cmidrule(lr){12-13}
    & SR $\uparrow$ & Score $\uparrow$
    & SR $\uparrow$ & Score $\uparrow$
    & SR $\uparrow$ & Score $\uparrow$
    & SR $\uparrow$ & Score $\uparrow$
    & SR $\uparrow$ & Score $\uparrow$
    & SR $\uparrow$ & Score $\uparrow$ \\
    \midrule

    Ours
    & 14.00 & 22.80
    & 4.00  & 4.00
    & 0.00  & 0.00
    & 36.00 & 36.00
    & 0.00  & 6.80
    & 10.80 & 13.92 \\

    \rowcolor{gray!12}
    \textbf{Ours+Astra}
    & \textbf{80.00} & \textbf{88.00}
    & \textbf{20.00} & \textbf{20.00}
    & \textbf{20.00} & \textbf{20.00}
    & \textbf{76.00} & \textbf{76.00}
    & \textbf{40.00} & \textbf{56.00}
    & \textbf{47.20} & \textbf{52.00} \\

    \bottomrule
\end{tabular}%
}
\end{table*}
\section{Conclusion}

\paragraph{Summary.}

We have presented \modelname, a world-action model that learns action-relevant dynamic representations by combining pretrained visual dynamics, task-conditioned scene semantics, and robot action generation within a directed Mixture-of-Transformers architecture.
A pretrained video expert is coupled with an action expert grounded in scene semantics from a frozen VLM, while sparse visual memory provides both episode-level context and recent interaction history.
Causal Imprint uses future supervision to capture changes relevant to subsequent interaction, and training-only 4D-aware distillation further introduces geometric and motion priors into the video representation.
Together, these designs allow predictive information learned during training to directly support action generation without requiring future-video rollout at inference.

Beyond the model, we have developed a scalable data, training, and deployment recipe built on over 20K hours of processed robot, UMI, egocentric, and Ego2Robot data.
Canonical state and action representations, systematic quality filtering, and temporal alignment enable heterogeneous pretraining followed by target-embodiment post-training.
Across multiple simulation benchmarks and real-robot platforms, the same pretrained checkpoint can be adapted to different task distributions, embodiments, and control interfaces, including both gripper-based and dexterous-hand manipulation.

We have further developed  training and inference infrastructure for efficient model iteration and online execution, and will release the associated code, checkpoints, training recipes, data-processing tools, and evaluation and deployment utilities.

\paragraph{Limitations.}
The current work does not exhaustively explore all components of the proposed training and deployment framework.
First, although egocentric human demonstrations constitute an important part of our pretraining mixture, our study of their contribution remains limited.
We have not yet systematically examined how different egocentric data sources, conversion strategies, scaling ratios, and supervision forms affect downstream robot performance.
Second, our investigation of agent-assisted control is still preliminary.
The current experiments only consider a limited corrective setting, and do not systematically study agent invocation policies, long-horizon planning, hierarchical decision making, or tighter integration between external agents and the world-action model.
We leave a broader study of egocentric data utilization and agent-enhanced control to future work.

\section{Team}

\begingroup
\renewcommand{\thefootnote}{\fnsymbol{footnote}}

\subsection*{Core Contributors}

\noindent\textbf{Real-Robot Data:} Zizun Li\footnote[1]{Equal contribution.}, Xingyu Miao\footnotemark[1], Xueyuan Wei

\noindent\textbf{Ego \& UMI Data:} Kaiwen Song\footnotemark[1], Tenghui Wang\footnotemark[1], Yuping He

\noindent\textbf{Model, Pretraining \& Infrastructure:} Xingyu Miao\footnotemark[1], Zizun Li\footnotemark[1], Hanxue Zhang, Yating Wang

\noindent\textbf{Post-Training \& Real-Robot Deployment:} Baole Fang\footnotemark[1], Xudong Li\footnotemark[1], Xijie Yang\footnotemark[1]

\noindent\textbf{Technical Lead (listed alphabetically):} Junting Dong\footnote[2]{Corresponding authors. 
}, Haoyu Guo\footnotemark[2], Tao Lu\footnotemark[2], Mulin Yu\footnotemark[2] 

\noindent\textbf{Project Lead:} Bowen Zhou, Bin Zhao, Tianfan Xue, Weinan Zhang\footnotemark[2], Chunhua Shen\footnotemark[2]

\subsection*{Contributors}

\noindent Xueyue Zhu, Chao Gao, Zeyu He, Yuanzhen Zhou, Jianyang Zhang, Siwei Cui, Xing Gao, Yifei Yao, Qiaojun Yu, Kailin Li, Ming Zhou, Xinzhe Wang, Yingxiang Xu, Mu Huang, Zetao Cai, Fuxian Huang, Yunsong Zhou, Yufei Xue, Wenqi Guo, Jianjun Zhou, Xinyue Li, Kerui Ren, Weiguang Zhao, Ni Yang, Wenze Cui, Bingqi Jiang, Rong Fu, Hengjie Li

\endgroup

\clearpage
\bibliographystyle{plainnat}
\bibliography{refs}

@string{iccv = "International Conference on Computer Vision (ICCV)"}

@string{icra = "IEEE International Conference on Robotics and Automation (ICRA)"}

@string{rss = "Robotics: Science and Systems Conference (RSS)"}

@string{chi = "ACM Conference on Human Factors in Computing Systems (CHI)"}

@misc{chen2026definition,
title         = {A Definition and Roadmap for World Models},
author        = {Chen, Xinyuan and Guo, Haoyu and Guo, Shi and Jiang, Bingqi
                 and Shen, Chunhua and Shen, Xing and Xue, Tianfan
                 and Xue, Yufei and Yu, Mulin and Zhang, Weinan
                 and Zhao, Bin and Zhou, Bowen and Zhou, Ming},
year          = {2026},
eprint        = {2607.06401},
archivePrefix = {arXiv},
primaryClass  = {cs.AI},
url           = {https://arxiv.org/abs/2607.06401}
}

@misc{yuan2026fastwam,
title         = {Fast-WAM: Do World Action Models Need Test-time Future Imagination?},
author        = {Yuan, Tianyuan and Dong, Zibin and Liu, Yicheng and Zhao, Hang},
year          = {2026},
eprint        = {2603.16666},
archivePrefix = {arXiv},
primaryClass  = {cs.CV},
url           = {https://arxiv.org/abs/2603.16666}
}

@article{wan2025wan,
  title         = {Wan: Open and Advanced Large-Scale Video Generative Models},
  author        = {{Wan Team}},
  journal       = {arXiv preprint arXiv:2503.20314},
  year          = {2025},
  eprint        = {2503.20314},
  archivePrefix = {arXiv},
  primaryClass  = {cs.CV},
  url           = {https://arxiv.org/abs/2503.20314}
}

@article{raffel2020t5,
  title   = {Exploring the Limits of Transfer Learning with a Unified Text-to-Text Transformer},
  author  = {Raffel, Colin and Shazeer, Noam and Roberts, Adam and Lee, Katherine and Narang, Sharan and Matena, Michael and Zhou, Yanqi and Li, Wei and Liu, Peter J.},
  journal = {Journal of Machine Learning Research},
  volume  = {21},
  number  = {140},
  pages   = {1--67},
  year    = {2020},
  url     = {https://www.jmlr.org/papers/v21/20-074.html}
}

@inproceedings{ichter2023saycan,
  title     = {Do As I Can, Not As I Say: Grounding Language in Robotic Affordances},
  author    = {Ichter, Brian and Brohan, Anthony and Chebotar, Yevgen and Finn, Chelsea and Hausman, Karol and Herzog, Alexander and Ho, Daniel and Ibarz, Julian and Irpan, Alex and Jang, Eric and Julian, Ryan and Kalashnikov, Dmitry and Levine, Sergey and Lu, Yao and Parada, Carolina and Rao, Kanishka and Sermanet, Pierre and Toshev, Alexander T. and Vanhoucke, Vincent and Xia, Fei and Xiao, Ted and Xu, Peng and Yan, Mengyuan and Brown, Noah and Ahn, Michael and Cortes, Omar and Sievers, Nicolas and Tan, Clayton and Xu, Sichun and Reyes, Diego and Rettinghouse, Jarek and Quiambao, Jornell and Pastor, Peter and Luu, Linda and Lee, Kuang-Huei and Kuang, Yuheng and Jesmonth, Sally and Joshi, Nikhil J. and Jeffrey, Kyle and Ruano, Rosario Jauregui and Hsu, Jasmine and Gopalakrishnan, Keerthana and David, Byron and Zeng, Andy and Fu, Chuyuan Kelly},
  booktitle = {Proceedings of the 6th Conference on Robot Learning},
  series    = {Proceedings of Machine Learning Research},
  volume    = {205},
  pages     = {287--318},
  publisher = {PMLR},
  year      = {2023},
  url       = {https://proceedings.mlr.press/v205/ichter23a.html}
}

@article{brohan2023rt2,
  title         = {{RT-2}: Vision-Language-Action Models Transfer Web Knowledge to Robotic Control},
  author        = {Brohan, Anthony and Brown, Noah and Carbajal, Justice and Chebotar, Yevgen and Chen, Xi and Choromanski, Krzysztof and Ding, Tianli and Driess, Danny and Dubey, Avinava and Finn, Chelsea and Florence, Pete and Fu, Chuyuan and Gonzalez Arenas, Montse and Gopalakrishnan, Keerthana and Han, Kehang and Hausman, Karol and Herzog, Alexander and Hsu, Jasmine and Ichter, Brian and Irpan, Alex and Joshi, Nikhil and Julian, Ryan and Kalashnikov, Dmitry and Kuang, Yuheng and Leal, Isabel and Lee, Lisa and Lee, Tsang-Wei Edward and Levine, Sergey and Lu, Yao and Michalewski, Henryk and Mordatch, Igor and Pertsch, Karl and Rao, Kanishka and Reymann, Krista and Ryoo, Michael and Salazar, Grecia and Sanketi, Pannag and Sermanet, Pierre and Singh, Jaspiar and Singh, Anikait and Soricut, Radu and Tran, Huong and Vanhoucke, Vincent and Vuong, Quan and Wahid, Ayzaan and Welker, Stefan and Wohlhart, Paul and Wu, Jialin and Xia, Fei and Xiao, Ted and Xu, Peng and Xu, Sichun and Yu, Tianhe and Zitkovich, Brianna},
  journal       = {arXiv preprint arXiv:2307.15818},
  year          = {2023},
  eprint        = {2307.15818},
  archivePrefix = {arXiv},
  primaryClass  = {cs.RO},
  url           = {https://arxiv.org/abs/2307.15818}
}

@inproceedings{jiang2023vima,
  title     = {{VIMA}: Robot Manipulation with Multimodal Prompts},
  author    = {Jiang, Yunfan and Gupta, Agrim and Zhang, Zichen and Wang, Guanzhi and Dou, Yongqiang and Chen, Yanjun and Fei-Fei, Li and Anandkumar, Anima and Zhu, Yuke and Fan, Linxi},
  booktitle = {Proceedings of the 40th International Conference on Machine Learning},
  series    = {Proceedings of Machine Learning Research},
  volume    = {202},
  pages     = {14975--15022},
  publisher = {PMLR},
  year      = {2023},
  url       = {https://proceedings.mlr.press/v202/jiang23b.html}
}

@inproceedings{huang2023innermonologue,
  title     = {Inner Monologue: Embodied Reasoning through Planning with Language Models},
  author    = {Huang, Wenlong and Xia, Fei and Xiao, Ted and Chan, Harris and Liang, Jacky and Florence, Pete and Zeng, Andy and Tompson, Jonathan and Mordatch, Igor and Chebotar, Yevgen and Sermanet, Pierre and Jackson, Tomas and Brown, Noah and Luu, Linda and Levine, Sergey and Hausman, Karol and Ichter, Brian},
  booktitle = {Proceedings of the 6th Conference on Robot Learning},
  series    = {Proceedings of Machine Learning Research},
  volume    = {205},
  pages     = {1769--1782},
  publisher = {PMLR},
  year      = {2023},
  url       = {https://proceedings.mlr.press/v205/huang23c.html}
}

@article{lingbot-va2026,
  title={Causal World Modeling for Robot Control},
  author={Li, Lin and Zhang, Qihang and Luo, Yiming and Yang, Shuai and Wang, Ruilin and Han, Fei and Yu, Mingrui and Gao, Zelin and Xue, Nan and Zhu, Xing and Shen, Yujun and Xu, Yinghao},
  journal={arXiv preprint arXiv:2601.21998},
  year={2026}
}

@misc{rtc_training_time_arxiv2512_05964,
  title         = {Training-Time Action Conditioning for Efficient Real-Time Chunking},
  author        = {Black, Kevin and Ren, Allen Z. and Equi, Michael and Levine, Sergey},
  year          = {2025},
  eprint        = {2512.05964},
  archivePrefix = {arXiv},
  primaryClass  = {cs.RO},
  doi           = {10.48550/arXiv.2512.05964},
  url           = {https://arxiv.org/abs/2512.05964}
}

@misc{king2026gen4u,
  title        = {Gen4U: Unifying Video Generation and Understanding via Diffusion},
  author       = {King, Michael and Mahendran, Aravindh and Grimes, Matthew Koichi and Kitashov, Fedor and Elarabawy, Adham and Velez, Pedro and Ovsjanikov, Maks and P{\u{a}}tr{\u{a}}ucean, Viorica},
  year         = {2026},
  eprint       = {2607.06856},
  archivePrefix= {arXiv},
  primaryClass = {cs.CV}
}

@misc{esmati2026invisible,
  title        = {The Invisible Hand of Physics: When Video Diffusion Models Know More Than They Show},
  author       = {Esmati, Parsa and Nath, Somjit and Hofmann, Katja and Nowrouzezahrai, Derek and Kahou, Samira Ebrahimi and Mirmehdi, Majid},
  year         = {2026},
  eprint       = {2606.05328},
  archivePrefix= {arXiv},
  primaryClass = {cs.GR}
}

@article{song2026habit,
  author  = {Jaehwi Song and Suchae Jeong and Byeongguk Jeon and Sungdong Kim and Minjoon Seo and Hyungmok Son and Kimin Lee},
  title   = {HABIT: Human-Aware Behavior and Interaction Training Dataset for Robot Manipulation},
  journal = {arXiv preprint arXiv:2606.31682},
  year    = {2026}
}

@article{fourier2025actionnet,
  author    = {Fourier ActionNet Team, Yao Mu},
  title     = {ActionNet: A dataset for dexterous bimanual manipulation},
  year      = {2025},
}

@article{lu2026track4world,
  title={Track4World: Feedforward World-centric Dense 3D Tracking of All Pixels},
  author={Lu, Jiahao and Xu, Jiayi and Hu, Wenbo and Zhu, Ruijie and Zhao, Chengfeng and Yeung, Sai-Kit and Shan, Ying and Liu, Yuan},
  journal={arXiv preprint arXiv:2603.02573},
  year={2026}
}

@article{track4action,
  title={{Track4Action: Distilling World-Centric 3D Tracker into Vision-Language-Action Policies}},
  author={Wang, Chenyi and Wang, Xinkai and Lin, Bokai and Tian, Jialin and Zhang, Fucheng and Lu, Cewu and Yang, Lixin},
  journal={arXiv preprint arXiv:2608.03727},
  year={2026},
  eprint={2608.03727},
  archivePrefix={arXiv},
  url={https://arxiv.org/abs/2608.03727}
}

@misc{fang2026molmoact2actionreasoningmodels,
      title={MolmoAct2: Action Reasoning Models for Real-world Deployment},
      author={Haoquan Fang and Jiafei Duan and Donovan Clay and Sam Wang and Shuo Liu and Weikai Huang and Xiang Fan and Wei-Chuan Tsai and Shirui Chen and Yi Ru Wang and Shanli Xing and Jaemin Cho and Jae Sung Park and Ainaz Eftekhar and Peter Sushko and Karen Farley and Angad Wadhwa and Cole Harrison and Winson Han and Ying-Chun Lee and Eli VanderBilt and Rose Hendrix and Suveen Ellawela and Lucas Ngoo and Joyce Chai and Zhongzheng Ren and Ali Farhadi and Dieter Fox and Ranjay Krishna},
      year={2026},
      eprint={2605.02881},
      archivePrefix={arXiv},
      primaryClass={cs.RO},
      url={https://arxiv.org/abs/2605.02881},
}

@misc{realsourceworld,
  title={RealSource World: A Large-Scale Real-World Dual-Arm Manipulation Dataset},
  author={RealSource},
  howpublished={\url{https://huggingface.co/datasets/RealSourceData/RealSource-World}},
  year={2025}
}

@inproceedings{RoboHive,
  title     = {RoboHive -- A Unified Framework for Robot Learning},
  author    = {Vikash Kumar and Rutav Shah and Gaoyue Zhou and Vincent Moens and Vittorio Caggiano and Jay Vakil and Abhishek Gupta and Aravind Rajeswaran},
  booktitle = {NeurIPS: Conference on Neural Information Processing Systems},
  year      = {2023},
  url       = {https://sites.google.com/view/robohive},
  eprint    = {https://arxiv.org/abs/2310.06828},

}

@article{liu2024rdt,
  title={RDT-1B: a Diffusion Foundation Model for Bimanual Manipulation},
  author={Liu, Songming and Wu, Lingxuan and Li, Bangguo and Tan, Hengkai and Chen, Huayu and Wang, Zhengyi and Xu, Ke and Su, Hang and Zhu, Jun},
  journal={arXiv preprint arXiv:2410.07864},
  year={2024}
}

@inproceedings{fang2024rh20t,
  title        = {RH20T: A Comprehensive Robotic Dataset for Learning Diverse Skills in One-Shot},
  author       = {Fang, Hao-Shu and Fang, Hongjie and Tang, Zhenyu and Liu, Jirong and Wang, Chenxi and Wang, Junbo and Zhu, Haoyi and Lu, Cewu},
  booktitle    = {2024 IEEE International Conference on Robotics and Automation (ICRA)},
  pages        = {653--660},
  year         = {2024},
  organization = {IEEE}
}

@article{RoboCOINReport,
  author = {Shihan Wu and Xuecheng Liu and Shaoxuan Xie and Pengwei Wang and Xinghang Li and Bowen Yang and Zhe Li and Kai Zhu and Hongyu Wu and Yiheng Liu and Zhaoye Long and Yue Wang and Chong Liu and Dihan Wang and Ziqiang Ni and Xiang Yang and You Liu and Ruoxuan Feng and Runtian Xu and Lei Zhang and Denghang Huang and Chenghao Jin and Anlan Yin and Xinlong Wang and Zhenguo Sun and Junkai Zhao and Mengfei Du and Mingyu Cao and Xiansheng Chen and Hongyang Cheng and Xiaojie Zhang and Yankai Fu and Ning Chen and Cheng Chi and Sixiang Chen and Huaihai Lyu and Xiaoshuai Hao and Yequan Wang and Bo Lei and Dong Liu and Xi Yang and Yance Jiao and Tengfei Pan and Yunyan Zhang and Songjing Wang and Ziqian Zhang and Xu Liu and Ji Zhang and Caowei Meng and Zhizheng Zhang and Jiyang Gao and Song Wang and Xiaokun Leng and Zhiqiang Xie and Zhenzhen Zhou and Peng Huang and Wu Yang and Yandong Guo and Yichao Zhu and Suibing Zheng and Hao Cheng and Xinmin Ding and Yang Yue and Huanqian Wang and Chi Chen and Jingrui Pang and YuXi Qian and Haoran Geng and Lianli Gao and Haiyuan Li and Bin Fang and Gao Huang and Yaodong Yang and Hao Dong and He Wang and Hang Zhao and Yadong Mu and Di Hu and Hao Zhao and Tiejun Huang and Shanghang Zhang and Yonghua Lin and Zhongyuan Wang and Guocai Yao},
  title = {RoboCOIN: An Open-Sourced Bimanual Robotic Data Collection for Integrated Manipulation},
  year = {2025},
  url = {https://github.com/FlagOpen/RoboCOIN}
}

@article{galaxea2025,
  title={Galaxea G0: Open-World Dataset and Dual-System VLA Model},
  author={Galaxea Team},
  journal={arXiv preprint arXiv:2509.00576},
  year={2025}
}

@misc{abc2026,
  title         = {Scalable Behavior Cloning with Open Data, Training, and Evaluation},
  author        = {Arthur Allshire and Himanshu Gaurav Singh and Ritvik Singh and Adam Rashid and Hongsuk Choi and David McAllister and Justin Yu and Yiyuan Chen and Huang Huang and Pieter Abbeel and Xi Chen and Rocky Duan and Phillip Isola and Jitendra Malik and Fred Shentu and Guanya Shi and Philipp Wu and Angjoo Kanazawa},
  year          = {2026},
  eprint        = {2606.27375},
  archivePrefix = {arXiv},
  primaryClass  = {cs.RO},
  doi           = {10.48550/arXiv.2606.27375},
  url           = {https://arxiv.org/abs/2606.27375},
}

@inproceedings{wu2025robomind,
              title={Robomind: Benchmark on multi-embodiment intelligence normative data for robot manipulation},
              author={Wu, Kun and Hou, Chengkai and Liu, Jiaming and Che, Zhengping and Ju, Xiaozhu and Yang, Zhuqin and Li, Meng and Zhao, Yinuo and Xu, Zhiyuan and Yang, Guang and others},
              booktitle={Robotics: Science and Systems (RSS) 2025},
              year={2025},
              publisher={Robotics: Science and Systems Foundation},
              url={https://www.roboticsproceedings.org/rss21/p152.pdf}
}

@misc{hou2025robomind20multimodalbimanual,
      title={RoboMIND 2.0: A Multimodal, Bimanual Mobile Manipulation Dataset for Generalizable Embodied Intelligence},
      author={Chengkai Hou and Kun Wu and Jiaming Liu and Zhengping Che and Di Wu and Fei Liao and Guangrun Li and Jingyang He and Qiuxuan Feng and Zhao Jin and Chenyang Gu and Zhuoyang Liu and Nuowei Han and Xiangju Mi and Yaoxu Lv and Yankai Fu and Gaole Dai and Langzhe Gu and Tao Li and Yuheng Zhang and Yixue Zhang and Xinhua Wang and Shichao Fan and Meng Li and Zhen Zhao and Ning Liu and Zhiyuan Xu and Pei Ren and Junjie Ji and Haonan Liu and Kuan Cheng and Shanghang Zhang and Jian Tang},
      year={2025},
      eprint={2512.24653},
      archivePrefix={arXiv},
      primaryClass={cs.RO},
      url={https://arxiv.org/abs/2512.24653},
}

@misc{contributors2025internroboticsrepo,
  title={InternData-A1},
  author={InternData-A1 contributors},
  howpublished={\url{https://github.com/InternRobotics/InternManip}},  year={2025}
}

@misc{contributors2024agibotworldrepo,
  title={AgiBot World Colosseum},
  author={AgiBot World Colosseum contributors},
  howpublished={\url{https://github.com/OpenDriveLab/AgiBot-World}},
  year={2024}
}

@misc{contributors2026litegenrepo,
  title={LiteGen},
  author={LiteGen contributors},
  howpublished={\url{https://github.com/DeepLink-org/LiteGen}},
  year={2026}
}

@misc{rw_rl_dataset_2026,
  title        = {RW-RL Dataset: Real-World Reinforcement Learning Dataset},
  author       = {Boden Intelligence and Junpu Innovation Center and MINT Lab, Shanghai Jiao Tong University},
  year         = {2026},
  howpublished = {https://huggingface.co/datasets/MINT-SJTU/RW-RL-Dataset}
}

@misc{zhang2026dexoraopensourcevlahighdof,
      title={Dexora: Open-source VLA for High-DoF Bimanual Dexterity},
      author={Zongzheng Zhang and Jingrui Pang and Zhuo Yang and Kun Li and Minwen Liao and Saining Zhang and Guoxuan Chi and Jinbang Guo and Huan-ang Gao and Modi Shi and Dongyun Ge and Yao Mu and Jiayuan Gu and Rui Chen and Hao Dong and Huazhe Xu and Li Yi and Yixin Zhu and Hang Zhao and Pengwei Wang and Shanghang Zhang and Guocai Yao and Jianyu Chen and Hongyang Li and Hao Zhao},
      year={2026},
      eprint={2605.18722},
      archivePrefix={arXiv},
      primaryClass={cs.RO},
      url={https://arxiv.org/abs/2605.18722},
}

@misc{hoque2025egodex,
  title         = {{EgoDex}: Learning Dexterous Manipulation from
                   Large-Scale Egocentric Video},
  author        = {Ryan Hoque and Peide Huang and David J. Yoon
                   and Mouli Sivapurapu and Jian Zhang},
  year          = {2025},
  eprint        = {2505.11709},
  archivePrefix = {arXiv},
  primaryClass  = {cs.CV},
  url           = {https://arxiv.org/abs/2505.11709}
}

@misc{punamiya2026egoverse,
  title         = {{EgoVerse}: An Egocentric Human Dataset for
                   Robot Learning from Around the World},
  author        = {Ryan Punamiya and Simar Kareer and Zeyi Liu
                   and others},
  year          = {2026},
  eprint        = {2604.07607},
  archivePrefix = {arXiv},
  primaryClass  = {cs.RO},
  url           = {https://arxiv.org/abs/2604.07607}
}

@article{zhang2026hy,
  title   = {{Hy-Embodied-0.5-VLA}: From Vision-Language-Action Models to a Real-World Robot Learning Stack},
  author  = {Zhang, He and Xiang, Lingzhu and Lin, Haitao and
             Huang, Zeyu and Wang, Minghui and Zhong, Dingyan and
             Dong, Yubo and Wu, Yihao and Rao, Yongming and
             Zhang, Dongsheng and others},
  journal = {arXiv preprint arXiv:2606.14409},
  year    = {2026}
}

@article{wang2025pi3,
    title   = {{$\pi^3$}: Permutation-Equivariant Visual Geometry Learning},
    author  = {Wang, Yifan and Zhou, Jianjun and Zhu, Haoyi
               and Chang, Wenzheng and Zhou, Yang and Li, Zizun
               and Chen, Junyi and Pang, Jiangmiao and Shen, Chunhua
               and He, Tong},
    journal = {arXiv preprint arXiv:2507.13347},
    year    = {2025}
  }

@article{lai2026cowtracker,
    title   = {{CoWTracker}: Tracking by Warping instead of Correlation},
    author  = {Lai, Zihang and Insafutdinov, Eldar
               and Sucar, Edgar and Vedaldi, Andrea},
    journal = {arXiv preprint arXiv:2602.04877},
    year    = {2026}
  }

@inproceedings{libero,
  title = {{LIBERO: Benchmarking Knowledge Transfer for Lifelong Robot Learning}},
  author = {Bo Liu and Yifeng Zhu and Chongkai Gao and Yihao Feng and Qiang Liu and Yuke Zhu and
            Peter Stone},
  booktitle = {Advances in Neural Information Processing Systems},
  year = {2023},
  volume = {36},
  url = {https://proceedings.neurips.cc/paper_files/paper/2023/hash/8c3c666820ea055a77726d66fc7d447f-Abstract-Datasets_and_Benchmarks.html},
}

@inproceedings{libero_plus,
  title = {{LIBERO-Plus: A Progressive Robustness Benchmark for Visual-Language-Action Models}},
  author = {Senyu Fei and Siyin Wang and Junhao Shi and Zihao Dai and Jikun Cai and
            Pengfang Qian and Li Ji and Xinzhe He and Shiduo Zhang and Zhaoye Fei and Jinlan Fu and
            Jingjing Gong and Xipeng Qiu},
  booktitle = {Proceedings of the IEEE/CVF Conference on Computer Vision and Pattern Recognition},
  year = {2026},
  url = {https://openaccess.thecvf.com/content/CVPR2026/html/Fei_LIBERO-Plus_A_Progressive_Robustness_Benchmark_for_Visual-Language-Action_Models_CVPR_2026_paper.html},
}

@inproceedings{robotwin2,
  title = {{RoboTwin 2.0: A Scalable Data Generator and Benchmark with Strong Domain Randomization for Robust Bimanual Robotic Manipulation}},
  author = {Tianxing Chen and Zanxin Chen and Baijun Chen and Zijian Cai and Yibin Liu and
            Zixuan Li and Qiwei Liang and Xianliang Lin and Yiheng Ge and Zhenyu Gu and
            Weiliang Deng and Yubin Guo and Tian Nian and Xuanbing Xie and Qiangyu Chen and
            Kailun Su and Tianling Xu and Guodong Liu and Mengkang Hu and Huan-ang Gao and
            Kaixuan Wang and Zhixuan Liang and Yusen Qin and Xiaokang Yang and Ping Luo and Yao Mu},
  booktitle = {Proceedings of the International Conference on Machine Learning},
  year = {2026},
  url = {https://arxiv.org/abs/2506.18088},
}

@article{ebench,
  title = {{EBench: Elemental Diagnosis of Generalist Mobile Manipulation Policies}},
  author = {Ning Gao and Jinliang Zheng and Xing Gao and Haoxiang Ma and Hanqing Wang and
            Yukai Wang and Jiantong Chen and Zanxin Chen and Shujie Zhang and Mingda Jia and
            Xuekun Jiang and Zihou Zhu and Xinyu Li and Shuai Wang and Hao Li and Wenzhe Cai and
            Yuqiang Yang and Xudong Xu and Zhaoyang Lyu and Yao Mu and Tai Wang and
            Jiangmiao Pang and Jia Zeng and Weinan Zhang and Chunhua Shen},
  journal = {arXiv preprint arXiv:2606.18239},
  year = {2026},
  eprint = {2606.18239},
  archivePrefix = {arXiv},
  url = {https://arxiv.org/abs/2606.18239},
}

@article{robodojo,
  title = {{RoboDojo: A Unified Sim-and-Real Benchmark for Comprehensive Evaluation of Generalist Robot Manipulation Policies}},
  author = {Tianxing Chen and Yue Chen and Zixuan Li and Junyuan Tang and Kailun Su and
            Haoran Lu and Weijie Wan and Baijun Chen and Songling Liu and Haowen Yan and
            Honghao Su and Zhiyang Dou and Kaixuan Wang and Dandan Zhang and Yunze Liu and
            Yan Qin and Qiwei Liang and Qiwei Wu and Zijian Lin and Wenwei Lin and Yuran Wang and
            Minghua He and Tianshu Wu and Ruihai Wu and Jingquan Zhou and Kai-Chong Lei and
            Haibao Yu and Yuanfeng Ji and Weiyang Jin and Guanyu Lin and Xiaofan Li and Qi Xiong and
            Renjing Xu and Zhongyu Li and Wenhao Chai and Enze Xie and Ziwei Wang and Yao Mu and
            Hao Dong and Wojciech Matusik and Mingyu Ding and Wenbo Ding and Ping Luo and
            Masayoshi Tomizuka},
  journal = {arXiv preprint arXiv:2607.04434},
  year = {2026},
  eprint = {2607.04434},
  archivePrefix = {arXiv},
  url = {https://arxiv.org/abs/2607.04434},
}

@article{qwen_robotmanip,
  title = {{Qwen-RobotManip Technical Report: Alignment Unlocks Scale for Robotic Manipulation Foundation Models}},
  author = {Haoqi Yuan and Zhixuan Liang and Anzhe Chen and Ye Wang and Haoyang Li and Pei Lin and
            Yiyang Huang and Zixing Lei and Tong Zhang and Jiazhao Zhang and Jie Zhang and
            Jingyang Fan and Gengze Zhou and Qihang Peng and Chenxu Lv and Xiaoyue Chen and
            An Yang and Fei Huang and Junyang Lin and Dayiheng Liu and Jingren Zhou and
            Chenfei Wu and Xiong-Hui Chen},
  journal = {arXiv preprint arXiv:2606.17846},
  year = {2026},
  eprint = {2606.17846},
  archivePrefix = {arXiv},
  url = {https://arxiv.org/abs/2606.17846},
}

@inproceedings{pi0,
  title = {{$\pi_{0}$: A Vision-Language-Action Flow Model for General Robot Control}},
  author = {Kevin Black and Noah Brown and Danny Driess and Adnan Esmail and Michael Robert Equi and
            Chelsea Finn and Niccolo Fusai and Lachy Groom and Karol Hausman and Brian Ichter and
            Szymon Jakubczak and Tim Jones and Liyiming Ke and Sergey Levine and Adrian Li-Bell and
            Mohith Mothukuri and Suraj Nair and Karl Pertsch and Lucy Xiaoyang Shi and
            Laura Smith and James Tanner and Quan Vuong and Anna Walling and Haohuan Wang and
            Ury Zhilinsky},
  booktitle = {Proceedings of Robotics: Science and Systems},
  year = {2025},
  doi = {10.15607/RSS.2025.XXI.010},
  url = {https://www.roboticsproceedings.org/rss21/p010.html},
}

@inproceedings{pi0_fast,
  title = {{FAST: Efficient Action Tokenization for Vision-Language-Action Models}},
  author = {Karl Pertsch and Kyle Stachowicz and Brian Ichter and Danny Driess and Suraj Nair and
            Quan Vuong and Oier Mees and Chelsea Finn and Sergey Levine},
  booktitle = {Proceedings of Robotics: Science and Systems},
  year = {2025},
  doi = {10.15607/RSS.2025.XXI.012},
  url = {https://www.roboticsproceedings.org/rss21/p012.html},
}

@article{ript_vla,
  title = {{Interactive Post-Training for Vision-Language-Action Models}},
  author = {Shuhan Tan and Kairan Dou and Yue Zhao and Philipp Kr{\"a}henb{\"u}hl},
  journal = {arXiv preprint arXiv:2505.17016},
  year = {2025},
  eprint = {2505.17016},
  archivePrefix = {arXiv},
  url = {https://arxiv.org/abs/2505.17016},
}

@inproceedings{openvla_oft,
  title = {{Fine-Tuning Vision-Language-Action Models: Optimizing Speed and Success}},
  author = {Moo Jin Kim and Chelsea Finn and Percy Liang},
  booktitle = {Proceedings of Robotics: Science and Systems},
  year = {2025},
  doi = {10.15607/RSS.2025.XXI.017},
  url = {https://www.roboticsproceedings.org/rss21/p017.html},
}

@article{starvla,
  title = {{StarVLA: A Lego-like Codebase for Vision-Language-Action Model Developing}},
  author = {{StarVLA Community}},
  journal = {arXiv preprint arXiv:2604.05014},
  year = {2026},
  eprint = {2604.05014},
  archivePrefix = {arXiv},
  url = {https://arxiv.org/abs/2604.05014},
}

@inproceedings{vla_jepa,
  title = {{VLA-JEPA: Enhancing Vision-Language-Action Model with Latent World Model}},
  author = {Jingwen Sun and Wenyao Zhang and Zekun Qi and Shaojie Ren and Zezhi Liu and
            Hanxin Zhu and Guangzhong Sun and Xin Jin and Zhibo Chen},
  booktitle = {European Conference on Computer Vision},
  year = {2026},
  url = {https://arxiv.org/abs/2602.10098},
}

@article{vlact,
  title = {{Beyond Data Scaling: Representation-Centric Continued Pre-training for Vision-Language-Action Models}},
  author = {Senqiao Yang and Chengyao Wang and Yuxin Chen and Zixuan Wang and Longxiang Tang and
            Haokun Gui and Jinhui Ye and Changsheng Lu and Xiaoyang Wu and Mingkang Zhu and
            Pengguang Chen and Shu Liu and Zhuotao Tian and Hengshuang Zhao and Bei Yu and Jiaya Jia},
  journal = {arXiv preprint arXiv:2608.27550},
  year = {2026},
  eprint = {2608.27550},
  archivePrefix = {arXiv},
  url = {https://arxiv.org/abs/2608.27550},
}

@inproceedings{pi05,
  title = {{$\pi_{0.5}$: a Vision-Language-Action Model with Open-World Generalization}},
  author = {Kevin Black and Noah Brown and James Darpinian and Karan Dhabalia and Danny Driess and
            Adnan Esmail and Michael Robert Equi and Chelsea Finn and Niccolo Fusai and
            Manuel Y. Galliker and Dibya Ghosh and Lachy Groom and Karol Hausman and
            Brian Ichter and Szymon Jakubczak and Tim Jones and Liyiming Ke and Devin LeBlanc and
            Sergey Levine and Adrian Li-Bell and Mohith Mothukuri and Suraj Nair and
            Karl Pertsch and Allen Z. Ren and Lucy Xiaoyang Shi and Laura Smith and
            Jost Tobias Springenberg and Kyle Stachowicz and James Tanner and Quan Vuong and
            Homer Walke and Anna Walling and Haohuan Wang and Lili Yu and Ury Zhilinsky},
  booktitle = {Proceedings of The 9th Conference on Robot Learning},
  year = {2025},
  pages = {17--40},
  volume = {305},
  series = {Proceedings of Machine Learning Research},
  publisher = {PMLR},
  url = {https://proceedings.mlr.press/v305/black25a.html},
}

@article{internvla_a15,
  title = {{InternVLA-A1.5: Unifying Understanding, Latent Foresight, and Action for Compositional Generalization}},
  author = {Haoxiang Ma and Junhao Cai and Xiaoxu Xu and Hao Li and Yuyin Yang and Yang Tian and
            Jiafei Cao and Hongrui Zhu and Zherui Qiu and Zhaxizhuoma and Yuqiang Yang and
            Jiaqi Peng and Xueyuan Wei and Yangkun Zhu and Jiahao Jiang and Xing Gao and
            Hanqing Wang and Feng Yuan and Kailin Li and Xueyue Zhu and Tai Wang and Yan Ding and
            Jiangmiao Pang and Jia Zeng and Jingjing Zhang and Bowen Zhou and Yao Mu and
            Chunhua Shen and Weinan Zhang},
  journal = {arXiv preprint arXiv:2607.04988},
  year = {2026},
  eprint = {2607.04988},
  archivePrefix = {arXiv},
  url = {https://arxiv.org/abs/2607.04988},
}

@inproceedings{caron2021emerging,
  title     = {Emerging Properties in Self-Supervised Vision Transformers},
  author    = {Caron, Mathilde and
               Touvron, Hugo and
               Misra, Ishan and
               J{\'e}gou, Herv{\'e} and
               Mairal, Julien and
               Bojanowski, Piotr and
               Joulin, Armand},
  booktitle = {Proceedings of the International Conference on Computer Vision (ICCV)},
  year      = {2021}
}

@article{internvla_a1,
  title   = {{InternVLA-A1}: Unifying Understanding, Generation
             and Action for Robotic Manipulation},
  author  = {Cai, Junhao and Cai, Zetao and Cao, Jiafei
             and Chen, Yilun and He, Zeyu and Jiang, Lei
             and Li, Hang and Li, Hengjie and Li, Yang
             and Liu, Yufei and others},
  journal = {arXiv preprint arXiv:2601.02456},
  year    = {2026},
  url     = {https://arxiv.org/abs/2601.02456}
}

@article{wam4d,
  title = {{4D-WAM: Infusing Spatiotemporal Awareness into World Action Models through Trajectory Fields}},
  author = {Lishan Yang and Wenxuan Song and Xi Wang and Pingyue Sheng and Zheng Fang and
            Ziyang Zhou and Junjie He and Haodong Yan and Jiayi Chen and Nan Sun and Qiao Sun and
            Pengwei Wang and Lingqiao Liu and Yan Wang and Yuxiang Gao and Feras Dayoub and
            Haoang Li},
  journal = {arXiv preprint arXiv:2608.08023},
  year = {2026},
  eprint = {2608.08023},
  archivePrefix = {arXiv},
  url = {https://arxiv.org/abs/2608.08023},
}

@article{st_wam,
  title = {{ST-WAM: Semantic-Temporal World Action Model for Robust Manipulation under Visual Distribution Shifts}},
  author = {Mingxin Wang and Bin Hu and Bin Qian and Kaitao Jiang and Haoning Wu and Feng Yan and
            Bowen Jing and Ruiyang Hao and Enyi Wang and Kangning Niu and Yandan Yang and Mu Xu and
            Yan Wang and Houde Liu and Tianlun Li},
  journal = {arXiv preprint arXiv:2607.28993},
  year = {2026},
  eprint = {2607.28993},
  archivePrefix = {arXiv},
  url = {https://arxiv.org/abs/2607.28993},
}

@article{faster_wam_depth,
  title = {{Faster-WAM: Do World Action Models Need Deep Action Modules?}},
  author = {Liheng Ma and Rui Heng Yang and Zhanguang Zhang and Mateo Clemente and Ziwen Hu and
            Tongtong Cao and Yingxue Zhang},
  journal = {arXiv preprint arXiv:2608.02365},
  year = {2026},
  eprint = {2608.02365},
  archivePrefix = {arXiv},
  url = {https://arxiv.org/abs/2608.02365},
}

@article{jepa_wam,
  title = {{JEPA-WAM: Learning Vision-Language-Action Policies with Joint-Embedding World Modeling}},
  author = {Yihan Lin and Jiawei He and Shifeng Bao and Chen Zhao and Yang Li and Xiaobo Wang and
            Yan Wang and Cheng Chi and Jing Zhang},
  journal = {arXiv preprint arXiv:2608.09381},
  year = {2026},
  eprint = {2608.09381},
  archivePrefix = {arXiv},
  url = {https://arxiv.org/abs/2608.09381},
}

@article{cosmos_policy,
  title = {{Cosmos Policy: Fine-Tuning Video Models for Visuomotor Control and Planning}},
  author = {Moo Jin Kim and Yihuai Gao and Tsung-Yi Lin and Yen-Chen Lin and Yunhao Ge and
            Grace Lam and Percy Liang and Shuran Song and Ming-Yu Liu and Chelsea Finn and Jinwei Gu},
  journal = {arXiv preprint arXiv:2601.16163},
  year = {2026},
  eprint = {2601.16163},
  archivePrefix = {arXiv},
  url = {https://arxiv.org/abs/2601.16163},
}

@article{imagewam,
  title = {{ImageWAM: Do World Action Models Really Need Video Generation, or Just Image Editing?}},
  author = {Yuyang Zhang and Wenyao Zhang and Zekun Qi and He Zhang and Haitao Lin and
            Jingbo Zhang and Yao Mu and Xiaokang Yang and Wenjun Zeng and Xin Jin},
  journal = {arXiv preprint arXiv:2606.19531},
  year = {2026},
  eprint = {2606.19531},
  archivePrefix = {arXiv},
  url = {https://arxiv.org/abs/2606.19531},
}

@article{abot_m05,
  title = {{ABot-M0.5: Unified Mobility-and-Manipulation World Action Model}},
  author = {Ronghan Chen and Yandan Yang and Zuojin Tang and Dongjie Huo and Tong Lin and
            Haoning Wu and Haoyun Liu and Yuzhi Chen and Lulu Zheng and Botai Yuan and
            Tianlun Li and Mingxin Wang and Dekang Qi and Bin Hu and Wei Mei and Yuze Xuan and
            Haolong Yang and Yanqing Zhu and Mu Xu and Zhiheng Ma and Xinyuan Chang},
  journal = {arXiv preprint arXiv:2607.00678},
  year = {2026},
  eprint = {2607.00678},
  archivePrefix = {arXiv},
  url = {https://arxiv.org/abs/2607.00678},
}

@article{being_h07,
  title = {{Being-H0.7: A Latent World-Action Model from Egocentric Videos}},
  author = {Hao Luo and Wanpeng Zhang and Yicheng Feng and Sipeng Zheng and Haiweng Xu and
            Chaoyi Xu and Ziheng Xi and Yuhui Fu and Zongqing Lu},
  journal = {arXiv preprint arXiv:2605.00078},
  year = {2026},
  eprint = {2605.00078},
  archivePrefix = {arXiv},
  url = {https://arxiv.org/abs/2605.00078},
}

@article{openwam_alpha,
  title = {{OpenWAM: An Open, Modular Exploration Towards Systematic World-Action Model Pretraining}},
  author = {Yuran Wang and Siqiao Huang and Mingleyang Li and Chenhao Zhang and Jiaqi Liang and
            Weiyang Jin and Yue Chen and Xuemin Chi and Donghao Zhou and Qize Yu and Yu-Kai Wang and
            Yuhan Rui and Shenzhe Yao and Zhen Yuan and Zhenhao Shen and Kefei Zhu and Zijie Zhu and
            Ning Gao and Xiaowei Chi and Guanqi He and Shanghang Zhang and Hao Dong and Lin Shao and
            Hang Zhao},
  journal = {arXiv preprint arXiv:2609.07398},
  year = {2026},
  eprint = {2609.07398},
  archivePrefix = {arXiv},
  url = {https://arxiv.org/abs/2609.07398},
}

@article{x_wam,
  title = {{Unified 4D World Action Modeling from Video Priors with Asynchronous Denoising}},
  author = {Jun Guo and Qiwei Li and Peiyan Li and Zilong Chen and Nan Sun and Yifei Su and
            Heyun Wang and Yuan Zhang and Xinghang Li and Huaping Liu},
  journal = {arXiv preprint arXiv:2604.26694},
  year = {2026},
  eprint = {2604.26694},
  archivePrefix = {arXiv},
  url = {https://arxiv.org/abs/2604.26694},
}

@article{aha_wam,
  title = {{AHA-WAM: Asynchronous Horizon-Adaptive World-Action Modeling with Observation-Guided Context Routing}},
  author = {Jisong Cai and Long Ling and Shiwei Chu and Zhongshan Liu and Jiayue Kang and
            Zhixuan Liang and Wenjie Xu and Yinan Mao and Weinan Zhang and Xiaokang Yang and
            Ru Ying and Ran Zheng and Yao Mu},
  journal = {arXiv preprint arXiv:2606.09811},
  year = {2026},
  eprint = {2606.09811},
  archivePrefix = {arXiv},
  url = {https://arxiv.org/abs/2606.09811},
}

@misc{groot_n17,
  title = {{NVIDIA Isaac GR00T N1.7-3B}},
  author = {{NVIDIA}},
  year = {2026},
  howpublished = {Hugging Face model card},
  note = {Model version N1.7. Accessed: 2026-09-18},
  url = {https://huggingface.co/nvidia/GR00T-N1.7-3B},
}

@inproceedings{x_vla,
  title = {{X-VLA: Soft-Prompted Transformer as Scalable Cross-Embodiment Vision-Language-Action Model}},
  author = {Jinliang Zheng and Jianxiong Li and Zhihao Wang and Dongxiu Liu and Xirui Kang and
            Yuchun Feng and Yinan Zheng and Jiayin Zou and Yilun Chen and Jia Zeng and
            Ya-Qin Zhang and Jiangmiao Pang and Jingjing Liu and Tai Wang and Xianyuan Zhan},
  booktitle = {International Conference on Learning Representations},
  year = {2026},
  url = {https://arxiv.org/abs/2510.10274},
}

@inproceedings{spatial_forcing,
  title = {{Spatial Forcing: Implicit Spatial Representation Alignment for Vision-Language-Action Model}},
  author = {Fuhao Li and Wenxuan Song and Han Zhao and Jingbo Wang and Pengxiang Ding and
            Donglin Wang and Long Zeng and Haoang Li},
  booktitle = {International Conference on Learning Representations},
  year = {2026},
  url = {https://arxiv.org/abs/2510.12276},
}

@article{abot_m0,
  title = {{ABot-M0: VLA Foundation Model for Robotic Manipulation with Action Manifold Learning}},
  author = {Yandan Yang and Shuang Zeng and Tong Lin and Xinyuan Chang and Dekang Qi and
            Junjin Xiao and Haoyun Liu and Ronghan Chen and Yuzhi Chen and Dongjie Huo and
            Feng Xiong and Xing Wei and Zhiheng Ma and Mu Xu},
  journal = {arXiv preprint arXiv:2602.11236},
  year = {2026},
  eprint = {2602.11236},
  archivePrefix = {arXiv},
  url = {https://arxiv.org/abs/2602.11236},
}

@article{gigabrain07,
  title = {{GigaBrain-0.7: Scaling Embodied Foundation Models to Emergent Capabilities with a Three-System Architecture}},
  author = {{GigaBrain Team} and Angen Ye and Axiang Sun and Can Jin and Chenxi Cheng and
            Chong Shi and Dengke Shang and Dingqian Zhang and Guan Huang and Guangqiang Wang and
            Guangqing Ding and Guo Li and Hangcong Li and Hengyu Zhong and Hongtao Lu and
            Jianbo Qin and Jiming Mao and Jing Zhu and Jindi Lv and Jingzhi Cui and Junjie Xie and
            Junyi Bao and Kai Liu and Lei Yuan and Limin Long and Lv Feng and Mingming Yu and
            Peng Li and Pengfei Yi and Qi Li and Qianli Zhang and Qingfang Li and Qitang Hu and
            Rui Zhang and Shaoyan Sun and Shibo Sun and Shiying Duan and Tenghui Chen and
            Tianze Liu and Weijie Ke and Wenyao Xue and Xiaofeng Wang and Xiaoyu Tian and
            Xinyu Liu and Xinze Chen and Yang Wang and Yankai Wang and Yejun Zeng and Yifan Li and
            Yifei Nie and Yilong Li and Yilong Liu and Yongchao Feng and Yumeng Wang and Yun Ye and
            Zhichao Liu and Ziheng He and Zonghai Yang and Zheng Zhu},
  journal = {arXiv preprint arXiv:2608.15875},
  year = {2026},
  eprint = {2608.15875},
  archivePrefix = {arXiv},
  url = {https://arxiv.org/abs/2608.15875},
}

@article{galaxea_g05,
  title = {{G0.5: One Autoregressive Stream for Robot Reasoning and Action}},
  author = {Yicheng Liu and Zibin Dong and Baijun Ye and Tianyuan Yuan and Tao Jiang and
            Anqi Yang and Shicheng Cao and Haonan Liu and Yue Sun and Zihan Guo and Xiao Liu and
            Dong Ke and Changxun Pan and Chenru Wu and Tailai Cheng and Xiaoshu Ren and
            Xinlei Zhang and Jianning Cui and Zijie Zhao and Haoyu Zhang and Kaiming Xu and
            Haodong Yang and Bowen Zhang and Jiahui Niu and Shaoting Zhu and Shiduo Zhang and
            Hang Zhao},
  journal = {arXiv preprint arXiv:2608.11739},
  year = {2026},
  eprint = {2608.11739},
  archivePrefix = {arXiv},
  url = {https://arxiv.org/abs/2608.11739},
}

@misc{dm05,
  title = {{DM0.5: From the Lab to the Open World}},
  author = {{Dexmal}},
  year = {2026},
  howpublished = {Technical blog and model release},
  note = {Released 2026-07-09. Accessed: 2026-09-18},
  url = {https://www.dexmal.com/blog/dm0.5?lang=en-US},
}

@article{gigaworld_policy,
  title = {{GigaWorld-Policy: An Efficient Action-Centered World--Action Model}},
  author = {Angen Ye and Boyuan Wang and Chaojun Ni and Guan Huang and Guosheng Zhao and Hao Li and
            Hengtao Li and Jie Li and Jindi Lv and Jingyu Liu and Min Cao and Peng Li and
            Qiuping Deng and Wenjun Mei and Xiaofeng Wang and Xinze Chen and Xinyu Zhou and
            Yang Wang and Yifan Chang and Yifan Li and Yukun Zhou and Yun Ye and Zhichao Liu and
            Zheng Zhu},
  journal = {arXiv preprint arXiv:2603.17240},
  year = {2026},
  eprint = {2603.17240},
  archivePrefix = {arXiv},
  url = {https://arxiv.org/abs/2603.17240},
}

@misc{wangEgo2RobotScalableRobot2026,
	title = {{Ego2Robot}: {Scalable} {Robot} {Data} {Synthesis} from {Egocentric} {Human} {Data}},
	url = {http://arxiv.org/abs/2608.02580},
	doi = {10.48550/arXiv.2608.02580},
	language = {en},
	publisher = {arXiv},
	author = {Wang, Ye and Lin, Pei and Chen, Xiong-Hui and Yuan, Haoqi and Liang, Zhixuan and Huang, Yiyang and Chen, Anzhe and Lei, Zixing and Zhang, Jie and Zhang, Tao and Li, Haoyang and Zhang, Tong and Xiao, Chenxi and Jiao, Ziyuan and Jin, Qin},
	year = {2026},
	note = {arXiv:2608.02580 [cs.RO]},
}

@inproceedings{todorov2012mujoco,
  title={MuJoCo: A physics engine for model-based control},
  author={Todorov, Emanuel and Erez, Tom and Tassa, Yuval},
  booktitle={2012 IEEE/RSJ International Conference on Intelligent Robots and Systems},
  pages={5026--5033},
  year={2012},
  organization={IEEE},
  doi={10.1109/IROS.2012.6386109}
}

@misc{carion_sam_2025,
	title = {{SAM} 3: {Segment} {Anything} with {Concepts}},
	copyright = {arXiv.org perpetual, non-exclusive license},
	shorttitle = {{SAM} 3},
	url = {https://arxiv.org/abs/2511.16719},
	doi = {10.48550/ARXIV.2511.16719},
	language = {en},
	urldate = {2026-09-20},
	publisher = {arXiv},
	author = {Carion, Nicolas and Gustafson, Laura and Hu, Yuan-Ting and Debnath, Shoubhik and Hu, Ronghang and Suris, Didac and Ryali, Chaitanya and Alwala, Kalyan Vasudev and Khedr, Haitham and Huang, Andrew and Lei, Jie and Ma, Tengyu and Guo, Baishan and Kalla, Arpit and Marks, Markus and Greer, Joseph and Wang, Meng and Sun, Peize and Rädle, Roman and Afouras, Triantafyllos and Mavroudi, Effrosyni and Xu, Katherine and Wu, Tsung-Han and Zhou, Yu and Momeni, Liliane and Hazra, Rishi and Ding, Shuangrui and Vaze, Sagar and Porcher, Francois and Li, Feng and Li, Siyuan and Kamath, Aishwarya and Cheng, Ho Kei and Dollár, Piotr and Ravi, Nikhila and Saenko, Kate and Zhang, Pengchuan and Feichtenhofer, Christoph},
	year = {2025},
	note = {Version Number: 2},
}

@inproceedings{zhou_propainter_2023,
	address = {Paris, France},
	title = {{ProPainter}: {Improving} {Propagation} and {Transformer} for {Video} {Inpainting}},
	shorttitle = {{ProPainter}},
	url = {https://ieeexplore.ieee.org/document/10378438/},
	doi = {10.1109/ICCV51070.2023.00961},
	language = {en},
	urldate = {2026-09-20},
	booktitle = {2023 {IEEE}/{CVF} {International} {Conference} on {Computer} {Vision} ({ICCV})},
	publisher = {IEEE},
	author = {Zhou, Shangchen and Li, Chongyi and Chan, Kelvin C.K. and Change, Chen},
	month = oct,
	year = {2023},
	pages = {10443--10452},
}

@inproceedings{hu2025videopredictionpolicy,
  title = {{Video Prediction Policy: A Generalist Robot Policy with Predictive Visual Representations}},
  author = {Hu, Yucheng and Guo, Yanjiang and Wang, Pengchao and Chen, Xiaoyu and Wang, Yen-Jen and Zhang, Jianke and Sreenath, Koushil and Lu, Chaochao and Chen, Jianyu},
  booktitle = {Proceedings of the 42nd International Conference on Machine Learning},
  pages = {24328--24346},
  year = {2025},
  volume = {267},
  series = {Proceedings of Machine Learning Research},
  publisher = {PMLR},
  url = {https://proceedings.mlr.press/v267/hu25g.html},
  eprint = {2412.14803},
  archivePrefix = {arXiv},
}

@article{bardes2024vjepa,
  title = {{Revisiting Feature Prediction for Learning Visual Representations from Video}},
  author = {Bardes, Adrien and Garrido, Quentin and Ponce, Jean and Chen, Xinlei and Rabbat, Michael and LeCun, Yann and Assran, Mahmoud and Ballas, Nicolas},
  journal = {arXiv preprint arXiv:2404.08471},
  year = {2024},
  eprint = {2404.08471},
  archivePrefix = {arXiv},
  url = {https://arxiv.org/abs/2404.08471},
}

@article{assran2025vjepa2,
  title = {{V-JEPA 2: Self-Supervised Video Models Enable Understanding, Prediction and Planning}},
  author = {Assran, Mido and Bardes, Adrien and Fan, David and Garrido, Quentin and Howes, Russell and Komeili, Mojtaba and Muckley, Matthew J. and Rizvi, Ammar and Roberts, Claire and Sinha, Koustuv and Zholus, Artem and Arnaud, Sergio and Gejji, Abha and Martin, Ada and Hogan, Francois Robert and Dugas, Daniel and Bojanowski, Piotr and Khalidov, Vasil and Labatut, Patrick and Massa, Francisco and Szafraniec, Marc and Krishnakumar, Kapil and Li, Yong and Ma, Xiaodong and Chandar, Sarath and Meier, Franziska and LeCun, Yann and Rabbat, Michael and Ballas, Nicolas},
  journal = {arXiv preprint arXiv:2506.09985},
  year = {2025},
  eprint = {2506.09985},
  archivePrefix = {arXiv},
  url = {https://arxiv.org/abs/2506.09985},
}

@article{kim2024openvla,
  title = {{OpenVLA: An Open-Source Vision-Language-Action Model}},
  author = {Moo Jin Kim and Karl Pertsch and Siddharth Karamcheti and Ted Xiao and
            Ashwin Balakrishna and Suraj Nair and Rafael Rafailov and Ethan Foster and
            Grace Lam and Pannag Sanketi and Quan Vuong and Thomas Kollar and
            Benjamin Burchfiel and Russ Tedrake and Dorsa Sadigh and Sergey Levine and
            Percy Liang and Chelsea Finn},
  journal = {arXiv preprint arXiv:2406.09246},
  year = {2024},
  eprint = {2406.09246},
  archivePrefix = {arXiv},
  url = {https://arxiv.org/abs/2406.09246},
}

@article{bjorck2025groot,
  title = {{GR00T N1: An Open Foundation Model for Generalist Humanoid Robots}},
  author = {{NVIDIA} and Johan Bjorck and Fernando Casta{\~n}eda and Nikita Cherniadev and
            Xingye Da and Runyu Ding and Linxi "Jim" Fan and Yu Fang and Dieter Fox and
            Fengyuan Hu and Spencer Huang and Joel Jang and Zhenyu Jiang and Jan Kautz and
            Kaushil Kundalia and Lawrence Lao and Zhiqi Li and Zongyu Lin and Kevin Lin and
            Guilin Liu and Edith Llontop and Loic Magne and Ajay Mandlekar and Avnish Narayan and
            Soroush Nasiriany and Scott Reed and You Liang Tan and Guanzhi Wang and Zu Wang and
            Jing Wang and Qi Wang and Jiannan Xiang and Yuqi Xie and Yinzhen Xu and
            Zhenjia Xu and Seonghyeon Ye and Zhiding Yu and Ao Zhang and Hao Zhang and
            Yizhou Zhao and Ruijie Zheng and Yuke Zhu},
  journal = {arXiv preprint arXiv:2503.14734},
  year = {2025},
  eprint = {2503.14734},
  archivePrefix = {arXiv},
  url = {https://arxiv.org/abs/2503.14734},
}

@article{wu2026lingbotvla,
  title = {{A Pragmatic VLA Foundation Model}},
  author = {Wei Wu and Fan Lu and Yunnan Wang and Shuai Yang and Shi Liu and Fangjing Wang and
            Qian Zhu and He Sun and Yong Wang and Shuailei Ma and Yiyu Ren and Kejia Zhang and
            Hui Yu and Jingmei Zhao and Shuai Zhou and Zhenqi Qiu and Houlong Xiong and
            Ziyu Wang and Zechen Wang and Ran Cheng and Yong-Lu Li and Yongtao Huang and
            Xing Zhu and Yujun Shen and Kecheng Zheng},
  journal = {arXiv preprint arXiv:2601.18692},
  year = {2026},
  eprint = {2601.18692},
  archivePrefix = {arXiv},
  url = {https://arxiv.org/abs/2601.18692},
}

@article{wu2026lingbotvla2,
  title = {{From Foundation to Application: Improving VLA Models in Practice}},
  author = {Wei Wu and Fangjing Wang and Fan Lu and He Sun and Shi Liu and Yunnan Wang and
            Yibin Yan and Yong Wang and Shuailei Ma and Xinyang Wang and Yibin Liu and
            Shuai Yang and Tianxiang Zhou and Kejia Zhang and Lei Zhou and Cheng Su and
            Nan Xue and Bin Tan and Han Zhang and Youchao Zhang and Fei Liao and Xing Zhu and
            Yujun Shen and Kecheng Zheng},
  journal = {arXiv preprint arXiv:2607.06403},
  year = {2026},
  eprint = {2607.06403},
  archivePrefix = {arXiv},
  url = {https://arxiv.org/abs/2607.06403},
}

@article{bi2025motus,
  title = {{Motus: A Unified Latent Action World Model}},
  author = {Hongzhe Bi and Hengkai Tan and Shenghao Xie and Zeyuan Wang and Shuhe Huang and
            Haitian Liu and Ruowen Zhao and Yao Feng and Chendong Xiang and Yinze Rong and
            Hongyan Zhao and Hanyu Liu and Zhizhong Su and Lei Ma and Hang Su and Jun Zhu},
  journal = {arXiv preprint arXiv:2512.13030},
  year = {2025},
  eprint = {2512.13030},
  archivePrefix = {arXiv},
  url = {https://arxiv.org/abs/2512.13030},
}

@article{ye2026dreamzero,
  title = {{World Action Models are Zero-shot Policies}},
  author = {Seonghyeon Ye and Yunhao Ge and Kaiyuan Zheng and Shenyuan Gao and Sihyun Yu and
            George Kurian and Suneel Indupuru and You Liang Tan and Chuning Zhu and Jiannan Xiang and
            Ayaan Malik and Kyungmin Lee and William Liang and Nadun Ranawaka and Jiasheng Gu and
            Yinzhen Xu and Guanzhi Wang and Fengyuan Hu and Avnish Narayan and Johan Bjorck and
            Jing Wang and Gwanghyun Kim and Dantong Niu and Ruijie Zheng and Yuqi Xie and Jimmy Wu and
            Qi Wang and Ryan Julian and Danfei Xu and Yilun Du and Yevgen Chebotar and Scott Reed and
            Jan Kautz and Yuke Zhu and Linxi "Jim" Fan and Joel Jang},
  journal = {arXiv preprint arXiv:2602.15922},
  year = {2026},
  eprint = {2602.15922},
  archivePrefix = {arXiv},
  url = {https://arxiv.org/abs/2602.15922},
}

@article{zhang2026lingbotva2,
  title = {{Native Video-Action Pretraining for Generalizable Robot Control}},
  author = {Qihang Zhang and Lin Li and Luyao Zhang and Shuai Yang and Yiming Luo and Shuaiting Li and
            Ruilin Wang and Junke Wang and Jiahao Shao and Gangwei Xu and Jiaming Zhou and Yishu Shen and
            Yudong Jin and Fangyi Xu and Shuailei Ma and Jiaqi Liao and Guanxing Lu and Zifan Shi and
            Yongkun Wen and Yujie Zhao and Weixuan Tang and Xinyang Wang and Chaojian Li and Jiapeng Zhu and
            Ka Leong Cheng and Nan Xue and Xing Zhu and Yujun Shen and Yinghao Xu},
  journal = {arXiv preprint arXiv:2607.08639},
  year = {2026},
  eprint = {2607.08639},
  archivePrefix = {arXiv},
  url = {https://arxiv.org/abs/2607.08639},
}

@misc{zhang2026unexpectedrobotpolicyearly,
      title={An Unexpected Robot Policy: Early Evaluations of GPT-6 Astra on RoboDojo and Beyond}, 
      author={Wenbo Zhang and Kaixuan Wang and Yutao Ouyang and Xiaoyu Huang and Liyang Li and Kailun Su and Weiyang Jin and Wenhao Chai and Haotian Liang and Zhiyang Dou and Yue Chen and Tianxing Chen},
      year={2026},
      eprint={2609.24170},
      archivePrefix={arXiv},
      primaryClass={cs.CV},
      url={https://arxiv.org/abs/2609.24170}, 
}

@inproceedings{lerobot,
  title = {{LeRobot: An Open-Source Library for End-to-End Robot Learning}},
  author = {Cadene, Remi and Alibert, Simon and Capuano, Francesco and Aractingi, Michel and Zouitine, Adil and Kooijmans, Pepijn and Choghari, Jade and Russi, Martino and Pascal, Caroline and Palma, Steven and Shukor, Mustafa and Moss, Jess and Soare, Alexander and Aubakirova, Dana and Lhoest, Quentin and Gallou{\'e}dec, Quentin and Wolf, Thomas},
  booktitle = {The Fourteenth International Conference on Learning Representations},
  year = {2026},
  eprint = {2602.22818},
  archivePrefix = {arXiv},
  url = {https://arxiv.org/abs/2602.22818},
}

@misc{openpi,
  title = {{OpenPI: Open-Source Models and Packages for Robotics}},
  author = {{Physical Intelligence}},
  year = {2025},
  howpublished = {GitHub repository},
  url = {https://github.com/Physical-Intelligence/openpi},
}

@misc{rtc_inference_time_arxiv2506_07339,
  title         = {Real-Time Execution of Action Chunking Flow Policies},
  author        = {Black, Kevin and Galliker, Manuel Y. and Levine, Sergey},
  year          = {2025},
  eprint        = {2506.07339},
  archivePrefix = {arXiv},
  primaryClass  = {cs.RO},
  doi           = {10.48550/arXiv.2506.07339},
  url           = {https://arxiv.org/abs/2506.07339}
}

@misc{motubrain2026realtime,
  title         = {World Action Models in Real Time: An Empirical Study of Smooth Execution via Asynchronous Deployment},
  author        = {{Motubrain Team}},
  year          = {2026},
  eprint        = {2608.01880},
  archivePrefix = {arXiv},
  primaryClass  = {cs.RO},
  url           = {https://arxiv.org/abs/2608.01880}
}

@misc{dang2026rynnbrainopenembodiedfoundation,
      title={RynnBrain: Open Embodied Foundation Models}, 
      author={Ronghao Dang and Jiayan Guo and Bohan Hou and Sicong Leng and Kehan Li and Xin Li and Jiangpin Liu and Yunxuan Mao and Zhikai Wang and Yuqian Yuan and Minghao Zhu and Xiao Lin and Yang Bai and Qian Jiang and Yaxi Zhao and Minghua Zeng and Junlong Gao and Yuming Jiang and Jun Cen and Siteng Huang and Liuyi Wang and Wenqiao Zhang and Chengju Liu and Jianfei Yang and Shijian Lu and Deli Zhao},
      year={2026},
      eprint={2602.14979},
      archivePrefix={arXiv},
      primaryClass={cs.RO},
      url={https://arxiv.org/abs/2602.14979}, 
}

@article{team2026xiaomi,
  title={Xiaomi-Robotics-1: Scaling Vision-Language-Action Models with over 100K Hours of Real-World Trajectories},
  author={Team, Xiaomi Robotics and Guo, Jun and Jin, Piaopiao and Li, Jason and Li, Peiyan and Li, Yingyan and Liu, Futeng and Peng, Wanli and Qin, Optimus and Su, Yifei and others},
  journal={arXiv preprint arXiv:2607.15330},
  year={2026}
}

@misc{su2026astra,
  title        = {{GPT 6 Astra} as an Embodied Policy},
  author       = {Su, Jiayi and Zheng, Yixin and Yan, Mi and Yi, Li and Zhang, Zhizheng and Wang, He},
  year         = {2026},
  howpublished = {Technical report and code},
  url          = {https://github.com/anonymous-report-421/eval-of-gpt-6-astra-as-policy}
}

\clearpage
\appendix
\appendix
\section{Human-to-Robot IK Details}
\label{app:h2r-ik}

At each local IK iteration for a fixed robot base placement, the solver combines
position and approach tracking, contact-point and orientation refinement, and
projected joint-space regularization:
\begin{equation}
    \Delta\mathbf{q}
    = \underbrace{\Delta\mathbf{q}_p}_{\substack{\text{position and}\\\text{approach tracking}}}
    + \underbrace{\Delta\mathbf{q}_s}_{\substack{\text{contact-point and}\\\text{orientation refinement}}}
    + \underbrace{\mathbf{N}_p\mathbf{g}_{\mathrm{aux}}}_{\substack{\text{joint-space}\\\text{regularization}}},
    \label{eq:h2r-ik-hierarchical-update}
\end{equation}
where $\mathbf{q}$ is the current joint configuration and $\Delta\mathbf{q}$ is
its local update. The increment $\Delta\mathbf{q}_p$ tracks TCP position and
approach direction, $\Delta\mathbf{q}_s$ refines contact-point alignment and
orientation through the damped projection $\mathbf{N}_p$, and
$\mathbf{N}_p\mathbf{g}_{\mathrm{aux}}$ applies projected joint-space
preferences. The subscript $p$ denotes position and approach tracking, $s$
denotes contact-point and orientation refinement, and $\mu$ is the
pseudoinverse damping. The individual terms and update bounds are specified
below.

\textit{Position and approach tracking.}
This step reduces TCP position error and approach-direction error beyond
the allowed tolerance:
\begin{equation}
    \Delta\mathbf{q}_p = \mathbf{J}_{p,\mu}^{\#}\mathbf{e}_p,
    \qquad
    \mathbf{N}_p = \mathbf{I}-\mathbf{J}_{p,\mu}^{\#}\mathbf{J}_p.
    \label{eq:h2r-ik-projection}
\end{equation}
Here $(\mathbf{J}_p,\mathbf{e}_p)$ contains the weighted position-and-approach
Jacobian and residual. The damped inverse is defined by
$\mathbf{J}^{\#}_{\mu}=\mathbf{J}^{\top}
(\mathbf{J}\mathbf{J}^{\top}+\mu\mathbf{I})^{-1}$,
with adaptive damping $\mu>0$.

\textit{Contact-point and orientation refinement.}
Optional gripper contact-point and full-orientation targets are refined after
the position-and-approach step. Their correction accounts for the residual left
by that step and uses the position-and-approach projection $\mathbf{N}_p$ to
attenuate interference:
\begin{equation}
    \Delta\mathbf{q}_s
    = \mathbf{N}_p(\mathbf{J}_s\mathbf{N}_p)^{\#}_{4\mu}
      (\mathbf{e}_s-\mathbf{J}_s\Delta\mathbf{q}_p),
\end{equation}
where $(\mathbf{J}_s,\mathbf{e}_s)$ stacks the weighted Jacobians and residuals
of the enabled contact-point and orientation terms. Contact-point alignment
drives the left and right gripper-pad centers toward two virtual points on
the human pinch axis. These targets specify geometric alignment. This term
is zero when neither refinement is active.

\textit{Joint-space regularization.}
The auxiliary adjustment encourages continuity, a preferred home configuration,
and separation from joint limits:
\begin{equation}
    \mathbf{g}_{\mathrm{aux}}
    = \lambda_r(\mathbf{q}^{\mathrm{ref}}-\mathbf{q})
    + \lambda_h(\mathbf{q}^{\mathrm{home}}-\mathbf{q})
    + \lambda_l\mathbf{d}_{\mathrm{lim}}.
\end{equation}
The reference $\mathbf{q}^{\mathrm{ref}}$ is the warm-reference configuration when
available, or otherwise the continuation seed. Attraction to this reference is
proportional to the negative gradient of
$\tfrac{1}{2}\|\mathbf{q}-\mathbf{q}^{\mathrm{ref}}\|_2^2$.
The remaining terms pull toward the home configuration $\mathbf{q}^{\mathrm{home}}$
and follow the joint-limit avoidance direction $\mathbf{d}_{\mathrm{lim}}$.
The weights $\lambda_r$, $\lambda_h$, and $\lambda_l$ control the respective
contributions, with disabled terms set to zero. All three contributions pass
through $\mathbf{N}_p$ in the total update.

The damped projection is not an exact null-space projector. The combined increment
is clipped before updating the joint configuration, followed by joint-limit
enforcement and bounds on displacement from the continuation seed. Accepted
iterates are selected using task-priority criteria rather than a single aggregate
loss. Continuity depends on reference attraction, solution reuse, step bounds,
and trajectory-level validation.

\end{document}